\documentclass[sn-mathphys,pdflatex,Numbered]{sn-jnl}

\usepackage{graphicx}%
\usepackage{rotating}
\usepackage{booktabs}%
\usepackage{longtable}
\usepackage{multirow}%
\usepackage{lscape}
\usepackage{float}
\usepackage{amsmath,amssymb,amsfonts}%
\usepackage{amsthm}%
\usepackage{mathrsfs}%
\usepackage[title]{appendix}%
\usepackage[dvipsnames,table,xcdraw]{xcolor}%
\usepackage{textcomp}%
\usepackage{manyfoot}%
\usepackage{listings}%
\usepackage{bookmark}
\usepackage{natbib}
\usepackage{geometry}
\usepackage{listings}
\usepackage[frozencache,cachedir=.]{minted}
\usepackage{soul}
\usepackage[many]{tcolorbox}
\tcbuselibrary{breakable}

\newcommand{\hlgreen}[2][LimeGreen!30]{\sethlcolor{#1}\hl{#2}}
\newcommand{\hlyellow}[2][Yellow!30]{\sethlcolor{#1}\hl{#2}}
\newcommand{\hlred}[2][Red!30]{\sethlcolor{#1}\hl{#2}}
\newcommand{\hlpurple}[2][Orchid!30]{\sethlcolor{#1}\hl{#2}}
\newcommand{\hlpink}[2][Lavender!30]{\sethlcolor{#1}\hl{#2}}
\newcommand{\hlblue}[2][SkyBlue!30]{\sethlcolor{#1}\hl{#2}}
\newcommand{\hlorange}[2][Orange!30]{\sethlcolor{#1}\hl{#2}}
\newcommand{\hlbrown}[2][Brown!30]{\sethlcolor{#1}\hl{#2}}

\newminted{json}{
    breaklines,
    breakanywhere,
    breaksymbolleft=~~,
    breaksymbolright=~~,
    fontsize=\small,
}

\begin{document}

\title[Article Title]{Large Language Model Capabilities in Perioperative Risk Prediction and Prognostication}


\author[1]{\fnm{Philip} \sur{Chung}}
\author[2]{\fnm{Christine T.} \sur{Fong}}
\author[2]{\fnm{Andrew M.} \sur{Walters}}
\author[1]{\fnm{Nima} \sur{Aghaeepour}}
\author[3,4]{\fnm{Meliha} \sur{Yetisgen}}
\author[2]{\fnm{Vikas N.} \sur{O'Reilly-Shah}}

\affil[1]{\orgdiv{Department of Anesthesiology, Perioperative and Pain Medicine}, \orgname{Stanford University}}
\affil[2]{\orgdiv{Department Anesthesia \& Pain Medicine}, \orgname{University of Washington}}
\affil[3]{\orgdiv{Department of Biomedical Informatics and Medical Education}, \orgname{University of Washington}}
\affil[4]{\orgdiv{Department of Linguistics}, \orgname{University of Washington}}


\abstract{We investigate whether general-domain large language models such as GPT-4 Turbo can perform risk stratification and predict post-operative outcome measures using a description of the procedure and a patient’s clinical notes derived from the electronic health record. We examine predictive performance on 8 different tasks: prediction of ASA Physical Status Classification, hospital admission, ICU admission, unplanned admission, hospital mortality, PACU Phase 1 duration, hospital duration, and ICU duration. Few-shot and chain-of-thought prompting improves predictive performance for several of the tasks. We achieve F1 scores of 0.50 for ASA Physical Status Classification, 0.81 for ICU admission, and 0.86 for hospital mortality. Performance on duration prediction tasks were universally poor across all prompt strategies. Current generation large language models can assist clinicians in perioperative risk stratification on classification tasks and produce high-quality natural language summaries and explanations.}


\maketitle

\section{Introduction}\label{introduction}

Instruction-tuned large language models (LLMs) have been successful at knowledge retrieval\cite{Brown2020-sb, Ouyang2022-ko, Zhang2023-fh, Taori2023-cv}, text extraction \cite{Agrawal2022-rv, Singhal2023-yl, Ramachandran2023-to, Ramachandran2023-wk}, summarization\cite{Zhang2023-sv, Stiennon2020-tv, Wu2021-wz}, and reasoning\cite{Wei2022-kj, Wei2022-yk, Kojima2022-ls, Yao2022-pl, Yao2023-mx} tasks without requiring domain-specific fine-tuning. Prompting has emerged as a means for task and domain specification without requiring domain-specific fine-tuning\cite{Radford2019-tu}. General-domain models have been shown to excel at medicine-specific tasks such as United States Medical Licensing Exam (USMLE) questions\cite{Singhal2023-kc, Nori2023-vf, Nori2023-xz} or summarization of electronic health record text.\cite{Van_Veen2023-sr}

We investigate whether general-domain LLMs can perform the task of preoperative risk stratification and prognostication–that is, given information about the patient and a surgery or procedure, assign a risk score or predict a post-operative outcome metric. This assessment of post-operative outcome assists proceduralists, surgeons, and anesthesiologists weighing the risks and benefits of proceeding with a procedure versus canceling or delaying the procedure for medical optimization. 

Since there is no single post-operative outcome measure of risk, we survey LLM capabilities on 8 different tasks: (1) assignment of the American Society of Anesthesiologists Physical Status (ASA-PS) classification system\cite{Saklad1941-tp, Mayhew2019-cy, Horvath2021-nt}, (2) prediction of post-anesthesia care unit (PACU) duration, (3) prediction of hospital admission, (4) prediction of hospital duration, (5) prediction of intensive care unit (ICU) admission, (6) prediction of ICU duration, (7) prediction of whether the patient will have an unanticipated hospital admission, (8) prediction of whether the patient will die in the hospital. We only examine outcomes where quantitative ground truth labels can be derived from the electronic health record (EHR). We also explore the effect of commonly applied prompting techniques such as in-context (few-shot) learning and chain-of-thought (CoT) reasoning, which has been shown to improve question and answering (Q\&A) performance. In-context learning involves adding representative task \& solution examples into the prompt prior to the actual query task to demonstrate the desired pattern of task and response \cite{Brown2020-sb}. CoT is a prompting strategy that instructs language models to respond with step by step reasoning prior to providing a final answer \cite{Wei2022-yk, Kojima2022-ls}.

While LLMs have been previously explored in medical Q\&A with models attaining a passing score on the USMLE medical licensing exam, these Q\&A datasets are not reflective of the real-world clinical setting\cite{Singhal2023-kc, Nori2023-vf, Nori2023-xz}. Most of these datasets are multiple choice questions or have unambiguous answers that exist within a well-defined knowledge source such as a medical text book\cite{Wang2023-jn, Zakka2023-tz}. Real-world EHRs often contain patient contexts with uncertain, incomplete, or erroneous information, and a clear answer may be elusive. It is within this real-world context that we derive our dataset and conduct our investigations to benchmark the capabilities of LLMs in perioperative risk prediction and prognostication.

\begin{figure}[b]
    \centering
    \includegraphics[width=1\linewidth]{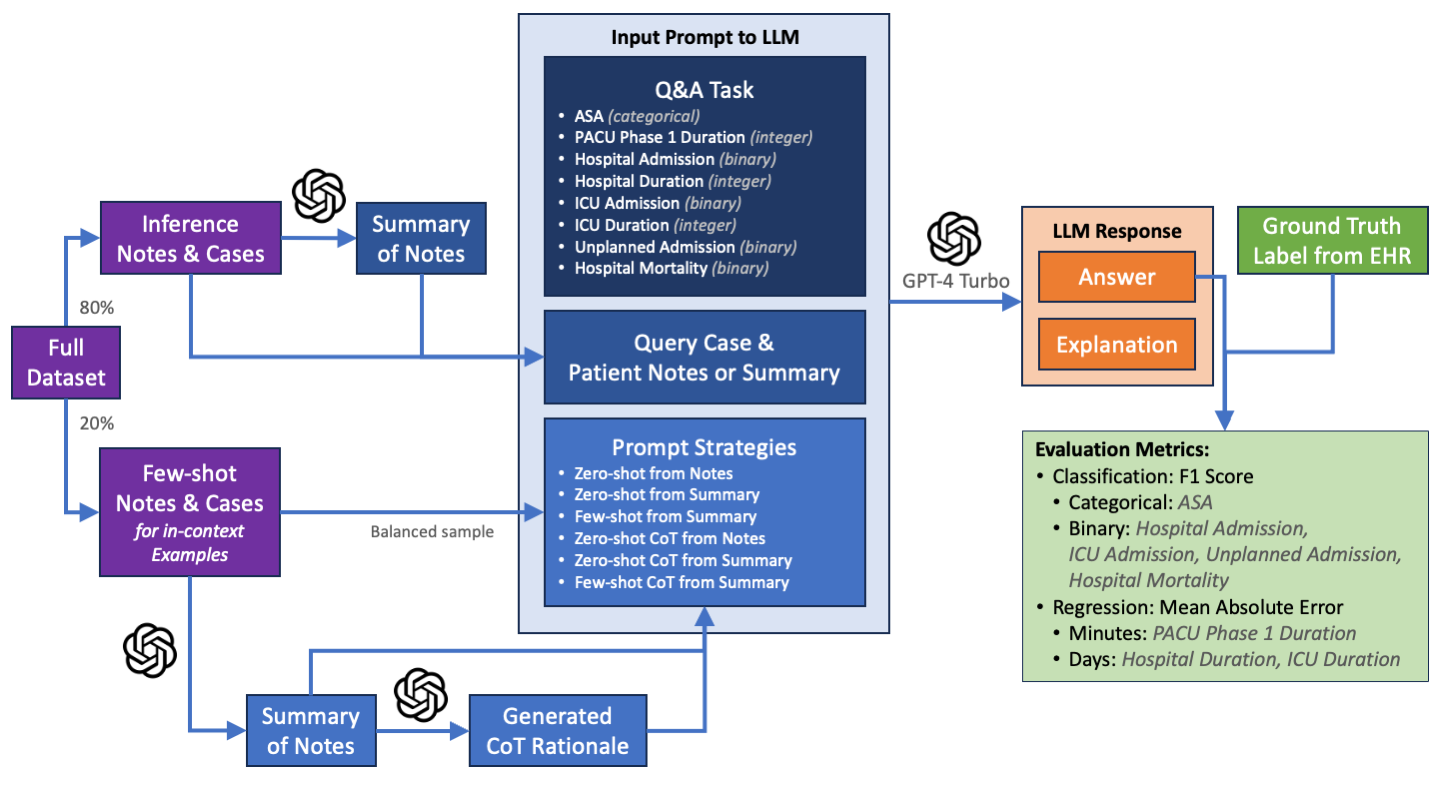}
    \caption{Overview of the experimental apparatus. GPT-4 Turbo is used as the large language model (LLM) in all steps. Each prompt to the LLM is unique based on the task, prompt strategy and query case for which an answer and explanation is generated. Zero-shot prompt strategy is conducted with both original clinical notes and a summary of the clinical notes. Few-shot prompts utilize in-context examples derived from the few-shot dataset. Each in-context example is a question, procedure description, summary of patient notes, and answer. Summaries are generated using LLM. The few-shot chain-of-thought (CoT) prompt strategy requires a CoT rationale for each in-context example that links the question to the answer, which is also generated using LLM. Answers provided by the LLM are compared against the ground truth label derived from electronic health record (EHR) data, and either F1 score or mean absolute error (MAE) is computed, depending on whether the outcome variable for the task is categorical/binary or integer.}
    \label{fig:fig1-experimentapparatus}
\end{figure}

\section{Methods}\label{methods}

This is a retrospective study of routinely collected health records data, approved by the University of Washington (UW) Institutional Review Board with a waiver of consent. The computational environment for use of protected health information (PHI) and personally identifiable information (PII) was reviewed and approved by UW Medicine Information Technology. The study followed the Transparent Reporting of a Multivariable Prediction Model for Individual Prognosis or Diagnosis (TRIPOD) guideline\cite{Collins2015-to}.

\subsection{Study Cohort and Dataset Definition}\label{methods-dataset}
Inclusion criteria were patients who had a surgery or procedure with anesthesia at 3 hospitals (UW Medical Center-Montlake, UW Medical Center-Northwest, Harborview Medical Center) in Seattle, WA from April 1, 2021 to May 5, 2023 where the patient also had an anesthesia preoperative evaluation note associated with the case and at least one other clinician note filed in the EHR prior to the case. The notes used in our experiment included up to the last 10 clinician-written notes filed in the EHR prior to the surgery, excluding the anesthesia preoperative evaluation note associated with the case. Short notes less than 100 token lengths were excluded. PACU phase 1 timestamps and Admit/Discharge/Transfer (ADT) events for these patients were used to determine postoperative PACU, hospital, and ICU admission and length of stay duration. Unanticipated admission was determined by comparing expected patient class in the surgeon’s case booking and actual postoperative patient class.

From the extracted cases, notes, and ADT events, seven datasets were created which were used for each of the eight prediction tasks (dataset for ICU admission and duration prediction are the same). For patients who had multiple cases, a single case was randomly selected. The case number was then downsampled to a target of 1250 cases and then split 80\%-20\% into inference and few-shot data splits for 1000 inference cases and 250 few-shot cases. The few-shot data splits are held-out sets that are only used for crafting in-context examples for few-shot prompting. We employed the following strategies to combat class imbalance and data skew in the outcome variables: (1) The ASA-PS dataset was constructed with inverse frequency sampling of the ASA-PS; (2) PACU and hospital durations were binned to form 20 and 100 groups, respectively, and the datasets were constructed with inverse frequency sampling of the groups; (3) For hospital admission, the outer product of whether a patient was admitted and the patient class was used to form groups, and then dataset was constructed using inverse frequency sampling of the groups; (4) For ICU admission, ICU duration, unplanned admission, and hospital mortality datasets, the outcome variables were rare occurrences, so we sampled a 50\%-50\% mix of cases with the presence and absence of ICU admission, unplanned admission, and hospital mortality for each respective dataset.

\subsection{Experimental Approach}\label{methods-experiment}
An overview of the experimental apparatus is shown in Figure \ref{fig:fig1-experimentapparatus}. We used GPT-4 Turbo (gpt-4-1106)\cite{noauthor_undated-cx} as our LLM through Microsoft Azure OpenAI Service with disabled content filters because of the presence of PHI/PII. Temperature and top p parameters were set to 1.0 for all experiments. A unique random seed was used for each case in the datasets. If an LLM generation was unsuccessful or the response was inappropriately formatted and cannot be extracted, the LLM generation was retried with a different random seed up to 5 times. For each of the 8 prediction tasks, we experiment with 6 prompting strategies: (1) zero-shot Q\&A using original notes, (2) zero-shot Q\&A using note summaries, (3) few-shot Q\&A using note summaries, (4) zero-shot CoT \&A using original notes, (5) zero-shot CoT Q\&A using note summaries, (6) few-shot CoT Q\&A using note summaries.  For few-shot prompts, we used 5-shot, 10-shot, 20-shot, and 50-shot in-context examples. Each prompt contained an instruction, task/question, procedure information (procedure, procedure description, diagnosis, provider service), patient representation (original notes or note summaries), and instruction on how to format the response. Note summaries were generated by GPT-4 Turbo. Few-shot prompt in-context examples were drawn only from the few-shot dataset, using inverse frequency sampling to balance the representation of outcome variables in the in-context examples. For few-shot CoT Q\&A prompts, each in-context example had a CoT rationale generated by GPT-4 Turbo specific to the in-context example and task. When CoT prompting was used, the LLM was instructed to provide a step-by-step explanation prior to the answer to allow the LLM to attend to the explanation while generating the answer. When CoT prompting was not used, the LLM was instructed to generate an answer and then provide an explanation. Representative text prompts used for experiments are depicted in Supplemental Figure \ref{suppfig1-prompts}. 

We applied each of the prompting strategies to the 8 tasks. For each example, the answer was compared against the ground truth label derived from the EHR. F1 score was the primary measure of performance for categorical and binary outcomes and mean absolute error (MAE) was the primary measure of performance for duration outcomes. ASA-PS prediction was a 6-class prediction task whereas hospital admission, ICU admission, unplanned admission, and hospital mortality prediction tasks were binary classification tasks. PACU Phase 1 duration was measured in minutes whereas hospital duration and ICU duration were measured in days. Reported metrics and 95\% confidence intervals were estimated using 2500 bootstrap iterations. We also stratified the cases based on token length of all 10 notes into three categories–short, medium, and long–and reported the performance of each strata. Statistical significance testing was performed using Wilcoxon signed-rank test for pairwise comparison of prompt strategies.

\section{Results}\label{results}

\subsection{Datasets}\label{results-datasets}
Our inclusion criteria yielded 90,844 patients undergoing 137,535 surgeries or procedures with anesthesia care, 133,500 anesthesia preoperative evaluation notes, 226,821 other clinician-written notes, and 2,253,286 admit/discharge/transfer events, which was then processed and downsampled to construct the task-specific datasets used for our experiments. The hospital mortality and unplanned admission outcomes were rare and resulted in datasets with 720 and 1186 cases prior to the 80\%-20\% data split. All other datasets reached the target size of 1250 cases followed by 80\%-20\% split for 1000 inference cases and 250 few-shot cases per task. Demographics, distribution of input and outcome variables for each dataset and data split are described in Table \ref{tab:table1}. Details to dataset creation is shown in Supplemental Figure \ref{fig:suppfig2-flowdiagram}. The degree of overlap of cases in each of the datasets is shown in Supplemental Figure \ref{fig:suppfig3-datasetoverlap}. Counts for note type and author provider type are shown in Supplemental Table \ref{tab:supptable1-notetypeauthortype}. The direct Microsoft Azure API costs of conducting these experiments was \$6000 (Supplemental Table \ref{tab:supptable2-experimentcosts}).

\begin{figure}[b!]
    \centering
    \includegraphics[width=1\linewidth]{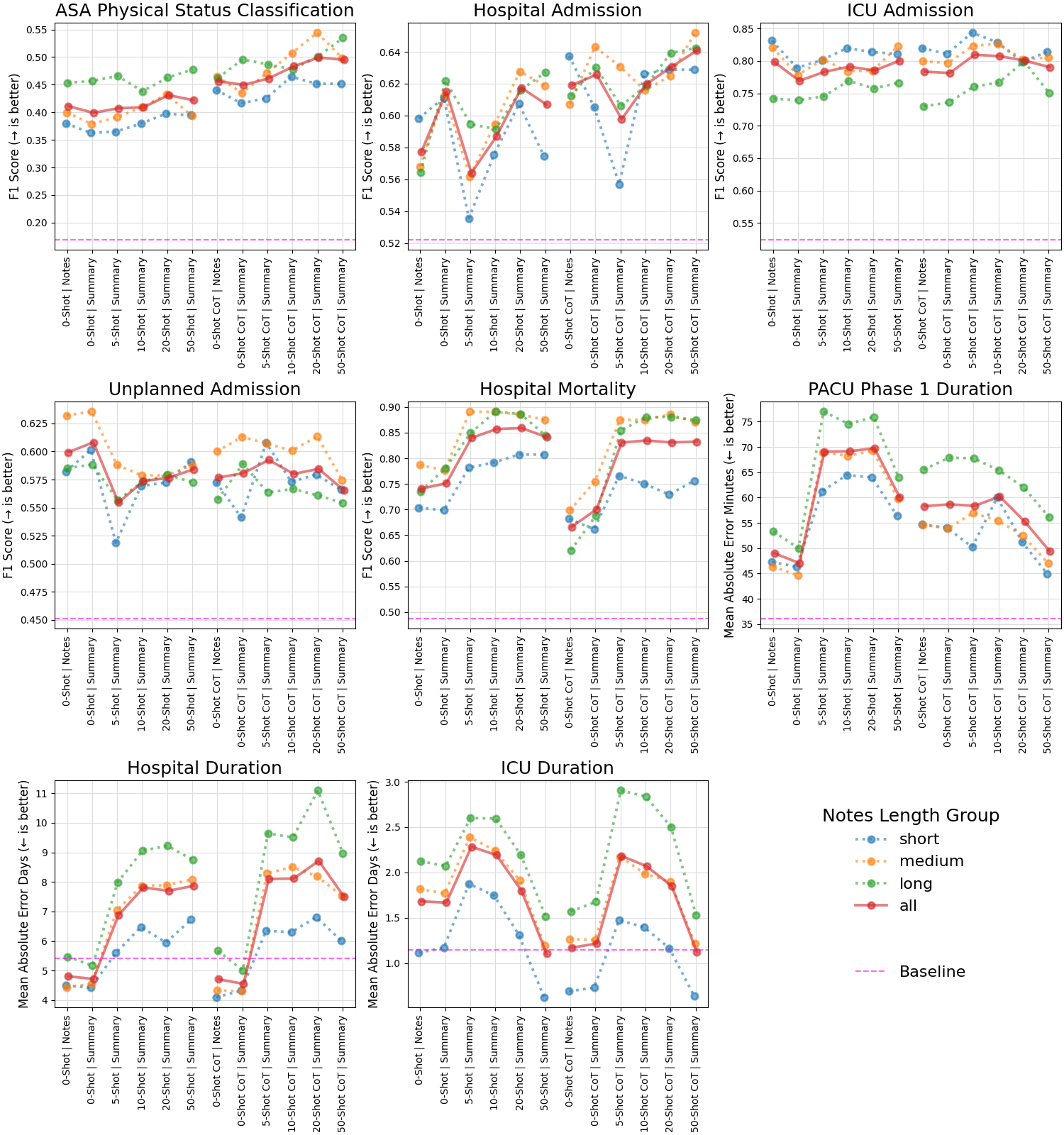}
    \caption{Performance for the 8 perioperative prediction tasks.  X-axis shows the different prompt strategies with the first six without chain-of-thought reasoning and the second six with chain-of-thought reasoning. “Notes” indicates that original clinical notes were inserted into the prompt whereas “Summary” indicates that clinical notes were first summarized using GPT-4 Turbo and then the summary was inserted into the prompt. All in-context examples for few-shot prompts used note summaries. Y-axis is F1 Score for classification tasks where higher score is better, and Mean Absolute Error for regression tasks where lower error is better. Baseline for classification tasks represent score achieved by random guessing. Baseline for regression tasks represent the MAE achieved by a regressor that always predicts the mean value in the dataset. The clinical notes are stratified into short, medium, and long length groups which represent the $\frac{1}{3}$ shortest, $\frac{1}{3}$ middle, and $\frac{1}{3}$ longest notes in the dataset and performance is shown for each stratification.}
    \label{fig:fig2-taskperformance}
\end{figure}

\subsection{Effect of Prompt Strategy on Perioperative Risk Prediction Tasks}\label{results-prompt_strategy}
Performance of each prompt strategy for each of the perioperative risk prediction tasks is summarized in Figure \ref{fig:fig2-taskperformance}. A detailed breakdown of performance metrics with confidence intervals is shown in Supplemental Tables \ref{tab:supptable3-asa_f1}-\ref{tab:supptable10-icuduration_mae}. This includes F1, Matthew’s Correlation Coefficient (MCC), Sensitivity, Specificity, PPV, NPV for classification tasks, and MAE for regression tasks. The difference between nearly all prompt strategies is statistically significant (Supplemental Figures \ref{fig:suppfig4-asa_pvalues}-\ref{fig:suppfig11-icuduration_pvalues}). GPT-4 Turbo provided a valid answer to all prompts except for 1 case in the unplanned admission dataset, which was excluded from our evaluation metrics. During dataset creation, most clinical notes written for administrative purposes such as attending attestations or for billing purposes were found to be less than 100 tokens in length and were removed. However, some of these notes were still present in the final dataset and in cases where the sole note did not contain clinically-relevant content as in the case for the unplanned admission dataset, GPT-4 Turbo correctly identified this situation and refused to make a prediction across all 5 allotted retries.

\begin{figure}[b!]
    \centering
    \includegraphics[width=1\linewidth]{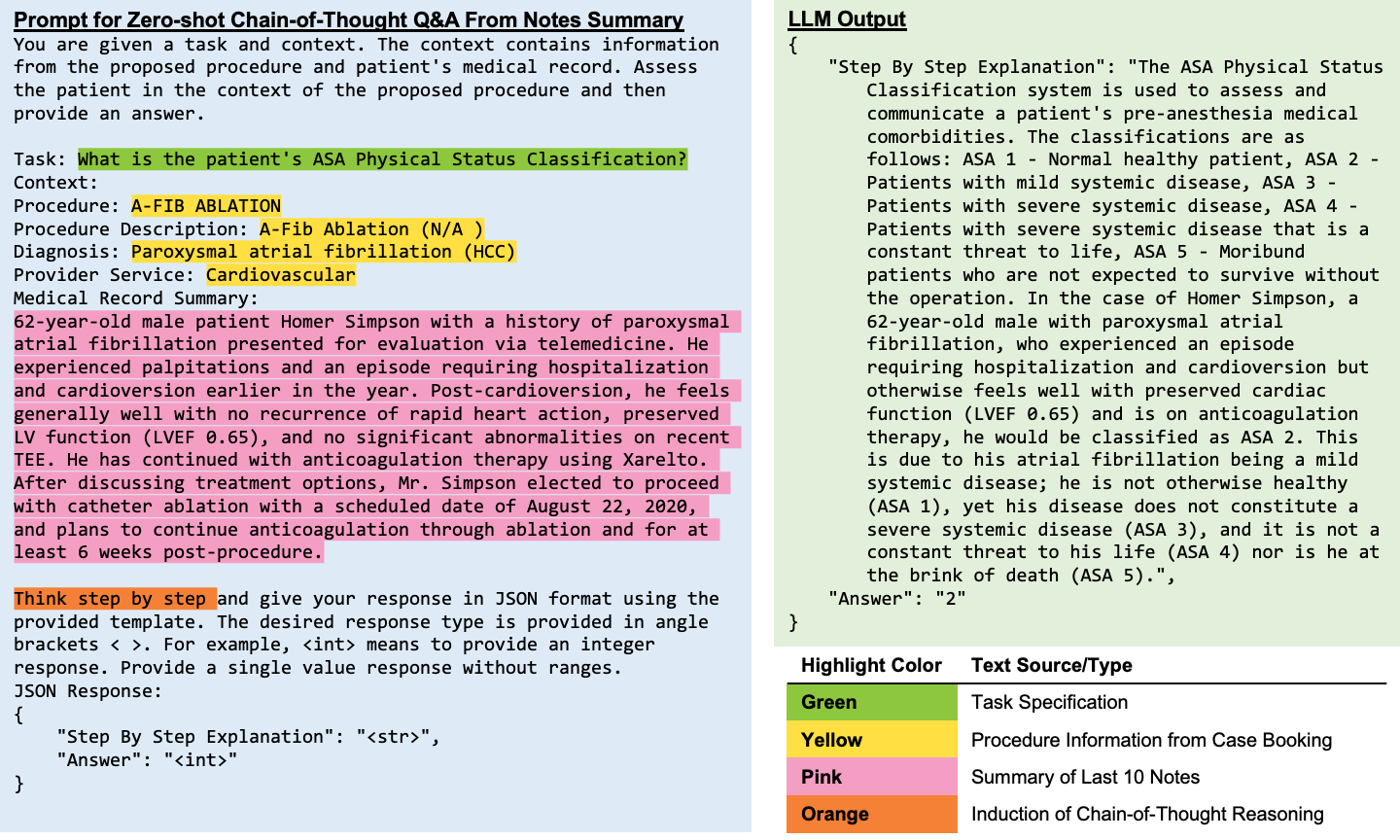}
    \caption{Prompt and LLM Output for Zero-shot chain-of-thought Q\&A from notes summary prompt strategy. Note summaries are generated from raw clinical notes from the LLM prior to insertion into the prompt. The LLM output shows that the LLM understands the definition for ASA Physical Status Classification (ASA-PS) and provides a valid rationale for which ASA-PS class the patient should be classified. All prompt strategies using this patient and procedure example are depicted in Supplemental Figure 1. While the content of this example is derived from a real patient and case from the electronic health record, all PHI and PII are removed with names obfuscated, and dates and times shifted.}
    \label{fig:fig3-prompt}
\end{figure}

ASA-PS, hospital admission, ICU admission, unplanned admission, and hospital mortality are classification tasks with performance measured using F1 score where a higher score corresponds to better performance. The baseline presented in each plot in Figure 2 for these outcomes represents the score for random guessing. All prompt strategies outperformed the random baseline for classification tasks. CoT resulted in significant performance gains for predicting ASA-PS and hospital admission, but not ICU admission, unplanned admission, and hospital mortality. Figure \ref{fig:fig3-prompt} depicts a prompt and LLM output for ASA-PS prediction using CoT, illustrating the LLM’s rich understanding of the task in the context of the patient and procedure. The logical reasoning is easily understood by any clinician and can rapidly assist a clinician with perioperative evaluation of the patient. Prompts and LLM outputs for all prompt strategies are shown in Supplemental Figure \ref{suppfig1-prompts} to illustrate how each prompt strategy affects the LLM output. Few-shot prompting showed consistent performance gains with ASA-PS, hospital admission, and hospital mortality. Gains for CoT and few-shot prompting were synergistic for ASA-PS and hospital admission, but for hospital mortality, CoT resulted in slightly worse performance. ICU admission performance was not significantly affected by the prompt strategies, but all prompt strategies had high prediction performance, suggesting the LLM is easily able to perform this task. Zero-shot without CoT was best performing for unplanned admission and demonstrated that rationalizing does not help with all prediction tasks.

PACU phase 1 duration, hospital duration, and ICU duration are regression tasks with performance measured using MAE where a lower score corresponds to better performance. The baseline is the MAE for a dummy regressor that always predicts the mean duration in the dataset. Most prompt strategies performed worse than this baseline. For PACU phase 1 duration prediction, all prompt strategies performed worse than the dummy regressor, suggesting that the LLM struggled with this task. For hospital duration prediction, zero-shot and zero-shot CoT slightly outperformed the baseline dummy regressor with an error of 4.5 days compared to 5.4 days, which is not a clinically significant difference. Addition of few-shot prompting worsened hospital duration prediction, despite few-shot and CoT improving hospital admission classification, which is the analogous binary prediction task. For ICU duration prediction, few-shot and CoT did help with predictions, but were only able to achieve parity with the baseline dummy regressor. This is in contrast to the analogous binary prediction task, ICU admission prediction, for which the LLM far exceeded its baseline.

The best performing prompt strategies for each prediction task were:

\begin{enumerate}
    \item \textbf{ASA-PS}: 20-shot CoT and 50-shot CoT, both which achieved F1 of 0.50.
    \item \textbf{Hospital admission}: 50-shot CoT with F1 of 0.64.
    \item \textbf{ICU admission}: 5-shot CoT prompt strategy with F1 score of 0.81.
    \item \textbf{Unplanned admission}: Zero-shot using note summaries with F1 score of 0.61.
    \item \textbf{Hospital mortality}: 10-shot and 20-shot with F1 score of 0.86.
    \item \textbf{PACU phase 1 duration}: Zero-shot using original notes with MAE of 49 minutes.
    \item \textbf{Hospital duration}: Zero-shot CoT using notes summary with MAE of 4.5 days.
    \item \textbf{ICU duration}: Both 50-shot and 50-shot CoT with MAE of 1.1 days.
\end{enumerate}

\subsection{Effect of Summary Representation of Notes}\label{results-note_summary}
Prior work has shown that LLM-generated summaries in the clinical domain may be preferable to human-written summaries\cite{Van_Veen2023-sr}. Comparison of zero-shot prompts using original notes versus zero-shot prompts using LLM generated summaries resulted in slight degradation of performance as seen in ASA-PS, ICU admission, PACU phase 1 duration, and hospital duration, but also resulted in a boost in performance for hospital admission, unplanned admission, and hospital mortality prediction. The magnitude of these effects were small indicating that while summaries sometimes resulted in slight loss of information useful for the predictive task, summaries may also help focus relevant information. However, a distinct advantage in using a summary representation of patient history was the ability to scale to a large number of in-context examples when using few-shot prompting, such as the 50-shot prompt strategies. In several tasks, this resulted in significantly better predictive performance, but summary representations of patient histories were necessary to compress patient notes so they fit within the input context of the LLM.

\subsection{Effect of Note Length on Perioperative Risk Prediction Tasks}\label{results-note_length}
Note length had a differential effect on several tasks including better performance for ASA-PS prediction and hospital mortality prediction. Since up to the last 10 clinical notes were used in the input to LLM, increased note length was due to either longer notes or more notes being written about the patient. However, for ICU admission prediction, PACU phase 1 duration prediction, and hospital duration prediction, longer input note lengths resulted in worse prediction performance. 

\subsection{Numerical Prediction Tasks}\label{results-numerical_prediction}
\begin{figure}[b!]
    \centering
    \includegraphics[width=1\linewidth]{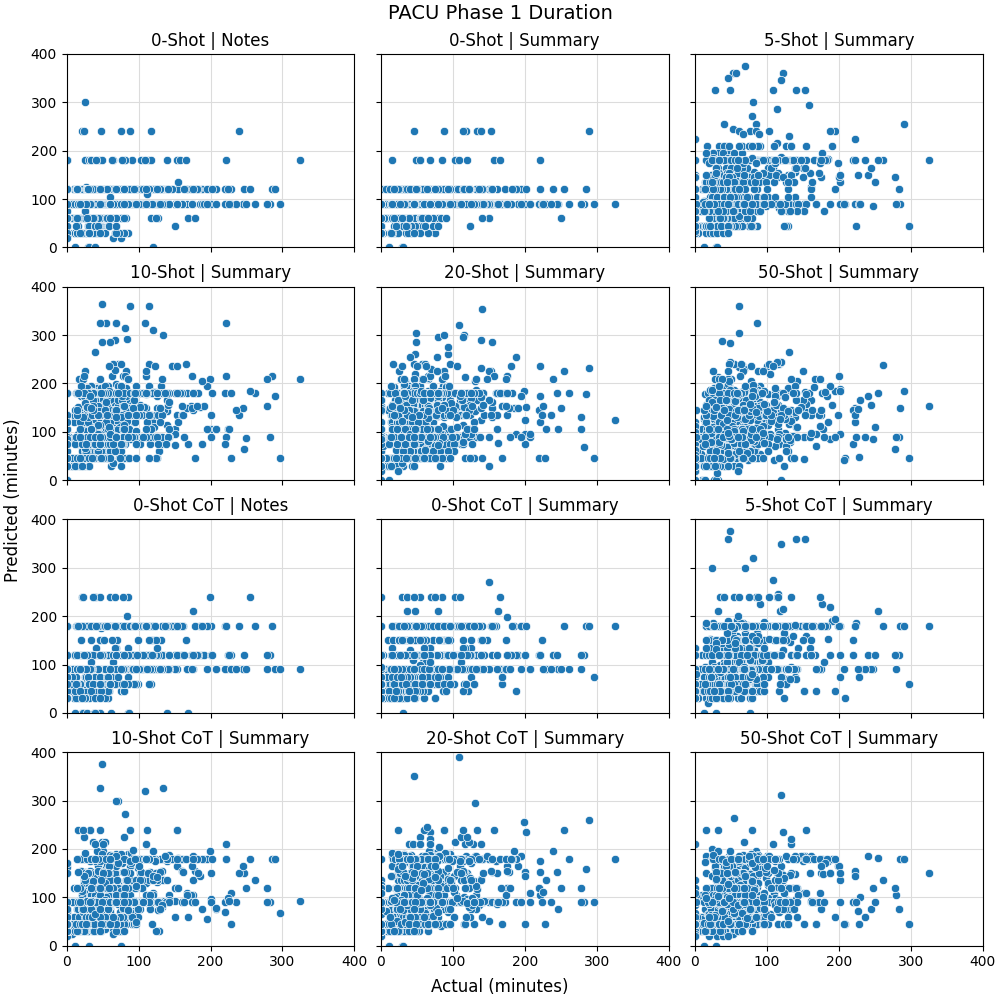}
    \caption{Scatter plot of predicted and actual post-anesthesia care unit (PACU) Phase 1 recovery durations across all 12 prompt strategies. Without few-shot and CoT prompting, predictions are heavily quantized to specific values and exhibit a ceiling effect where the LLM rarely predicts beyond 180 minutes. The progressive addition of few-shot and CoT prompting removes this effect, but predictive performance remains poor.}
    \label{fig:fig4-phase1duration}
\end{figure}

Performance on numerical predictions (PACU phase 1 duration, hospital duration, icu duration) was poor. LLM predictions were often worse than simply guessing the mean value of the dataset. However, when the same task was recast as a binary prediction as in the case of hospital admission or ICU admission prediction, LLM prediction performance was significantly better. Visualizing PACU phase 1 duration predictions in Figure \ref{fig:fig4-phase1duration} revealed that without few-shot and CoT prompting, LLMs tend to predict quantized outputs, often with a ceiling effect. Few-shot and CoT prompting helps remove these effects, but results in worse performance. Similar findings were seen for hospital duration prediction (Supplemental Figure \ref{fig:suppfig18-hospitalduration_scatter}) and ICU duration prediction (Supplemental Figure \ref{fig:suppfig19-icuduration_scatter}).

\section{Discussion}\label{discussion}

Our investigations indicate that general-domain LLMs such as GPT-4 Turbo can be applied to clinical notes and procedure description to achieve perioperative risk assessment and prognostication. As Nori et al.\cite{Nori2023-vf} has observed, the addition of few-shot prompting and chain-of-thought prompting helps boost performance, though not unilaterally across all tasks.

We observe strong overall performance for prediction of postoperative ICU admission and hospital mortality across all prompt strategies. ICU admission prediction exhibits high sensitivity \(\ge0.78\) and specificity \(\ge0.74\) across all prompt strategies (Supplemental Table). Hospital mortality prediction can be enhanced using few-shot prompting techniques and exhibits high sensitivity \(\ge0.93\) across all prompt strategies with highest specificity of 0.81 from 20-shot prompt strategy (Supplemental Table \ref{tab:supptable7-hospitalmortality-tpr_npr_ppv_npv}). We also observe good predictive performance for ASA-PS. ASA-PS assignment is known to be subjective and has only moderate interrater agreement among human anesthesiologists\cite{Cuvillon2011-rm, Sankar2014-fn}, so it is unlikely that any prediction system can achieve a perfect score. In this context, a multiclass ASA-PS F1 micro score of 0.5 has meaningful clinical utility (Supplemental Table \ref{tab:supptable3-asa_f1}); confusion matrices also show that ASA-PS misclassifications made by the LLM with few-shot and CoT are almost always an adjacent ASA class (Supplemental Figure \ref{fig:suppfig12-asa_confusionmatrix}). Hospital admission prediction and unplanned admission prediction performance is better than random guessing, but not as impressive as outcome measures like ICU admission and hospital mortality where the illness severity of a patient is likely more apparent and it is easier for the LLM to make prediction. Still, it is remarkable that GPT-4 Turbo can achieve this kind of predictive performance from only procedure description and clinical notes with no specialized clinical training and no fine-tuning for perioperative risk prediction tasks.

Few-shot and chain-of-thought prompting reveal significant gains in predictive performance in tasks where synthesizing prior clinical knowledge is important, such as determination of ASA-PS, hospital admission prediction, and hospital mortality. These effects are additive and synergistic but the benefits of these prompting techniques do not apply to all outcomes. We believe the aforementioned prediction tasks benefit from the prompting strategies because they are heavily dependent on preoperative illness severity, which would be reflected in a patient’s clinical notes. In-context examples will help the LLM compare and contrast among similar cases whereas CoT rationales will help expand upon the concepts mentioned in clinical notes, both of which will guide the LLM toward more accurate predictions. In contrast, we suspect that these gains are not seen in outcomes such as unplanned admission because factors leading to unexpected admission are predominantly due to intraoperative factors which are not accounted for in pre-operative clinical notes and procedure booking data presented to the LLM, and no amount of deliberation or rationalization would affect the outcome. 

In their current incarnation, LLMs struggle with regression tasks involving prediction of continuous or numerical outcomes such as PACU phase 1 duration, hospital duration, and ICU duration. Our analysis of duration prediction tasks shows that LLMs tend to quantize values, which we suspect is due to the LLM memorizing length of stay estimations from hospital websites, textbooks, and journal articles (Figure \ref{fig:fig4-phase1duration}, Supplemental Figure \ref{fig:suppfig18-hospitalduration_scatter}). In-context demonstrations and prompting the LLM to rationalize about the patient’s procedure and medical history helps overcome this quantization phenomenon, but we believe the continued poor results to be attributed to the architectural design of LLMs. Namely, LLMs enforce a discrete tokenized output where each token’s representation is primarily derived from text contexts. For continuous-valued outcomes, it is meaningful to be able to interpolate between numerical values but a LLM’s training data and training process does not provide a robust way for the model to learn this concept. Potential strategies to overcome this limitation include multimodal enhancements to LLMs to treat numbers as distinct data modalities and directly mapping of continuous values to and from the embedding space of neural network layers\cite{Driess2023-sb, Belyaeva2023-bi, Xu2023-qu, Tu2023-lq}. While many visual-language\cite{Alayrac2022-hn, Moor2023-gx, Chen2022-ea} and multimodal models adopt these strategies to combine text and other data modalities such as pixel intensity in the same model, no widely available model has yet employed these solutions for general numerical predictions. Future foundational models for healthcare or EHR data should consider model architectures and pretraining routines that enable better performance for these kind of numerical prediction tasks. Another alternative is equipping LLMs with tool use\cite{Schick2023-yl, Qin2023-rl, Cai2023-ya, Goodell2023-wc}, but this relegates the LLM as a natural language information extractor and outsources the actual prediction task to an external model rather than taking full advantage of the LLM’s capability for information understanding and synthesis to make a prediction.

Our studies in stratifying prediction performance on note length groups and comparing zero-shot predictions from original notes versus LLM-generated summaries indicate that longer input contexts do not necessarily result in better performance, though the specific effects are task-specific. This is contrary to the intuition that providing the LLM more clinical context and information would enable a more accurate prediction. We believe that this is because increased note length from the EHR may be the result of automatically copied templates that do not substantially add knowledge over a carefully crafted clinical vignette statement. Conversely, we qualitatively find LLM-generated summaries to be high-quality (Figure \ref{fig:fig3-prompt}, Supplemental Figure \ref{suppfig1-prompts}) which concords with prior studies showing LLMs to have excellent clinical summarization capabilities. Furthermore, prior work has shown that many current LLMs are biased to pay more attention to the beginning and end of prompts, which may be an artifact of training data not containing long-context prediction or retrieval tasks\cite{Liu2023-je}. This would explain why some tasks perform better when note summaries are used instead of original notes. Similar to Nori et al.\cite{Nori2023-vf, Nori2023-xz}, we observe GPT-4 Turbo generates high-quality CoT rationalizations and answer explanations. We also find the LLM appropriately refuses to answer when the context is irrelevant to the task, which is a desirable property in clinical use cases.

Overall, our results indicate that currently available general-domain LLMs can have direct clinical impact to perioperative risk stratification workflows in hospitals and can be used to assist in stratifying the preoperative patient population for these outcomes. In particular, we find LLMs to exhibit very good performance at ASA Physical Status Classification prediction, ICU admission prediction, and Hospital Mortality prediction. LLMs underperform dedicated classification models utilizing tabular features\cite{Mudumbai2019-op, Graesner2023-ik, Lee2022-ii, Hill2019-uz, Bilimoria2013-fy, Chen2022-gc, Xu2023-nf, Meguid2016-bn, Tully2023-qg, Fang2023-kw, Gabriel2017-wn, Dyas2022-au, Le_Manach2016-dc, Smilowitz2020-qh} and if standalone automated triage and prediction is the goal, it is unlikely current generation LLMs will perform better. However traditional machine learning learning models are rarely utilized in the clinical setting because of difficulty in interpreting a model’s predictions. LLMs are unique in their ability to present natural language explanations understandable to human clinicians, and thus have significant clinical utility in summarizing the patient history and creating rationalizations against each outcome variable of interest. These explanations are valuable starting points for clinicians to perform a more comprehensive perioperative risk assessment, and are more useful clinically than standalone risk predictions.

Future research is needed to evaluate whether clinical domain-specific language models\cite{Chen2023-jw, Singhal2023-kc, Singhal2023-yl, Toma2023-uf}, trained on clinical or EHR text shows improved performance or more advanced few-shot prompting strategies such as dynamic kNN few-shot, shuffling few-shot examples, ensembling, or retrieval-augmentation\cite{Nori2023-vf, Wang2022-ol, Lewis2020-as, Zakka2023-tz, Wang2023-jn}.  Our investigations are centered around whether LLMs can perform these tasks and use outcome-balanced datasets to enable measurement of these rare outcomes. Real-world incidence of these outcomes is rare and of the 137535 cases considered in the 2 year span from which our datasets are derived, only 0.49\% of cases have associated postoperative ICU admissions, 0.43\% of cases had an unanticipated hospital admission, and 0.3\% of cases have associated postoperative in-hospital mortality (Figure \ref{fig:suppfig2-flowdiagram}. Future large-scale prospective clinical validation is necessary to verify our observed performance, especially for rare outcomes, and to compare against existing perioperative prediction algorithms\cite{Le_Manach2016-dc,Smilowitz2020-qh}.

\section{Conclusion}\label{conclusion}

General-domain text-only large language models are capable of perioperative risk prediction and prognostication when framed as classification or binary prediction tasks, but still fall short in being able to predict continuous-valued outcomes such as PACU, hospital, and ICU length of stay. Few-shot prompting and chain-of-thought reasoning improves prediction performance for perioperative prediction tasks. Hospitals should evaluate the effectiveness of using large language models as tools to assist perioperative risk stratification.

\section*{Declarations}

\bmhead{Acknowledgments}

The authors would like to acknowledge: University of Washington Anesthesia Department’s Perioperative \& Pain initiatives in Quality Safety Outcome group for assistance on data extraction and discussions in dataset and experimental design, University of Washington Department of Medicine for computational environment support, Roland Lai and Robert Fabiano from University of Washington Research IT for creating a digital research environment within the Microsoft Azure Cloud where experiments were performed, the University of Washington Biomedical Natural Language Processing group, and the Aghaeepour laboratory at Stanford University for providing early feedback on experimental design and results.

\bmhead{Funding}

Computational resources for this project were funded by the Microsoft Azure Cloud Compute Credits grant program from the University of Washington eScience Institute and Microsoft Azure. Financial support for this work was provided by the University of Washington Department of Anesthesiology \& Pain Medicine's Bonica Scholars Program, Stanford University Research in Anesthesia Training Program (ReAP) program (NIH 5T32GM089626), and NIH R35GM138353. Funding sources were not involved in design of the study, collection of data, analysis of data, interpretation of data or in writing the manuscript.

\bmhead{Contributions}
P.C. and V.O.R conceived of the initial research question. A.M.W. and M.Y. helped refine the research question, prompting strategies, and experimental approach. P.C. and C.T.F. refined inclusion and exclusion criteria and conducted data extraction, cleaning, and transformation into final datasets. P.C. wrote code and conducted experiments, analyzed data and wrote the manuscript. N.A., A.M.W., M.Y., and V.O.R provided feedback on initial results and refinement to experimental methodology and analysis. All authors reviewed and revised the paper.

\bmhead{Competing Interests}
None of the authors have any financial or non-financial competing interests.

\bmhead{Data Availability}
The raw data and datasets generated in our experiments are not publicly available because they are derived from electronic health records consisting of protected health information (PHI). Data may be requested by contacting Vikas O’Reilly-Shah at voreill@uw.edu or the University of Washington Center for Perioperative \& Pain Initiatives in Quality Safety Outcome (PPiQSO) at PPiQSO@uw.edu. Data access is contingent upon signing a data use agreement in accordance with UW Medicine policy. 

\bmhead{Code Availability}
Code for experiments and analysis results are publicly available at:\\ \url{https://github.com/philipchung/llm-periop-prediction}

\newpage
\section*{Tables}\label{tables}
\fontsize{7}{9}\selectfont
\subsection*{Table 1: Dataset Characteristics}
\begin{longtable}[c]{@{}lllll@{}}
\caption{Description of patient demographics, procedure specialty, patient class, note statistics, and outcome variables for each dataset: (A) American Society of Anesthesiologists Physical Status (ASA-PS) Classification, (B) Post-Anesthesia Care Unit (PACU) Phase 1 Duration, (C) Hospital Admission, (D) Hospital Duration, (E) ICU Admission \& ICU Duration, (F) Unplanned Admission, (G) Hospital Mortality.}
\label{tab:table1}\\
\toprule
\textbf{A.} &
   &
   &
  \multicolumn{2}{l}{\textbf{ASA Physical Status Dataset}} \\* \midrule
\endfirsthead
\multicolumn{5}{c}%
{{\bfseries Table 1-A continued from previous page}} \\
\toprule
\textbf{A.} &
   &
   &
  \multicolumn{2}{l}{\textbf{ASA Physical Status Dataset}} \\* \midrule
\endhead
 &
   &
   &
  \textbf{Inference} &
  \textbf{Fewshot} \\* \midrule
\multirow{5}{*}{\textbf{\begin{tabular}[c]{@{}l@{}}Patient\\ Info\end{tabular}}} &
  \multirow{2}{*}{\textbf{Demographics}} &
  \textbf{\begin{tabular}[c]{@{}l@{}}Case Counts, \\ no. (\%)\end{tabular}} &
  \multicolumn{1}{r}{1000} &
  \multicolumn{1}{r}{250} \\* \cmidrule(l){3-5} 
 &
   &
  \textbf{Age, mean (std)} &
  51.9 (19.9) &
  51.0 (21.6) \\* \cmidrule(l){2-5} 
 &
  \multirow{3}{*}{\textbf{Sex, no. (\%)}} &
  \textbf{Female} &
  447 (44.7\%) &
  108 (43.2\%) \\* \cmidrule(l){3-5} 
 &
   &
  \textbf{Male} &
  553 (55.3\%) &
  142 (56.8\%) \\* \cmidrule(l){3-5} 
 &
   &
  \textbf{\begin{tabular}[c]{@{}l@{}}Other or \\ Unknown\end{tabular}} &
  -- &
  -- \\* \midrule
\multirow{37}{*}{\textbf{\begin{tabular}[c]{@{}l@{}}Case\\ Info\end{tabular}}} &
  \multirow{3}{*}{\textbf{\begin{tabular}[c]{@{}l@{}}Anesthesia Type, \\ no. (\%)\end{tabular}}} &
  \textbf{General} &
  842 (84.2\%) &
  216 (86.4\%) \\* \cmidrule(l){3-5} 
 &
   &
  \textbf{MAC} &
  125 (12.5\%) &
  22 (8.8\%) \\
\bottomrule
\end{longtable}

\newpage
\begin{longtable}[c]{@{}lllll@{}}
\toprule
\textbf{B.} &
  \multicolumn{1}{c}{} &
   &
  \multicolumn{2}{l}{\textbf{PACU Phase 1 Duration Dataset}} \\* \midrule
\endfirsthead
\multicolumn{5}{c}%
{{\bfseries Table 1-B continued from previous page}} \\
\toprule
\textbf{B.} &
  \multicolumn{1}{c}{} &
   &
  \multicolumn{2}{l}{\textbf{PACU Phase 1 Duration Dataset}} \\* \midrule
\endhead
 &
   &
   &
  \textbf{Inference} &
  \textbf{Fewshot} \\* \midrule
\multirow{5}{*}{\textbf{\begin{tabular}[c]{@{}l@{}}Patient\\ Info\end{tabular}}} &
  \multirow{2}{*}{\textbf{Demographics}} &
  \textbf{\begin{tabular}[c]{@{}l@{}}Case Counts, \\ no. (\%)\end{tabular}} &
  \multicolumn{1}{r}{1000} &
  \multicolumn{1}{r}{250} \\* \cmidrule(l){3-5} 
 &
   &
  \textbf{Age, mean (std)} &
  50.0 (19.5) &
  50.5 (19.1) \\* \cmidrule(l){2-5} 
 &
  \multirow{3}{*}{\textbf{Sex, no. (\%)}} &
  \textbf{Female} &
  494 (49.4\%) &
  141 (56.4\%) \\* \cmidrule(l){3-5} 
 &
   &
  \textbf{Male} &
  506 (50.6\%) &
  109 (43.6\%) \\* \cmidrule(l){3-5} 
 &
   &
  \textbf{\begin{tabular}[c]{@{}l@{}}Other or \\ Unknown\end{tabular}} &
  -- &
  -- \\* \midrule
\multirow{37}{*}{\textbf{\begin{tabular}[c]{@{}l@{}}Case\\ Info\end{tabular}}} &
  \multirow{3}{*}{\textbf{\begin{tabular}[c]{@{}l@{}}Anesthesia Type, \\ no. (\%)\end{tabular}}} &
  \textbf{General} &
  867 (86.7\%) &
  209 (83.6\%) \\* \cmidrule(l){3-5} 
 &
   &
  \textbf{MAC} &
  69 (6.9\%) &
  18 (7.2\%) \\
\bottomrule
\end{longtable}

\newpage
\begin{longtable}[c]{@{}lllll@{}}
\toprule
\textbf{C.} &
  \multicolumn{1}{c}{} &
   &
  \multicolumn{2}{l}{\textbf{Hospital Admission Dataset}} \\* \midrule
\endfirsthead
\multicolumn{5}{c}%
{{\bfseries Table 1-C continued from previous page}} \\
\toprule
\textbf{C.} &
  \multicolumn{1}{c}{} &
   &
  \multicolumn{2}{l}{\textbf{Hospital Admission Dataset}} \\* \midrule
\endhead
 &
   &
   &
  \textbf{Inference} &
  \textbf{Fewshot} \\* \midrule
\multirow{5}{*}{\textbf{\begin{tabular}[c]{@{}l@{}}Patient\\ Info\end{tabular}}} &
  \multirow{2}{*}{\textbf{Demographics}} &
  \textbf{\begin{tabular}[c]{@{}l@{}}Case Counts, \\ no. (\%)\end{tabular}} &
  \multicolumn{1}{r}{1000} &
  \multicolumn{1}{r}{250} \\* \cmidrule(l){3-5} 
 &
   &
  \textbf{Age, mean (std)} &
  50.5 (19.3) &
  49.3 (20.8) \\* \cmidrule(l){2-5} 
 &
  \multirow{3}{*}{\textbf{Sex, no. (\%)}} &
  \textbf{Female} &
  439 (43.9\%) &
  116 (46.4\%) \\* \cmidrule(l){3-5} 
 &
   &
  \textbf{Male} &
  560 (56.0\%) &
  134 (53.6\%) \\* \cmidrule(l){3-5} 
 &
   &
  \textbf{\begin{tabular}[c]{@{}l@{}}Other or \\ Unknown\end{tabular}} &
  1 (0.1\%) &
  -- \\* \midrule
\multirow{37}{*}{\textbf{\begin{tabular}[c]{@{}l@{}}Case\\ Info\end{tabular}}} &
  \multirow{3}{*}{\textbf{\begin{tabular}[c]{@{}l@{}}Anesthesia Type, \\ no. (\%)\end{tabular}}} &
  \textbf{General} &
  826 (82.6\%) &
  210 (84.0\%) \\* \cmidrule(l){3-5} 
 &
   &
  \textbf{MAC} &
  106 (10.6\%) &
  24 (9.6\%) \\
\bottomrule
\end{longtable}

\newpage
\begin{longtable}[c]{@{}lllll@{}}
\toprule
\textbf{D.} &
  \multicolumn{1}{c}{} &
   &
  \multicolumn{2}{l}{\textbf{Hospital Duration Dataset}} \\* \midrule
\endfirsthead
\multicolumn{5}{c}%
{{\bfseries Table 1-D continued from previous page}} \\
\toprule
\textbf{D.} &
  \multicolumn{1}{c}{} &
   &
  \multicolumn{2}{l}{\textbf{Hospital Duration Dataset}} \\* \midrule
\endhead
 &
   &
   &
  \textbf{Inference} &
  \textbf{Fewshot} \\* \midrule
\multirow{5}{*}{\textbf{\begin{tabular}[c]{@{}l@{}}Patient\\ Info\end{tabular}}} &
  \multirow{2}{*}{\textbf{Demographics}} &
  \textbf{\begin{tabular}[c]{@{}l@{}}Case Counts, \\ no. (\%)\end{tabular}} &
  \multicolumn{1}{r}{1000} &
  \multicolumn{1}{r}{250} \\* \cmidrule(l){3-5} 
 &
   &
  \textbf{Age, mean (std)} &
  56.1 (19.1) &
  57.9 (18.0) \\* \cmidrule(l){2-5} 
 &
  \multirow{3}{*}{\textbf{Sex, no. (\%)}} &
  \textbf{Female} &
  456 (45.6\%) &
  131 (52.4\%) \\* \cmidrule(l){3-5} 
 &
   &
  \textbf{Male} &
  544 (54.4\%) &
  119 (47.6\%) \\* \cmidrule(l){3-5} 
 &
   &
  \textbf{\begin{tabular}[c]{@{}l@{}}Other or \\ Unknown\end{tabular}} &
  -- &
  -- \\* \midrule
\multirow{37}{*}{\textbf{\begin{tabular}[c]{@{}l@{}}Case\\ Info\end{tabular}}} &
  \multirow{3}{*}{\textbf{\begin{tabular}[c]{@{}l@{}}Anesthesia Type, \\ no. (\%)\end{tabular}}} &
  \textbf{General} &
  901 (90.1\%) &
  229 (91.6\%) \\* \cmidrule(l){3-5} 
 &
   &
  \textbf{MAC} &
  52 (5.2\%) &
  11 (4.4\%) \\
\bottomrule
\end{longtable}

\newpage
\begin{longtable}[c]{@{}lllll@{}}
\toprule
\textbf{E.} &
  \multicolumn{1}{c}{} &
   &
  \multicolumn{2}{l}{\textbf{ICU Admission \& ICU Duration Dataset}} \\* \midrule
\endfirsthead
\multicolumn{5}{c}%
{{\bfseries Table 1-E continued from previous page}} \\
\toprule
\textbf{E.} &
  \multicolumn{1}{c}{} &
   &
  \multicolumn{2}{l}{\textbf{ICU Admission \& ICU Duration Dataset}} \\* \midrule
\endhead
 &
   &
   &
  \textbf{Inference} &
  \textbf{Fewshot} \\* \midrule
\multirow{5}{*}{\textbf{\begin{tabular}[c]{@{}l@{}}Patient\\ Info\end{tabular}}} &
  \multirow{2}{*}{\textbf{Demographics}} &
  \textbf{\begin{tabular}[c]{@{}l@{}}Case Counts, \\ no. (\%)\end{tabular}} &
  \multicolumn{1}{r}{1000} &
  \multicolumn{1}{r}{250} \\* \cmidrule(l){3-5} 
 &
   &
  \textbf{Age, mean (std)} &
  53.5 (18.8) &
  55.8 (18.5) \\* \cmidrule(l){2-5} 
 &
  \multirow{3}{*}{\textbf{Sex, no. (\%)}} &
  \textbf{Female} &
  492 (49.2\%) &
  128 (51.2\%) \\* \cmidrule(l){3-5} 
 &
   &
  \textbf{Male} &
  507 (50.7\%) &
  122 (48.8\%) \\* \cmidrule(l){3-5} 
 &
   &
  \textbf{\begin{tabular}[c]{@{}l@{}}Other or \\ Unknown\end{tabular}} &
  1 (0.1\%) &
  -- \\* \midrule
\multirow{37}{*}{\textbf{\begin{tabular}[c]{@{}l@{}}Case\\ Info\end{tabular}}} &
  \multirow{3}{*}{\textbf{\begin{tabular}[c]{@{}l@{}}Anesthesia Type, \\ no. (\%)\end{tabular}}} &
  \textbf{General} &
  862 (86.2\%) &
  220 (88.0\%) \\* \cmidrule(l){3-5} 
 &
   &
  \textbf{MAC} &
  109 (10.9\%) &
  26 (10.4\%) \\
\bottomrule
\end{longtable}

\newpage
\begin{longtable}[c]{@{}lllll@{}}
\toprule
\textbf{F.} &
  \multicolumn{1}{c}{} &
   &
  \multicolumn{2}{l}{\textbf{Unplanned Admit Dataset}} \\* \midrule
\endfirsthead
\multicolumn{5}{c}%
{{\bfseries Table 1-F continued from previous page}} \\
\toprule
\textbf{F.} &
  \multicolumn{1}{c}{} &
   &
  \multicolumn{2}{l}{\textbf{Unplanned Admit Dataset}} \\* \midrule
\endhead
 &
   &
   &
  \textbf{Inference} &
  \textbf{Fewshot} \\* \midrule
\multirow{5}{*}{\textbf{\begin{tabular}[c]{@{}l@{}}Patient\\ Info\end{tabular}}} &
  \multirow{2}{*}{\textbf{Demographics}} &
  \textbf{\begin{tabular}[c]{@{}l@{}}Case Counts, \\ no. (\%)\end{tabular}} &
  \multicolumn{1}{r}{949} &
  \multicolumn{1}{r}{237} \\* \cmidrule(l){3-5} 
 &
   &
  \textbf{Age, mean (std)} &
  54.8 (18.6) &
  50.8 (19.2) \\* \cmidrule(l){2-5} 
 &
  \multirow{3}{*}{\textbf{Sex, no. (\%)}} &
  \textbf{Female} &
  471 (49.6\%) &
  130 (54.9\%) \\* \cmidrule(l){3-5} 
 &
   &
  \textbf{Male} &
  478 (50.4\%) &
  107 (45.1\%) \\* \cmidrule(l){3-5} 
 &
   &
  \textbf{\begin{tabular}[c]{@{}l@{}}Other or \\ Unknown\end{tabular}} &
  -- &
  -- \\* \midrule
\multirow{37}{*}{\textbf{\begin{tabular}[c]{@{}l@{}}Case\\ Info\end{tabular}}} &
  \multirow{3}{*}{\textbf{\begin{tabular}[c]{@{}l@{}}Anesthesia Type, \\ no. (\%)\end{tabular}}} &
  \textbf{General} &
  812 (85.6\%) &
  199 (84.0\%) \\* \cmidrule(l){3-5} 
 &
   &
  \textbf{MAC} &
  82 (8.6\%) &
  17 (7.2\%) \\
\bottomrule
\end{longtable}

\newpage
\begin{longtable}[c]{@{}lllll@{}}
\toprule
\textbf{G.} &
  \multicolumn{1}{c}{} &
   &
  \multicolumn{2}{l}{\textbf{Hospital Mortality Dataset}} \\* \midrule
\endfirsthead
\multicolumn{5}{c}%
{{\bfseries Table 1-G continued from previous page}} \\
\toprule
\textbf{G.} &
  \multicolumn{1}{c}{} &
   &
  \multicolumn{2}{l}{\textbf{Hospital Mortality Dataset}} \\* \midrule
\endhead
 &
   &
   &
  \textbf{Inference} &
  \textbf{Fewshot} \\* \midrule
\multirow{5}{*}{\textbf{\begin{tabular}[c]{@{}l@{}}Patient\\ Info\end{tabular}}} &
  \multirow{2}{*}{\textbf{Demographics}} &
  \textbf{\begin{tabular}[c]{@{}l@{}}Case Counts, \\ no. (\%)\end{tabular}} &
  \multicolumn{1}{r}{576} &
  \multicolumn{1}{r}{144} \\* \cmidrule(l){3-5} 
 &
   &
  \textbf{Age, mean (std)} &
  58.2 (18.9) &
  59.0 (19.7) \\* \cmidrule(l){2-5} 
 &
  \multirow{3}{*}{\textbf{Sex, no. (\%)}} &
  \textbf{Female} &
  240 (41.7\%) &
  67 (46.5\%) \\* \cmidrule(l){3-5} 
 &
   &
  \textbf{Male} &
  336 (58.3\%) &
  77 (53.5\%) \\* \cmidrule(l){3-5} 
 &
   &
  \textbf{\begin{tabular}[c]{@{}l@{}}Other or \\ Unknown\end{tabular}} &
  -- &
  -- \\* \midrule
\multirow{37}{*}{\textbf{\begin{tabular}[c]{@{}l@{}}Case\\ Info\end{tabular}}} &
  \multirow{3}{*}{\textbf{\begin{tabular}[c]{@{}l@{}}Anesthesia Type, \\ no. (\%)\end{tabular}}} &
  \textbf{General} &
  495 (85.9\%) &
  124 (86.1\%) \\* \cmidrule(l){3-5} 
 &
   &
  \textbf{MAC} &
  63 (10.9\%) &
  16 (11.1\%) \\
\bottomrule
\end{longtable}

\normalsize

\backmatter
\pagebreak
\begin{appendices}

\begin{landscape}
\section{Supplemental Tables}\label{supplemental-tables}

\subsection{Supplemental Table 1: Note Type \& Author Provider Type}\label{supptable1-section}
\fontsize{7}{9}\selectfont

\begin{longtable}[c]{@{}lllllllllll@{}}
\toprule
 & Dentist & Fellow & \begin{tabular}[c]{@{}l@{}}Nurse \\ Anesthetist\end{tabular} & \begin{tabular}[c]{@{}l@{}}Nurse \\ Practitioner\end{tabular} & Osteopath & Pharmacist & Physician & \begin{tabular}[c]{@{}l@{}}Physician \\ Assistant\end{tabular} & Podiatrist & Resident \\* \midrule
\endfirsthead
\multicolumn{11}{c}%
{{\bfseries Table \thetable\ continued from previous page}} \\
\toprule
 & Dentist & Fellow & \begin{tabular}[c]{@{}l@{}}Nurse \\ Anesthetist\end{tabular} & \begin{tabular}[c]{@{}l@{}}Nurse \\ Practitioner\end{tabular} & Osteopath & Pharmacist & Physician & \begin{tabular}[c]{@{}l@{}}Physician \\ Assistant\end{tabular} & Podiatrist & Resident \\* \midrule
\endhead
\begin{tabular}[c]{@{}l@{}}Anesthesia Preprocedure Evaluation\end{tabular} & -- & 47 & 179 & 54 & -- & -- & 3346 & 83 & -- & 418 \\* \midrule
\begin{tabular}[c]{@{}l@{}}Assessment \& Plan Note\end{tabular} & -- & 30 & -- & 21 & -- & -- & 305 & 4 & -- & 14 \\* \midrule
Brief Op Note & -- & 415 & -- & 4 & -- & -- & 535 & 8 & -- & 1014 \\* \midrule
\begin{tabular}[c]{@{}l@{}}Brief Procedure Note\end{tabular} & 1 & 280 & -- & 33 & -- & -- & 280 & 23 & -- & 464 \\* \midrule
Consults & 3 & 1056 & 1 & 360 & -- & 1 & 2601 & 666 & 23 & 5842 \\* \midrule
\begin{tabular}[c]{@{}l@{}}Discharge Summary\end{tabular} & -- & 13 & -- & 125 & -- & -- & 222 & 116 & -- & 277 \\* \midrule
\begin{tabular}[c]{@{}l@{}}ED Procedure Notes\end{tabular} & -- & 13 & -- & 20 & -- & -- & 521 & 12 & -- & 550 \\* \midrule
\begin{tabular}[c]{@{}l@{}}ED Provider Notes\end{tabular} & -- & 70 & -- & 402 & -- & -- & 3579 & 161 & -- & 3401 \\
\end{longtable}
\end{landscape}
\normalsize

\subsection{Supplemental Table 2: Experiment Costs}\label{supptable2-section}
\fontsize{7}{9}\selectfont
\begin{longtable}[c]{@{}lr@{}}
\toprule
\textbf{Task}                               & \multicolumn{1}{l}{\textbf{Cost}} \\* \midrule
\endfirsthead
\endhead
\bottomrule
\endfoot
\endlastfoot
\textbf{ASA Physical Status Classification} & \$765.50                          \\
\textbf{Hospital Admission}                 & \$743.02                          \\
\textbf{ICU Admission}                      & \$795.52                          \\
\textbf{Unplanned Admission}                & \$764.09                          \\
\textbf{Hospital Mortality}                 & \$468.70                          \\
\textbf{PACU Phase 1 Duration}              & \$831.08                          \\
\textbf{Hospital Duration}                  & \$834.44                          \\
\textbf{ICU Duration}                       & \$799.66                          \\
\textbf{TOTAL}                              & \$6,002.01                        \\* \bottomrule
\caption{Cost for API calls to GPT-4 Turbo (gpt-4-1106) using the Microsoft Azure OpenAI Service. This includes costs for both input prompt tokens as well as output LLM generated tokens.}
\label{tab:supptable2-experimentcosts}
\end{longtable}
\normalsize

\subsection{Supplemental Table 3: ASA Physical Status Classification - Metrics}\label{supptable3-section}
\fontsize{7}{9}\selectfont

\begin{longtable}[c]{@{}lllll@{}}
\toprule
\multicolumn{5}{c}{\textbf{ASA Physical Status Classification: F1 Score}}                                      \\* \midrule
\endfirsthead
\endhead
\bottomrule
\endfoot
\endlastfoot
\textbf{}                      & \textbf{All}      & \textbf{Long}     & \textbf{Medium}   & \textbf{Short}    \\* \midrule
\textbf{Baseline}              & 0.17 (0.15, 0.19) & --                & --                & --                \\
\textbf{0-Shot | Notes}        & 0.41 (0.38, 0.44) & 0.45 (0.40, 0.50) & 0.40 (0.35, 0.45) & 0.38 (0.33, 0.43) \\
\textbf{0-Shot | Summary}      & 0.40 (0.37, 0.43) & 0.46 (0.41, 0.51) & 0.38 (0.32, 0.43) & 0.36 (0.31, 0.42) \\
\textbf{5-Shot | Summary}      & 0.41 (0.38, 0.44) & 0.47 (0.41, 0.52) & 0.39 (0.34, 0.44) & 0.36 (0.31, 0.42) \\
\textbf{10-Shot | Summary}     & 0.41 (0.38, 0.44) & 0.44 (0.38, 0.49) & 0.41 (0.35, 0.46) & 0.38 (0.33, 0.43) \\
\textbf{20-Shot | Summary}     & 0.43 (0.40, 0.46) & 0.46 (0.41, 0.51) & 0.43 (0.38, 0.49) & 0.40 (0.34, 0.45) \\
\textbf{50-Shot | Summary}     & 0.42 (0.39, 0.45) & 0.48 (0.43, 0.53) & 0.39 (0.34, 0.44) & 0.39 (0.34, 0.45) \\
\end{longtable}

\begin{longtable}[c]{@{}lllll@{}}
\toprule
\multicolumn{5}{c}{\textbf{ASA Physical Status Classification: Matthew's Correlation Coefficient (MCC)}}        \\* \midrule
\endfirsthead
\endhead
\bottomrule
\endfoot
\endlastfoot
\textbf{}                      & \textbf{All}       & \textbf{Long}     & \textbf{Medium}   & \textbf{Short}    \\* \midrule
\textbf{Baseline}              & 0.00 (-0.03, 0.03) & --                & --                & --                \\
\textbf{0-Shot | Notes}        & 0.27 (0.23, 0.30)  & 0.29 (0.23, 0.36) & 0.25 (0.19, 0.32) & 0.24 (0.18, 0.31) \\
\textbf{0-Shot | Summary}      & 0.25 (0.21, 0.29)  & 0.31 (0.25, 0.37) & 0.22 (0.16, 0.29) & 0.22 (0.16, 0.28) \\
\textbf{5-Shot | Summary}      & 0.26 (0.23, 0.30)  & 0.31 (0.25, 0.38) & 0.24 (0.17, 0.31) & 0.22 (0.15, 0.29) \\
\textbf{10-Shot | Summary}     & 0.27 (0.23, 0.31)  & 0.28 (0.21, 0.34) & 0.26 (0.20, 0.33) & 0.24 (0.17, 0.30) \\
\textbf{20-Shot | Summary}     & 0.30 (0.26, 0.33)  & 0.31 (0.24, 0.38) & 0.29 (0.22, 0.35) & 0.26 (0.19, 0.33) \\
\textbf{50-Shot | Summary}     & 0.28 (0.24, 0.32)  & 0.33 (0.26, 0.40) & 0.24 (0.17, 0.31) & 0.25 (0.19, 0.32) \\
\end{longtable}

\begin{landscape}
\begin{longtable}[c]{@{}lllllll@{}}
\toprule
\multicolumn{7}{c}{\textbf{True Positive Rate (TPR, Sensitivity, Recall)}}                                                                          \\* \midrule
\endfirsthead
\multicolumn{7}{c}%
{{\bfseries Table \thetable\ continued from previous page}} \\
\toprule
\multicolumn{7}{c}{\textbf{True Positive Rate (TPR, Sensitivity, Recall)}}                                                                          \\* \midrule
\endhead
\bottomrule
\endfoot
\endlastfoot
 &
  \multicolumn{1}{r}{\textbf{1}} &
  \multicolumn{1}{r}{\textbf{2}} &
  \multicolumn{1}{r}{\textbf{3}} &
  \multicolumn{1}{r}{\textbf{4}} &
  \multicolumn{1}{r}{\textbf{5}} &
  \multicolumn{1}{r}{\textbf{6}} \\* \midrule
\textbf{Baseline}              & 0.18 (0.12, 0.24) & 0.24 (0.18, 0.31) & 0.22 (0.16, 0.28) & 0.23 (0.17, 0.29) & 0.12 (0.08, 0.17) & 0.02 (0.00, 0.04) \\
\textbf{0-Shot | Notes}        & 0.85 (0.62, 1.00) & 0.37 (0.32, 0.42) & 0.36 (0.31, 0.40) & 0.51 (0.44, 0.59) & 0.65 (0.52, 0.77) & 0.64 (0.00, 1.00) \\
\textbf{0-Shot | Summary}      & 0.89 (0.72, 1.00) & 0.35 (0.31, 0.40) & 0.35 (0.31, 0.40) & 0.52 (0.45, 0.61) & 0.55 (0.41, 0.68) & 0.62 (0.00, 1.00) \\
\textbf{5-Shot | Summary}      & 0.81 (0.61, 0.96) & 0.35 (0.30, 0.40) & 0.34 (0.30, 0.38) & 0.57 (0.49, 0.65) & 0.65 (0.52, 0.77) & 0.74 (0.00, 1.00) \\
\textbf{10-Shot | Summary}     & 0.89 (0.72, 1.00) & 0.35 (0.30, 0.40) & 0.35 (0.30, 0.40) & 0.53 (0.45, 0.61) & 0.62 (0.50, 0.74) & 0.86 (0.50, 1.00) \\
\textbf{20-Shot | Summary}     & 0.78 (0.63, 0.92) & 0.37 (0.33, 0.42) & 0.35 (0.30, 0.40) & 0.58 (0.50, 0.66) & 0.63 (0.52, 0.74) & 0.90 (0.67, 1.00) \\
\textbf{50-Shot | Summary}     & 0.67 (0.53, 0.80) & 0.37 (0.32, 0.42) & 0.33 (0.28, 0.37) & 0.55 (0.47, 0.64) & 0.64 (0.52, 0.74) & 0.80 (0.50, 1.00) \\
\end{longtable}

\begin{longtable}[c]{@{}lllllll@{}}
\toprule
\multicolumn{7}{c}{\textbf{True Negative Rate (TNR, Specificity)}}                                                                                     \\* \midrule
\endfirsthead
\multicolumn{7}{c}%
{{\bfseries Table \thetable\ continued from previous page}} \\
\toprule
\multicolumn{7}{c}{\textbf{True Negative Rate (TNR, Specificity)}}                                                                                     \\* \midrule
\endhead
\bottomrule
\endfoot
\endlastfoot
 &
  \multicolumn{1}{r}{\textbf{1}} &
  \multicolumn{1}{r}{\textbf{2}} &
  \multicolumn{1}{r}{\textbf{3}} &
  \multicolumn{1}{r}{\textbf{4}} &
  \multicolumn{1}{r}{\textbf{5}} &
  \multicolumn{1}{r}{\textbf{6}} \\* \midrule
\textbf{Baseline}              & 0.77 (0.74, 0.80) & 0.77 (0.74, 0.80) & 0.79 (0.76, 0.82) & 0.78 (0.75, 0.81) & 0.90 (0.88, 0.92) & 0.99 (0.99, 1.00) \\
\textbf{0-Shot | Notes}        & 0.79 (0.77, 0.82) & 0.85 (0.82, 0.87) & 0.88 (0.86, 0.91) & 0.83 (0.81, 0.86) & 0.93 (0.91, 0.94) & 0.99 (0.99, 1.00) \\
\textbf{0-Shot | Summary}      & 0.79 (0.77, 0.82) & 0.84 (0.81, 0.87) & 0.89 (0.86, 0.91) & 0.83 (0.80, 0.85) & 0.92 (0.90, 0.94) & 0.99 (0.99, 1.00) \\
\textbf{5-Shot | Summary}      & 0.79 (0.77, 0.82) & 0.83 (0.80, 0.86) & 0.88 (0.85, 0.91) & 0.84 (0.81, 0.86) & 0.93 (0.91, 0.95) & 0.99 (0.99, 1.00) \\
\textbf{10-Shot | Summary}     & 0.79 (0.77, 0.82) & 0.83 (0.80, 0.86) & 0.88 (0.85, 0.90) & 0.83 (0.81, 0.86) & 0.93 (0.92, 0.95) & 1.00 (0.99, 1.00) \\
\textbf{20-Shot | Summary}     & 0.80 (0.78, 0.82) & 0.85 (0.82, 0.88) & 0.87 (0.85, 0.90) & 0.84 (0.81, 0.86) & 0.94 (0.92, 0.95) & 1.00 (1.00, 1.00) \\
\textbf{50-Shot | Summary}     & 0.80 (0.78, 0.83) & 0.84 (0.81, 0.87) & 0.86 (0.83, 0.88) & 0.83 (0.81, 0.86) & 0.94 (0.93, 0.96) & 1.00 (1.00, 1.00) \\
\end{longtable}
\end{landscape}

\begin{landscape}
\begin{longtable}[c]{@{}lllllll@{}}
\toprule
\multicolumn{7}{c}{\textbf{Positive Predictive Value (PPV, Precision)}}                                                                                \\* \midrule
\endfirsthead
\multicolumn{7}{c}%
{{\bfseries Table \thetable\ continued from previous page}} \\
\toprule
\multicolumn{7}{c}{\textbf{Positive Predictive Value (PPV, Precision)}}                                                                                \\* \midrule
\endhead
\bottomrule
\endfoot
\endlastfoot
 &
  \multicolumn{1}{r}{\textbf{1}} &
  \multicolumn{1}{r}{\textbf{2}} &
  \multicolumn{1}{r}{\textbf{3}} &
  \multicolumn{1}{r}{\textbf{4}} &
  \multicolumn{1}{r}{\textbf{5}} &
  \multicolumn{1}{r}{\textbf{6}} \\* \midrule
\textbf{Baseline}              & 0.04 (0.03, 0.06) & 0.05 (0.04, 0.07) & 0.04 (0.03, 0.06) & 0.05 (0.04, 0.07) & 0.02 (0.01, 0.03) & 0.00 (0.00, 0.01) \\
\textbf{0-Shot | Notes}        & 0.01 (0.01, 0.02) & 0.18 (0.15, 0.21) & 0.18 (0.15, 0.21) & 0.11 (0.09, 0.14) & 0.04 (0.03, 0.05) & 0.00 (0.00, 0.01) \\
\textbf{0-Shot | Summary}      & 0.02 (0.01, 0.03) & 0.17 (0.14, 0.20) & 0.19 (0.15, 0.22) & 0.10 (0.08, 0.13) & 0.03 (0.02, 0.04) & 0.00 (0.00, 0.00) \\
\textbf{5-Shot | Summary}      & 0.02 (0.01, 0.03) & 0.15 (0.12, 0.18) & 0.18 (0.15, 0.22) & 0.11 (0.09, 0.14) & 0.04 (0.03, 0.06) & 0.00 (0.00, 0.01) \\
\textbf{10-Shot | Summary}     & 0.02 (0.01, 0.03) & 0.16 (0.13, 0.19) & 0.18 (0.15, 0.21) & 0.11 (0.08, 0.13) & 0.05 (0.03, 0.06) & 0.01 (0.00, 0.01) \\
\textbf{20-Shot | Summary}     & 0.03 (0.02, 0.05) & 0.18 (0.15, 0.21) & 0.17 (0.14, 0.20) & 0.11 (0.09, 0.13) & 0.05 (0.04, 0.07) & 0.01 (0.00, 0.01) \\
\textbf{50-Shot | Summary}     & 0.04 (0.03, 0.05) & 0.17 (0.14, 0.20) & 0.15 (0.13, 0.19) & 0.11 (0.08, 0.13) & 0.06 (0.04, 0.07) & 0.01 (0.00, 0.01) \\
\end{longtable}

\begin{longtable}[c]{@{}lllllll@{}}
\toprule
\multicolumn{7}{c}{\textbf{Negative Predictive Value (NPV)}}                                                                                           \\* \midrule
\endfirsthead
\multicolumn{7}{c}%
{{\bfseries Table \thetable\ continued from previous page}} \\
\toprule
\multicolumn{7}{c}{\textbf{Negative Predictive Value (NPV)}}                                                                                           \\* \midrule
\endhead
\bottomrule
\endfoot
\endlastfoot
 &
  \multicolumn{1}{r}{\textbf{1}} &
  \multicolumn{1}{r}{\textbf{2}} &
  \multicolumn{1}{r}{\textbf{3}} &
  \multicolumn{1}{r}{\textbf{4}} &
  \multicolumn{1}{r}{\textbf{5}} &
  \multicolumn{1}{r}{\textbf{6}} \\* \midrule
\textbf{Baseline}              & 0.81 (0.78, 0.84) & 0.84 (0.81, 0.86) & 0.85 (0.82, 0.87) & 0.83 (0.80, 0.85) & 0.83 (0.81, 0.86) & 0.85 (0.82, 0.87) \\
\textbf{0-Shot | Notes}        & 1.00 (0.99, 1.00) & 0.70 (0.67, 0.73) & 0.68 (0.65, 0.71) & 0.89 (0.87, 0.91) & 0.98 (0.97, 0.99) & 1.00 (1.00, 1.00) \\
\textbf{0-Shot | Summary}      & 1.00 (0.99, 1.00) & 0.69 (0.66, 0.72) & 0.66 (0.63, 0.69) & 0.91 (0.89, 0.93) & 0.98 (0.96, 0.98) & 1.00 (1.00, 1.00) \\
\textbf{5-Shot | Summary}      & 0.99 (0.99, 1.00) & 0.71 (0.68, 0.75) & 0.64 (0.61, 0.68) & 0.91 (0.89, 0.93) & 0.98 (0.97, 0.99) & 1.00 (1.00, 1.00) \\
\textbf{10-Shot | Summary}     & 1.00 (0.99, 1.00) & 0.70 (0.67, 0.73) & 0.67 (0.64, 0.71) & 0.91 (0.88, 0.93) & 0.97 (0.96, 0.98) & 1.00 (1.00, 1.00) \\
\textbf{20-Shot | Summary}     & 0.99 (0.98, 1.00) & 0.71 (0.67, 0.74) & 0.68 (0.65, 0.72) & 0.92 (0.90, 0.94) & 0.97 (0.96, 0.98) & 1.00 (1.00, 1.00) \\
\textbf{50-Shot | Summary}     & 0.98 (0.97, 0.99) & 0.72 (0.69, 0.75) & 0.68 (0.65, 0.71) & 0.91 (0.90, 0.93) & 0.97 (0.96, 0.98) & 1.00 (0.99, 1.00) \\
\end{longtable}
\end{landscape}

\normalsize

\subsection{Supplemental Table 4: Hospital Admission - Metrics}\label{supptable4-section}
\fontsize{7}{9}\selectfont

\begin{longtable}[c]{@{}lllll@{}}
\toprule
\multicolumn{5}{c}{\textbf{Hospital Admission: F1 Score}}                                                      \\* \midrule
\endfirsthead
\multicolumn{5}{c}%
{{\bfseries Table \thetable\ continued from previous page}} \\
\toprule
\multicolumn{5}{c}{\textbf{Hospital Admission: F1 Score}}                                                      \\* \midrule
\endhead
\bottomrule
\endfoot
\endlastfoot
\textbf{}                      & \textbf{All}      & \textbf{Long}     & \textbf{Medium}   & \textbf{Short}    \\* \midrule
\textbf{Baseline}              & 0.52 (0.49, 0.55) & --                & --                & --                \\
\textbf{0-Shot | Notes}        & 0.58 (0.55, 0.61) & 0.56 (0.51, 0.62) & 0.57 (0.52, 0.62) & 0.60 (0.54, 0.65) \\
\textbf{0-Shot | Summary}      & 0.62 (0.58, 0.65) & 0.62 (0.57, 0.67) & 0.61 (0.56, 0.66) & 0.61 (0.56, 0.66) \\
\textbf{5-Shot | Summary}      & 0.56 (0.53, 0.59) & 0.59 (0.54, 0.65) & 0.56 (0.51, 0.62) & 0.54 (0.48, 0.59) \\
\textbf{10-Shot | Summary}     & 0.59 (0.56, 0.62) & 0.59 (0.54, 0.64) & 0.59 (0.54, 0.65) & 0.58 (0.52, 0.63) \\
\textbf{20-Shot | Summary}     & 0.62 (0.59, 0.65) & 0.62 (0.56, 0.67) & 0.63 (0.58, 0.68) & 0.61 (0.55, 0.66) \\
\textbf{50-Shot | Summary}     & 0.61 (0.58, 0.64) & 0.63 (0.58, 0.68) & 0.62 (0.57, 0.67) & 0.57 (0.52, 0.63) \\
\end{longtable}

\begin{longtable}[c]{@{}lllll@{}}
\toprule
\multicolumn{5}{c}{\textbf{Hospital Admission: Matthew's Correlation Coefficient (MCC)}}                        \\* \midrule
\endfirsthead
\multicolumn{5}{c}%
{{\bfseries Table \thetable\ continued from previous page}} \\
\toprule
\multicolumn{5}{c}{\textbf{Hospital Admission: Matthew's Correlation Coefficient (MCC)}}                        \\* \midrule
\endhead
\bottomrule
\endfoot
\endlastfoot
\textbf{}                      & \textbf{All}       & \textbf{Long}     & \textbf{Medium}   & \textbf{Short}    \\* \midrule
\textbf{Baseline}              & 0.05 (-0.01, 0.11) & --                & --                & --                \\
\textbf{0-Shot | Notes}        & 0.28 (0.22, 0.33)  & 0.20 (0.09, 0.29) & 0.28 (0.18, 0.36) & 0.31 (0.22, 0.40) \\
\textbf{0-Shot | Summary}      & 0.27 (0.21, 0.33)  & 0.29 (0.19, 0.39) & 0.26 (0.16, 0.36) & 0.22 (0.11, 0.32) \\
\textbf{5-Shot | Summary}      & 0.25 (0.20, 0.31)  & 0.28 (0.19, 0.37) & 0.24 (0.14, 0.34) & 0.22 (0.11, 0.31) \\
\textbf{10-Shot | Summary}     & 0.28 (0.23, 0.34)  & 0.27 (0.17, 0.36) & 0.28 (0.19, 0.37) & 0.27 (0.17, 0.36) \\
\textbf{20-Shot | Summary}     & 0.33 (0.27, 0.38)  & 0.30 (0.21, 0.39) & 0.33 (0.23, 0.42) & 0.31 (0.21, 0.40) \\
\textbf{50-Shot | Summary}     & 0.29 (0.23, 0.34)  & 0.30 (0.20, 0.40) & 0.32 (0.23, 0.41) & 0.23 (0.12, 0.33) \\
\end{longtable}

\begin{longtable}[c]{@{}lllll@{}}
\toprule
\multicolumn{5}{c}{\textbf{Hospital Admission}}                                                                \\* \midrule
\endfirsthead
\endhead
\bottomrule
\endfoot
\endlastfoot
 &
  \textbf{\begin{tabular}[c]{@{}l@{}}True Positive \\ Rate (TPR, \\ Sensitivity, \\ Recall)\end{tabular}} &
  \textbf{\begin{tabular}[c]{@{}l@{}}True Negative \\ Rate (TNR, \\ Specificity)\end{tabular}} &
  \textbf{\begin{tabular}[c]{@{}l@{}}Positive \\ Predictive \\ Value (PPV, \\ Precision)\end{tabular}} &
  \textbf{\begin{tabular}[c]{@{}l@{}}Negative \\ Predictive \\ Value (NPV)\end{tabular}} \\* \midrule
\textbf{Baseline}              & 0.42 (0.38, 0.46) & 0.63 (0.59, 0.67) & 0.36 (0.30, 0.41) & 0.51 (0.47, 0.55) \\
\textbf{0-Shot | Notes}        & 0.48 (0.44, 0.52) & 0.82 (0.77, 0.86) & 0.57 (0.50, 0.65) & 0.38 (0.35, 0.42) \\
\textbf{0-Shot | Summary}      & 0.51 (0.47, 0.55) & 0.76 (0.72, 0.80) & 0.49 (0.43, 0.56) & 0.53 (0.49, 0.57) \\
\textbf{5-Shot | Summary}      & 0.47 (0.44, 0.51) & 0.81 (0.76, 0.85) & 0.57 (0.49, 0.65) & 0.37 (0.33, 0.40) \\
\textbf{10-Shot | Summary}     & 0.49 (0.45, 0.52) & 0.82 (0.77, 0.86) & 0.56 (0.49, 0.64) & 0.41 (0.37, 0.45) \\
\textbf{20-Shot | Summary}     & 0.51 (0.47, 0.55) & 0.83 (0.79, 0.86) & 0.56 (0.49, 0.63) & 0.46 (0.43, 0.50) \\
\textbf{50-Shot | Summary}     & 0.50 (0.46, 0.54) & 0.79 (0.75, 0.83) & 0.53 (0.46, 0.60) & 0.48 (0.44, 0.52) \\
\end{longtable}

\normalsize

\subsection{Supplemental Table 5: Intensive Care Unit (ICU) Admission - Metrics}\label{supptable5-section}
\fontsize{7}{9}\selectfont

\begin{longtable}[c]{@{}lllll@{}}
\toprule
\multicolumn{5}{c}{\textbf{ICU Admission: F1 Score}}                                                           \\* \midrule
\endfirsthead
\multicolumn{5}{c}%
{{\bfseries Table \thetable\ continued from previous page}} \\
\toprule
\multicolumn{5}{c}{\textbf{ICU Admission: F1 Score}}                                                           \\* \midrule
\endhead
\bottomrule
\endfoot
\endlastfoot
\textbf{}                      & \textbf{All}      & \textbf{Long}     & \textbf{Medium}   & \textbf{Short}    \\* \midrule
\textbf{Baseline}              & 0.52 (0.49, 0.56) & --                & --                & --                \\
\textbf{0-Shot | Notes}        & 0.80 (0.77, 0.82) & 0.74 (0.70, 0.79) & 0.82 (0.78, 0.86) & 0.83 (0.79, 0.87) \\
\textbf{0-Shot | Summary}      & 0.77 (0.74, 0.80) & 0.74 (0.69, 0.78) & 0.78 (0.73, 0.82) & 0.79 (0.75, 0.84) \\
\textbf{5-Shot | Summary}      & 0.78 (0.76, 0.81) & 0.75 (0.70, 0.79) & 0.80 (0.76, 0.84) & 0.80 (0.76, 0.84) \\
\textbf{10-Shot | Summary}     & 0.79 (0.77, 0.82) & 0.77 (0.72, 0.81) & 0.78 (0.74, 0.83) & 0.82 (0.78, 0.86) \\
\textbf{20-Shot | Summary}     & 0.79 (0.76, 0.81) & 0.76 (0.71, 0.80) & 0.78 (0.74, 0.83) & 0.81 (0.77, 0.86) \\
\textbf{50-Shot | Summary}     & 0.80 (0.77, 0.82) & 0.77 (0.72, 0.81) & 0.82 (0.78, 0.86) & 0.81 (0.77, 0.85) \\
\end{longtable}

\begin{longtable}[c]{@{}lllll@{}}
\toprule
\multicolumn{5}{c}{\textbf{ICU Admission: Matthew's Correlation Coefficient (MCC)}}                             \\* \midrule
\endfirsthead
\multicolumn{5}{c}%
{{\bfseries Table \thetable\ continued from previous page}} \\
\toprule
\multicolumn{5}{c}{\textbf{ICU Admission: Matthew's Correlation Coefficient (MCC)}}                             \\* \midrule
\endhead
\bottomrule
\endfoot
\endlastfoot
\textbf{}                      & \textbf{All}       & \textbf{Long}     & \textbf{Medium}   & \textbf{Short}    \\* \midrule
\textbf{Baseline}              & 0.05 (-0.01, 0.11) & --                & --                & --                \\
\textbf{0-Shot | Notes}        & 0.60 (0.55, 0.65)  & 0.46 (0.36, 0.55) & 0.64 (0.55, 0.72) & 0.65 (0.56, 0.73) \\
\textbf{0-Shot | Summary}      & 0.54 (0.49, 0.59)  & 0.45 (0.35, 0.55) & 0.56 (0.48, 0.65) & 0.56 (0.47, 0.65) \\
\textbf{5-Shot | Summary}      & 0.57 (0.51, 0.62)  & 0.47 (0.36, 0.56) & 0.61 (0.52, 0.69) & 0.59 (0.49, 0.68) \\
\textbf{10-Shot | Summary}     & 0.58 (0.53, 0.63)  & 0.52 (0.42, 0.60) & 0.57 (0.49, 0.66) & 0.62 (0.54, 0.71) \\
\textbf{20-Shot | Summary}     & 0.57 (0.52, 0.62)  & 0.49 (0.39, 0.59) & 0.57 (0.48, 0.66) & 0.61 (0.52, 0.70) \\
\textbf{50-Shot | Summary}     & 0.60 (0.55, 0.65)  & 0.51 (0.42, 0.60) & 0.65 (0.56, 0.72) & 0.61 (0.52, 0.70) \\
\end{longtable}

\begin{longtable}[c]{@{}lllll@{}}
\toprule
\multicolumn{5}{c}{\textbf{ICU Admission}}                                                                     \\* \midrule
\endfirsthead
\endhead
\bottomrule
\endfoot
\endlastfoot
 &
  \textbf{\begin{tabular}[c]{@{}l@{}}True Positive \\ Rate (TPR, \\ Sensitivity, \\ Recall)\end{tabular}} &
  \textbf{\begin{tabular}[c]{@{}l@{}}True Negative \\ Rate (TNR, \\ Specificity)\end{tabular}} &
  \textbf{\begin{tabular}[c]{@{}l@{}}Positive \\ Predictive \\ Value (PPV, \\ Precision)\end{tabular}} &
  \textbf{\begin{tabular}[c]{@{}l@{}}Negative \\ Predictive \\ Value (NPV)\end{tabular}} \\* \midrule
\textbf{Baseline}              & 0.42 (0.38, 0.46) & 0.63 (0.59, 0.67) & 0.36 (0.30, 0.41) & 0.51 (0.47, 0.55) \\
\textbf{0-Shot | Notes}        & 0.48 (0.44, 0.52) & 0.82 (0.77, 0.86) & 0.57 (0.50, 0.65) & 0.38 (0.35, 0.42) \\
\textbf{0-Shot | Summary}      & 0.51 (0.47, 0.55) & 0.76 (0.72, 0.80) & 0.49 (0.43, 0.56) & 0.53 (0.49, 0.57) \\
\textbf{5-Shot | Summary}      & 0.47 (0.44, 0.51) & 0.81 (0.76, 0.85) & 0.57 (0.49, 0.65) & 0.37 (0.33, 0.40) \\
\textbf{10-Shot | Summary}     & 0.49 (0.45, 0.52) & 0.82 (0.77, 0.86) & 0.56 (0.49, 0.64) & 0.41 (0.37, 0.45) \\
\textbf{20-Shot | Summary}     & 0.51 (0.47, 0.55) & 0.83 (0.79, 0.86) & 0.56 (0.49, 0.63) & 0.46 (0.43, 0.50) \\
\textbf{50-Shot | Summary}     & 0.50 (0.46, 0.54) & 0.79 (0.75, 0.83) & 0.53 (0.46, 0.60) & 0.48 (0.44, 0.52) \\
\end{longtable}

\normalsize

\subsection{Supplemental Table 6: Unplanned Admission - Metrics}\label{supptable6-section}
\fontsize{7}{9}\selectfont

\begin{longtable}[c]{@{}lllll@{}}
\toprule
\multicolumn{5}{c}{\textbf{Unplanned Admission: F1 Score}}                                                     \\* \midrule
\endfirsthead
\multicolumn{5}{c}%
{{\bfseries Table \thetable\ continued from previous page}} \\
\toprule
\multicolumn{5}{c}{\textbf{Unplanned Admission: F1 Score}}                                                     \\* \midrule
\endhead
\bottomrule
\endfoot
\endlastfoot
\textbf{}                      & \textbf{All}      & \textbf{Long}     & \textbf{Medium}   & \textbf{Short}    \\* \midrule
\textbf{Baseline}              & 0.45 (0.42, 0.48) & --                & --                & --                \\
\textbf{0-Shot | Notes}        & 0.60 (0.57, 0.63) & 0.59 (0.53, 0.64) & 0.63 (0.58, 0.68) & 0.58 (0.53, 0.64) \\
\textbf{0-Shot | Summary}      & 0.61 (0.58, 0.64) & 0.59 (0.53, 0.64) & 0.64 (0.59, 0.69) & 0.60 (0.55, 0.66) \\
\textbf{5-Shot | Summary}      & 0.55 (0.52, 0.59) & 0.56 (0.50, 0.61) & 0.59 (0.53, 0.64) & 0.52 (0.46, 0.57) \\
\textbf{10-Shot | Summary}     & 0.57 (0.54, 0.60) & 0.57 (0.52, 0.63) & 0.58 (0.53, 0.63) & 0.57 (0.51, 0.63) \\
\textbf{20-Shot | Summary}     & 0.58 (0.54, 0.61) & 0.58 (0.52, 0.63) & 0.58 (0.53, 0.63) & 0.57 (0.52, 0.63) \\
\textbf{50-Shot | Summary}     & 0.58 (0.55, 0.61) & 0.57 (0.52, 0.63) & 0.59 (0.53, 0.64) & 0.59 (0.54, 0.64) \\
\end{longtable}

\begin{longtable}[c]{@{}lllll@{}}
\toprule
\multicolumn{5}{c}{\textbf{Unplanned Admission: Matthew's Correlation Coefficient (MCC)}}                          \\* \midrule
\endfirsthead
\multicolumn{5}{c}%
{{\bfseries Table \thetable\ continued from previous page}} \\
\toprule
\multicolumn{5}{c}{\textbf{Unplanned Admission: Matthew's Correlation Coefficient (MCC)}}                          \\* \midrule
\endhead
\bottomrule
\endfoot
\endlastfoot
\textbf{}                      & \textbf{All}         & \textbf{Long}     & \textbf{Medium}   & \textbf{Short}     \\* \midrule
\textbf{Baseline}              & -0.10 (-0.16, -0.03) & --                & --                & --                 \\
\textbf{0-Shot | Notes}        & 0.22 (0.15, 0.28)    & 0.20 (0.09, 0.30) & 0.28 (0.18, 0.38) & 0.17 (0.07, 0.28)  \\
\textbf{0-Shot | Summary}      & 0.22 (0.16, 0.28)    & 0.19 (0.08, 0.30) & 0.27 (0.17, 0.37) & 0.20 (0.10, 0.31)  \\
\textbf{5-Shot | Summary}      & 0.11 (0.05, 0.17)    & 0.12 (0.01, 0.23) & 0.17 (0.06, 0.28) & 0.04 (-0.07, 0.15) \\
\textbf{10-Shot | Summary}     & 0.15 (0.09, 0.21)    & 0.16 (0.05, 0.26) & 0.15 (0.05, 0.26) & 0.14 (0.03, 0.25)  \\
\textbf{20-Shot | Summary}     & 0.16 (0.09, 0.22)    & 0.17 (0.06, 0.27) & 0.16 (0.05, 0.26) & 0.15 (0.04, 0.26)  \\
\textbf{50-Shot | Summary}     & 0.17 (0.11, 0.23)    & 0.15 (0.04, 0.26) & 0.17 (0.07, 0.28) & 0.19 (0.09, 0.29)  \\
\end{longtable}

\begin{longtable}[c]{@{}lllll@{}}
\toprule
\multicolumn{5}{c}{\textbf{Unplanned Admission}}                                                               \\* \midrule
\endfirsthead
\endhead
\bottomrule
\endfoot
\endlastfoot
 &
  \textbf{\begin{tabular}[c]{@{}l@{}}True Positive \\ Rate (TPR, \\ Sensitivity, \\ Recall)\end{tabular}} &
  \textbf{\begin{tabular}[c]{@{}l@{}}True Negative \\ Rate (TNR, \\ Specificity)\end{tabular}} &
  \textbf{\begin{tabular}[c]{@{}l@{}}Positive \\ Predictive \\ Value (PPV, \\ Precision)\end{tabular}} &
  \textbf{\begin{tabular}[c]{@{}l@{}}Negative \\ Predictive \\ Value (NPV)\end{tabular}} \\* \midrule
\textbf{Baseline}              & 0.46 (0.41, 0.50) & 0.45 (0.40, 0.49) & 0.47 (0.39, 0.55) & 0.45 (0.40, 0.49) \\
\textbf{0-Shot | Notes}        & 0.57 (0.53, 0.61) & 0.66 (0.61, 0.72) & 0.82 (0.72, 0.94) & 0.39 (0.35, 0.43) \\
\textbf{0-Shot | Summary}      & 0.60 (0.56, 0.64) & 0.63 (0.58, 0.67) & 0.71 (0.61, 0.81) & 0.52 (0.48, 0.57) \\
\textbf{5-Shot | Summary}      & 0.55 (0.51, 0.59) & 0.56 (0.51, 0.61) & 0.67 (0.57, 0.77) & 0.45 (0.41, 0.50) \\
\textbf{10-Shot | Summary}     & 0.56 (0.52, 0.60) & 0.60 (0.54, 0.64) & 0.72 (0.62, 0.82) & 0.44 (0.40, 0.48) \\
\textbf{20-Shot | Summary}     & 0.56 (0.52, 0.60) & 0.60 (0.55, 0.65) & 0.73 (0.62, 0.84) & 0.44 (0.40, 0.48) \\
\textbf{50-Shot | Summary}     & 0.57 (0.53, 0.61) & 0.61 (0.56, 0.66) & 0.74 (0.64, 0.85) & 0.44 (0.39, 0.48) \\
\end{longtable}

\normalsize

\subsection{Supplemental Table 7: Hospital Mortality - Metrics}\label{supptable7-section}
\fontsize{7}{9}\selectfont

\begin{longtable}[c]{@{}lllll@{}}
\toprule
\multicolumn{5}{c}{\textbf{Hospital Mortality: F1 Score}}                                                      \\* \midrule
\endfirsthead
\multicolumn{5}{c}%
{{\bfseries Table \thetable\ continued from previous page}} \\
\toprule
\multicolumn{5}{c}{\textbf{Hospital Mortality: F1 Score}}                                                      \\* \midrule
\endhead
\bottomrule
\endfoot
\endlastfoot
\textbf{}                      & \textbf{All}      & \textbf{Long}     & \textbf{Medium}   & \textbf{Short}    \\* \midrule
\textbf{Baseline}              & 0.49 (0.45, 0.53) & --                & --                & --                \\
\textbf{0-Shot | Notes}        & 0.74 (0.70, 0.78) & 0.73 (0.67, 0.80) & 0.79 (0.73, 0.84) & 0.70 (0.64, 0.77) \\
\textbf{0-Shot | Summary}      & 0.75 (0.72, 0.79) & 0.78 (0.72, 0.84) & 0.78 (0.72, 0.83) & 0.70 (0.63, 0.76) \\
\textbf{5-Shot | Summary}      & 0.84 (0.81, 0.87) & 0.85 (0.80, 0.90) & 0.89 (0.84, 0.93) & 0.78 (0.72, 0.84) \\
\textbf{10-Shot | Summary}     & 0.86 (0.83, 0.89) & 0.89 (0.84, 0.93) & 0.89 (0.85, 0.93) & 0.79 (0.73, 0.85) \\
\textbf{20-Shot | Summary}     & 0.86 (0.83, 0.89) & 0.89 (0.84, 0.93) & 0.89 (0.84, 0.93) & 0.81 (0.75, 0.86) \\
\textbf{50-Shot | Summary}     & 0.84 (0.81, 0.87) & 0.84 (0.79, 0.90) & 0.88 (0.83, 0.92) & 0.81 (0.75, 0.86) \\
\end{longtable}

\begin{longtable}[c]{@{}lllll@{}}
\toprule
\multicolumn{5}{c}{\textbf{Hospital Mortality: Matthew's Correlation Coefficient (MCC)}}                         \\* \midrule
\endfirsthead
\multicolumn{5}{c}%
{{\bfseries Table \thetable\ continued from previous page}} \\
\toprule
\multicolumn{5}{c}{\textbf{Hospital Mortality: Matthew's Correlation Coefficient (MCC)}}                         \\* \midrule
\endhead
\bottomrule
\endfoot
\endlastfoot
\textbf{}                      & \textbf{All}        & \textbf{Long}     & \textbf{Medium}   & \textbf{Short}    \\* \midrule
\textbf{Baseline}              & -0.02 (-0.11, 0.06) & --                & --                & --                \\
\textbf{0-Shot | Notes}        & 0.56 (0.51, 0.61)   & 0.58 (0.50, 0.66) & 0.61 (0.52, 0.70) & 0.47 (0.38, 0.56) \\
\textbf{0-Shot | Summary}      & 0.58 (0.53, 0.63)   & 0.64 (0.56, 0.72) & 0.59 (0.50, 0.68) & 0.46 (0.37, 0.55) \\
\textbf{5-Shot | Summary}      & 0.70 (0.65, 0.75)   & 0.72 (0.63, 0.81) & 0.79 (0.71, 0.87) & 0.59 (0.48, 0.68) \\
\textbf{10-Shot | Summary}     & 0.73 (0.68, 0.78)   & 0.79 (0.71, 0.87) & 0.78 (0.69, 0.86) & 0.60 (0.50, 0.70) \\
\textbf{20-Shot | Summary}     & 0.73 (0.67, 0.78)   & 0.78 (0.68, 0.86) & 0.77 (0.68, 0.86) & 0.62 (0.52, 0.73) \\
\textbf{50-Shot | Summary}     & 0.70 (0.65, 0.75)   & 0.71 (0.62, 0.80) & 0.76 (0.67, 0.84) & 0.62 (0.51, 0.72) \\
\end{longtable}

\begin{longtable}[c]{@{}lllll@{}}
\toprule
\multicolumn{5}{c}{\textbf{Hospital Mortality}}                                                                \\* \midrule
\endfirsthead
\endhead
\bottomrule
\endfoot
\endlastfoot
 &
  \textbf{\begin{tabular}[c]{@{}l@{}}True Positive \\ Rate (TPR, \\ Sensitivity, \\ Recall)\end{tabular}} &
  \textbf{\begin{tabular}[c]{@{}l@{}}True Negative \\ Rate (TNR, \\ Specificity)\end{tabular}} &
  \textbf{\begin{tabular}[c]{@{}l@{}}Positive \\ Predictive \\ Value (PPV, \\ Precision)\end{tabular}} &
  \textbf{\begin{tabular}[c]{@{}l@{}}Negative \\ Predictive \\ Value (NPV)\end{tabular}} \\* \midrule
\textbf{Baseline}              & 0.50 (0.44, 0.55) & 0.48 (0.42, 0.54) & 0.52 (0.42, 0.63) & 0.47 (0.41, 0.53) \\
\textbf{0-Shot | Notes}        & 0.99 (0.97, 1.00) & 0.66 (0.61, 0.70) & 0.51 (0.42, 0.62) & 0.99 (0.98, 1.00) \\
\textbf{0-Shot | Summary}      & 0.99 (0.97, 1.00) & 0.67 (0.62, 0.71) & 0.53 (0.44, 0.64) & 0.99 (0.98, 1.00) \\
\textbf{5-Shot | Summary}      & 0.95 (0.92, 0.98) & 0.77 (0.73, 0.81) & 0.74 (0.62, 0.88) & 0.96 (0.94, 0.98) \\
\textbf{10-Shot | Summary}     & 0.95 (0.92, 0.97) & 0.79 (0.75, 0.84) & 0.79 (0.66, 0.93) & 0.96 (0.93, 0.98) \\
\textbf{20-Shot | Summary}     & 0.93 (0.90, 0.96) & 0.81 (0.76, 0.85) & 0.81 (0.68, 0.96) & 0.94 (0.91, 0.97) \\
\textbf{50-Shot | Summary}     & 0.93 (0.90, 0.96) & 0.78 (0.73, 0.82) & 0.76 (0.64, 0.91) & 0.95 (0.92, 0.97) \\
\end{longtable}

\normalsize

\subsection{Supplemental Table 8: Post-Anesthesia Care Unit (PACU) - Phase 1 Duration - Metrics}\label{supptable8-section}
\fontsize{7}{9}\selectfont
\begin{longtable}[c]{@{}lllll@{}}
\toprule
\multicolumn{5}{c}{\textbf{Phase 1 PACU Duration: Mean Absolute Error (MAE) in minutes}}                                   \\* \midrule
\endfirsthead
\multicolumn{5}{c}%
{{\bfseries Table \thetable\ continued from previous page}} \\
\toprule
\multicolumn{5}{c}{\textbf{Phase 1 PACU Duration: Mean Absolute Error (MAE) in minutes}}                                   \\* \midrule
\endhead
\bottomrule
\endfoot
\endlastfoot
\textbf{}                      & \textbf{All}         & \textbf{Long}        & \textbf{Medium}      & \textbf{Short}       \\* \midrule
\textbf{Baseline}              & 36.12 (34.02, 38.28) & --                   & --                   & --                   \\
\textbf{0-Shot | Notes}        & 49.01 (46.89, 51.15) & 53.30 (49.11, 57.59) & 46.37 (42.93, 50.03) & 47.24 (44.04, 50.67) \\
\textbf{0-Shot | Summary}      & 47.03 (45.11, 48.98) & 49.99 (46.56, 53.68) & 44.64 (41.46, 47.86) & 46.35 (43.23, 49.87) \\
\textbf{5-Shot | Summary}      & 69.06 (65.87, 72.25) & 76.98 (71.67, 82.52) & 68.89 (63.42, 74.61) & 61.24 (56.83, 65.85) \\
\textbf{10-Shot | Summary}     & 69.14 (66.09, 72.27) & 74.55 (69.20, 80.12) & 68.24 (62.82, 73.70) & 64.43 (59.63, 69.49) \\
\textbf{20-Shot | Summary}     & 69.74 (66.78, 72.86) & 75.90 (70.69, 81.18) & 69.23 (63.85, 74.85) & 63.96 (59.25, 68.74) \\
\textbf{50-Shot | Summary}     & 60.09 (57.19, 63.06) & 63.94 (58.47, 69.51) & 59.72 (54.91, 64.80) & 56.42 (51.73, 61.18) \\
\end{longtable}
\normalsize

\subsection{Supplemental Table 9: Hospital Duration - Metrics}\label{supptable9-section}
\fontsize{7}{9}\selectfont
\begin{longtable}[c]{@{}lllll@{}}
\toprule
\multicolumn{5}{c}{\textbf{Hospital Duration: Mean Absolute Error (MAE) in days}}                               \\* \midrule
\endfirsthead
\multicolumn{5}{c}%
{{\bfseries Table \thetable\ continued from previous page}} \\
\toprule
\multicolumn{5}{c}{\textbf{Hospital Duration: Mean Absolute Error (MAE) in days}}                               \\* \midrule
\endhead
\bottomrule
\endfoot
\endlastfoot
\textbf{}                      & \textbf{All}      & \textbf{Long}      & \textbf{Medium}   & \textbf{Short}    \\* \midrule
\textbf{Baseline}              & 5.42 (5.04, 5.85) & --                 & --                & --                \\
\textbf{0-Shot | Notes}        & 4.80 (4.40, 5.26) & 5.47 (4.66, 6.45)  & 4.43 (3.84, 5.11) & 4.49 (3.87, 5.23) \\
\textbf{0-Shot | Summary}      & 4.72 (4.31, 5.17) & 5.17 (4.37, 6.14)  & 4.55 (3.98, 5.20) & 4.42 (3.80, 5.18) \\
\textbf{5-Shot | Summary}      & 6.87 (6.39, 7.39) & 7.98 (6.99, 9.08)  & 7.03 (6.24, 7.87) & 5.60 (4.89, 6.39) \\
\textbf{10-Shot | Summary}     & 7.81 (7.30, 8.35) & 9.06 (8.04, 10.11) & 7.87 (7.02, 8.75) & 6.47 (5.72, 7.35) \\
\textbf{20-Shot | Summary}     & 7.70 (7.16, 8.24) & 9.22 (8.17, 10.31) & 7.89 (7.00, 8.82) & 5.95 (5.19, 6.84) \\
\textbf{50-Shot | Summary}     & 7.87 (7.35, 8.42) & 8.74 (7.83, 9.81)  & 8.09 (7.24, 8.98) & 6.72 (5.91, 7.65) \\
\end{longtable}
\normalsize

\subsection{Supplemental Table 10: Intensive Care Unit (ICU) Duration - Metrics}\label{supptable10-section}
\fontsize{7}{9}\selectfont
\begin{longtable}[c]{@{}lllll@{}}
\toprule
\multicolumn{5}{c}{\textbf{ICU Duration: Mean Absolute Error (MAE) in days}}                                   \\* \midrule
\endfirsthead
\multicolumn{5}{c}%
{{\bfseries Table \thetable\ continued from previous page}} \\
\toprule
\multicolumn{5}{c}{\textbf{ICU Duration: Mean Absolute Error (MAE) in days}}                                   \\* \midrule
\endhead
\bottomrule
\endfoot
\endlastfoot
\textbf{}                      & \textbf{All}      & \textbf{Long}     & \textbf{Medium}   & \textbf{Short}    \\* \midrule
\textbf{Baseline}              & 1.15 (0.99, 1.33) & --                & --                & --                \\
\textbf{0-Shot | Notes}        & 1.68 (1.51, 1.86) & 2.12 (1.80, 2.46) & 1.82 (1.47, 2.27) & 1.12 (0.96, 1.30) \\
\textbf{0-Shot | Summary}      & 1.67 (1.51, 1.84) & 2.07 (1.78, 2.38) & 1.77 (1.42, 2.20) & 1.17 (1.01, 1.34) \\
\textbf{5-Shot | Summary}      & 2.28 (2.12, 2.45) & 2.60 (2.31, 2.90) & 2.39 (2.05, 2.80) & 1.87 (1.67, 2.10) \\
\textbf{10-Shot | Summary}     & 2.19 (2.00, 2.38) & 2.59 (2.26, 2.94) & 2.24 (1.89, 2.67) & 1.75 (1.53, 1.99) \\
\textbf{20-Shot | Summary}     & 1.80 (1.62, 1.99) & 2.20 (1.90, 2.53) & 1.91 (1.55, 2.33) & 1.31 (1.10, 1.54) \\
\textbf{50-Shot | Summary}     & 1.11 (0.94, 1.29) & 1.51 (1.20, 1.86) & 1.19 (0.85, 1.61) & 0.62 (0.51, 0.74) \\
\end{longtable}
\normalsize

\pagebreak
\section{Supplemental Figures}\label{supplemental-figures}

\subsection{Supplemental Figure 1: Representative Text Prompts}\label{suppfig1-prompts}
\ref{suppfig-system_message} to \ref{suppfig-cot_rationale_generation} are representative examples of all prompts used in experiments. The prompts are given followed by the LLM output after the dashed line. Prompts are presented with color-coded highlights to assist identification of text source and salient portions of the prompt. We show all prompts and corresponding LLM outputs using the same patient case and procedure for \ref{suppfig-system_message} to \ref{suppfig-summary_generation} to enable comparison between the different prompt strategies. Few-shot prompts in \ref{suppfig-fewshot_summary} and \ref{suppfig-fewshot_cot_summary} are illustrated with 4 in-context examples, but in our experiments we use 5, 10, 20, and 50 in-context examples. The prompt in \ref{suppfig-cot_rationale_generation} uses a different patient case and procedure because it is only used for generating Chain-of-Thought rationale for the in-context examples depicted in prompt for \ref{suppfig-fewshot_cot_summary} “Few-shot CoT Q\&A From Notes Summary”.

While the content of these examples derived from a real patient and case from the electronic health record, all PHI and PII are removed with names obfuscated, and dates and times shifted. \\

\begin{tabular}{@{}ll@{}}
\centering
\textbf{Highlight Color} & \textbf{Text Source/Type} \\
\midrule
\hlgreen{\textbf{Green}} & Task Specification \\
\hlyellow{\textbf{Yellow}} & Procedure Information from Case Booking \\
\hlred{\textbf{Red}} & Note Metadata \\
\hlpurple{\textbf{Purple}} & Note Text for Query Case \\
\hlpink{\textbf{Pink}} & Summarized Note Text for Query Case \\
\hlblue{\textbf{Blue}} & In-Context Examples \\
\hlorange{\textbf{Orange}} & In-Context Examples Chain-of-Thought Rationale \\
\hlbrown{\textbf{Brown}} & Modifications to Prompt to Induce Chain-of-Thought Reasoning \\
\end{tabular}%

\subsubsection{System Message}\label{suppfig-system_message}
\begin{tcolorbox}[enhanced jigsaw,breakable,pad at break*=1mm,
  colback=green!5!white,colframe=green!50!black,
  colbacktitle=green!75!black,title=System Message]
  You are a physician working in a hospital surgery center who is assessing patients to determine their outcome after a procedure.
\end{tcolorbox}

\subsubsection{Zero-shot Q\&A From Original Notes}\label{suppfig-zeroshot_notes}
\begin{tcolorbox}[enhanced jigsaw,breakable,pad at break*=1mm,
  colback=yellow!5!white,colframe=yellow!50!black,
  colbacktitle=yellow!75!black,title=Prompt \& LLM Output]
\obeylines
You are given a task and context. The context contains information from the proposed procedure and patient's medical record. Assess the patient in the context of the proposed procedure and then provide an answer.
\vspace{8pt}
Task: \hlgreen{What is the patient's ASA Physical Status Classification?}
\vspace{4pt}
Context:
Procedure: \hlyellow{A-FIB ABLATION}
Procedure Description: \hlyellow{A-Fib Ablation (N/A )}
Diagnosis: \hlyellow{Paroxysmal atrial fibrillation (HCC)}
Provider Service: \hlyellow{Cardiovascular}
\vspace{4pt}
Medical Record Notes:
\hlred{Progress Notes written by Physician at 2020-07-02 12:23:00:}
\hlpurple{ELECTROPHYSIOLOGY CLINIC INITIAL CONSULT	REASON FOR CONSULT:  Atrial fibrillation 	REFERRING:  Self	PRIMARY CARE: No primary care provider on file.	Distant Site Telemedicine Encounter	I conducted this encounter from Marvin Monroe Memorial Hospital via secure, live, face-to-face video conference with the patient. Homer was located at home with his wife. I reviewed the risks and benefits of telemedicine as pertinent to this visit and the patient agreed to proceed.  	PROBLEM LIST  1. Atrial fibrillation 	HPI:   Mr. Simpson is a 62 year old male referred for evaluation of atrial fibrillation. He has a long standing history of episodic palpitations for several years. After an episode of the same earlier in 2020 he was seen by his PCP who noted that 12 lead ECG showed atrial fibrillation. These episodes were paroxysmal in nature without clear provoking triggers or palliating factors. He had a more sustained episode of palpitations and light headedness on a flight to Springfield where he ultimately was admitted to a hospital for evaluation. ECG and telemetry showed persistent atrial fibrillation and he underwent a TEE cardioversion. He was started on Xarelto and advised to continue for 4 weeks. He had a TTE showing normal LV function and no valvular heart disease. Imaging also showed possible cholecystitis though HIDA scan was most in keeping with biliary dyskinesia. 	Today he joins me via telemedicine. He has felt generally well since his cardioversion and discharge. He has not had a recurrence of rapid heart action. He has continued Xarelto. He is an avid outdoorsman and has been avoiding higher risk physical activity in light of his anticoagulation use. He denies pre-syncope or syncope. He has not had chest pain. He denies peripheral edema, orthopnea or PND. 	ECG June 19, 2020 (personally reviewed) shows atrial fibrillation with IVCD. 	CARDIOLOGY STUDIES	TEE [June 2020]: NO LAA thrombus, LVEF 0.65, trace  TR.   	PAST MEDICAL HISTORY  Patient Active Problem List	Diagnosis Date Noted   • Paroxysmal atrial fibrillation (HCC) [I48.0] 07/02/2020 	Added automatically from request for surgery 127730     	SOCIAL HISTORY:   Married. Non-smoker. Rare EtOH. Works for local tech company. 	FAMILY HISTORY:  No heart failure, SCD.	ALLERGIES:  Patient has no allergy information on record.	Current Outpatient Medications   Medication Sig Dispense Refill   • rivaroxaban (Xarelto) 20 MG tablet Take 1 tablet (20 mg) by mouth daily. 90 tablet 0 	No current facility-administered medications for this visit.    	REVIEW OF SYSTEMS 	Other than HPI, all other systems are negative. 	PHYSICAL EXAM  VITAL SIGNS: There were no vitals taken for this visit.  Appears well   	IMPRESSION:   Mr. Simpson is a 62 year old male with the following salient arrhythmic issues: 	1. Atrial fibrillation 	We discussed atrial fibrillation in detail, including the risks factors, triggers, natural history, stroke risk prediction and treatment options.  Specifically, we discussed a rate control strategy with medications versus a rhythm control strategy.  For many patients, rate control is sufficient to maintain quality of life and cardiovascular function.  We discussed the antiarrhythmic drugs in detail, including class Ic and class III AADs.  We also discussed the importance of lifestyle changes in improving both ablative and non-ablative approaches to AF, namely weight reduction and exercise.	Given his clearly symptomatic atrial fibrillation we discussed the rationale for rhythm control. Given his low resting heart rate and background significant physical activity, we discussed that AAD options are not likely to be tolerated or less well tolerated. 	I therefore discussed ablation.  I explained that the chances of success with this procedure is about 75\% but may need more than one procedure.  I explained how a CT or MRI  of his heart is completed ahead of time to assess his pulmonary vein anatomy.  The procedure is completed under general anesthesia, possibly with a transesophageal echocardiogram.  I explained that a transseptal puncture is completed and a wide-area circumferential ablation around the pulmonary veins is completed.  I explained that he may go home the same day or may spend one night in the hospital and go home the next day.  I said the risks of the procedure included, but were not limited to bleeding, infection, cardiac perforation and tamponade, 1 in 200 chance of a stroke, gastroparesis, pulmonary vein stenosis, atrial-esophageal fistula, and 1 in a 1000 chance of death.	After discussion, he would like to proceed to catheter ablation. 	PLAN:  1. AF ablation August 22, 2020  2. Cardiac CT for PV anatomy   3. Continue Xarelto through ablation and minimum 6 weeks after 	FOLLOW-UP:  Post ablation per protocol 	I spent a total of 60 minutes for the patient's care on the date of the service including chart review, history taking, physical exam, counseling, referrals and communication with other providers, and documentation.	Julius M Hibbert, MD  Cardiac Electrophysiology Section  Division of Cardiology}
\vspace{4pt}
Give your response in JSON format using the provided template. The desired response type is provided in angle brackets \verb|< >|. For example, \verb|<int>| means to provide an integer response. Provide a single value response without ranges.
\vspace{4pt}
JSON Response:
\begin{jsoncode}
{
    "Answer": "<int>",
    "Explanation": "<str>"
}
\end{jsoncode}
\tcblower
\begin{jsoncode}
{
    "Answer": "2",
    "Explanation": "The patient's ASA Physical Status Classification appears to be ASA II. According to the ASA classification system, an ASA II patient is defined as a patient with mild systemic disease. In this case, the patient has a history of paroxysmal atrial fibrillation, which is a chronic condition, but he has normal left ventricular function, no valvular heart disease, and is otherwise healthy without significant lifestyle constraints. He reports feeling generally well since the cardioversion and continues his regular activities as an outdoorsman, albeit with caution due to anticoagulation. There are no other significant comorbidities or systemic diseases mentioned in the medical record that would elevate his classification to ASA III or beyond. Meanwhile, his condition is not limited to a specific organ system or seems to have a substantive impact on his daily functioning, so he would not be classified as ASA I. Therefore, ASA II is the most fitting classification."
}
\end{jsoncode}
\end{tcolorbox}

\subsubsection{Zero-shot Q\&A From Notes Summary}\label{suppfig-zeroshot_summary}
\begin{tcolorbox}[enhanced jigsaw,breakable,pad at break*=1mm,
  colback=yellow!5!white,colframe=yellow!50!black,
  colbacktitle=yellow!75!black,title=Prompt \& LLM Output]
\obeylines
You are given a task and context. The context contains information from the proposed procedure and patient's medical record. Assess the patient in the context of the proposed procedure and then provide an answer.
\vspace{8pt}
Task: \hlgreen{What is the patient's ASA Physical Status Classification?}
\vspace{4pt}
Context:
Procedure: \hlyellow{A-FIB ABLATION}
Procedure Description: \hlyellow{A-Fib Ablation (N/A )}
Diagnosis: \hlyellow{Paroxysmal atrial fibrillation (HCC)}
Provider Service: \hlyellow{Cardiovascular}
\vspace{4pt}
Medical Record Summary:
\hlpink{62-year-old male patient Homer Simpson with a history of paroxysmal atrial fibrillation presented for evaluation via telemedicine. He experienced palpitations and an episode requiring hospitalization and cardioversion earlier in the year. Post-cardioversion, he feels generally well with no recurrence of rapid heart action, preserved LV function (LVEF 0.65), and no significant abnormalities on recent TEE. He has continued with anticoagulation therapy using Xarelto. After discussing treatment options, Mr. Simpson elected to proceed with catheter ablation with a scheduled date of August 22, 2020, and plans to continue anticoagulation through ablation and for at least 6 weeks post-procedure.}
\vspace{4pt}
Give your response in JSON format using the provided template. The desired response type is provided in angle brackets \verb|< >|. For example, \verb|<int>| means to provide an integer response. Provide a single value response without ranges.
\vspace{4pt}
JSON Response:
\begin{jsoncode}
{
    "Answer": "<int>",
    "Explanation": "<str>"
}
\end{jsoncode}
\tcblower
\begin{jsoncode}
{
    "Answer": "2",
    "Explanation": "The ASA Physical Status Classification for this patient would be ASA II. The classification ASA II is for patients with mild systemic disease. Even though Mr. Simpson experienced palpitations and required hospitalization and cardioversion, his current state includes preserved left ventricular function, no significant findings on recent transesophageal echocardiogram (TEE), and he is stable on anticoagulation therapy with Xarelto. There is no mention of any other systemic diseases that are poorly controlled. Thus, reflecting a patient with a mild systemic disease without substantive functional limitations fits the ASA II category."
}
\end{jsoncode}
\end{tcolorbox}

\subsubsection{Few-shot Q\&A From Notes Summary}\label{suppfig-fewshot_summary}
\begin{tcolorbox}[enhanced jigsaw,breakable,pad at break*=1mm,
  colback=yellow!5!white,colframe=yellow!50!black,
  colbacktitle=yellow!75!black,title=Prompt \& LLM Output]
\obeylines
You are given examples of task, context, and answer. The context contains information from the proposed procedure and patient's medical record which can be used to determine the answer.
\vspace{8pt}
\hlblue{Task: What is the patient's ASA Physical Status Classification?}
\vspace{4pt}
\hlblue{Context:}
\hlblue{Procedure: HYSTERECTOMY, TOTAL, VAGINAL, WITH VAGINAL VAULT SUSPENSION}
\hlblue{Procedure Description: LAPAROSCOPIC ASSISTED  VAGINAL HYSTERECTOMY REMOVAL BILATERALTUBES AND OVARIES WITH UTEROSACRAL LIGAMENT SUSPENSION AND CYSTOSCOPY.   (N/A Vagina)}
\hlblue{Diagnosis: Uterine prolapse|Cystocele, midline}
\hlblue{Provider Service: Gynecology}
\vspace{4pt}
\hlblue{Medical Record Summary:}
\hlblue{Marge Simpson is a 61-year-old female with a history of uterine prolapse treated with a pessary for 10-15 years, endometrial hyperplasia managed with progesterone cream, bladder infections, pelvic pressure, lipidemia, and thyroid disease. She is scheduled for a LAVH, BSO, USLS with a preoperative Caprini score of 3 indicating a high risk for VTE, for which heparin and SCDs are planned. She takes levothyroxine and liothyronine for thyroid disease and has documented allergies to nitrofurantoin and levofloxacin, causing GI upset. Physical examination reveals a well-nourished woman with stable vital signs and no signs of anxiety or agitation.}
\vspace{4pt}
\hlblue{JSON Response:}
\hlblue{\{}
\hlblue{~~~~"Answer": "2"}
\hlblue{\}}
\vspace{8pt}
\hlblue{Task: What is the patient's ASA Physical Status Classification?}
\vspace{4pt}
\hlblue{Context:}
\hlblue{Procedure: MICROLARYNGOSCOPY, USING LASER, WITH BRONCHOSCOPY}
\hlblue{Procedure Description: MICROLARYNGOSCOPY, DILATIN, STEROID INJECTION, WITH BRONCHOSCOPY (Bronchus)}
\hlblue{Diagnosis: Subglottic stenosis|Dyspnea, unspecified type|Dysphonia}
\hlblue{Provider Service: Otolaryngology}
\vspace{4pt}
\hlblue{Medical Record Summary:}
\hlblue{Selma Bouvier, a 59-year-old female with a long history of subglottic stenosis, has experienced recurrent symptoms of dyspnea, dysphonia, and mucus build-up leading to multiple interventions. Most recently, she underwent a microlaryngoscopy using CO2 laser with bronchoscopy. Postoperative reports indicate significant improvement in breathing and ability to carry out activities with intermittent post-nasal drip and productive cough attributed to allergies. Vital signs are stable, and recent physical exams show slight stridor, no acute distress, and a patent airway with 50-60\% subglottic stenosis. Current plan includes consideration of further dilation and potential steroid injections for long-term management. She is not interested in in-office balloon dilation or open tracheal resection at this time.}
\vspace{4pt}
\hlblue{JSON Response:}
\hlblue{\{}
\hlblue{~~~~"Answer": "3"}
\hlblue{\}}
\vspace{8pt} 
\hlblue{Task: What is the patient's ASA Physical Status Classification?}
\vspace{4pt}
\hlblue{Context:}
\hlblue{Procedure: LAPAROTOMY, EXPLORATORY}
\hlblue{Procedure Description: LAPAROTOMY, EXPLORATORY, PORTAL VEIN REPAIR, TEMPORARY ABDOMINAL CLOSURE (N/A Abdomen)}
\hlblue{Diagnosis: Intraabdominal hemorrhage}
\hlblue{Provider Service: General Surgery}
\vspace{4pt}
\hlblue{Medical Record Summary:}
\hlblue{Patty Bouvier is a 63-year-old G2P2 female with a history of breast cancer treated with tamoxifen, presenting with post-menopausal bleeding (PMB). Initial assessment revealed likely endometrial polyp and a simple paratubal cyst, with a negative endometrial biopsy for hyperplasia and malignancy. Patient subsequently underwent hysteroscopic polypectomy with pathology showing complex endometrial hyperplasia without atypia. A Mirena IUD was inserted for management of the hyperplasia with endometrial sampling planned for follow-up. Patient later requested IUD removal after discontinuing tamoxifen and reports minimal spotting. Advised to have repeat endometrial biopsy in 4-6 months. The patient's medical history also includes spinal fusion, appendectomy, and surgeries due to an accident; family history includes heart disease and stroke. Notable allergies include Lymphazurin, Vancomycin, Dexamethasone, Lamisil, and Pcn. At the age of 67, she experienced intraabdominal hemorrhage, suspected to be related to a pancreatic head mass with adjacent hemorrhage.}
\vspace{4pt}
\hlblue{JSON Response:}
\hlblue{\{}
\hlblue{~~~~"Answer": "5"}
\hlblue{\}}
\vspace{8pt} 
\hlblue{Task: What is the patient's ASA Physical Status Classification?}
\vspace{4pt}
\hlblue{Context:}
\hlblue{Procedure: SURGICAL PROCUREMENT, ORGAN}
\hlblue{Procedure Description: SURGICAL PROCUREMENT, ORGAN / LIVER AND KIDNEYS (N/A Abdomen)
Diagnosis: Other accident}
\hlblue{Provider Service: Organ Donor}
\vspace{4pt}
\hlblue{Medical Record Summary:}
\hlblue{Mr. Doe, a 60-year-old male with unknown past medical history, suffered a severe traumatic brain injury from a fall from a ladder. He had an initial Glasgow Coma Scale of 3, was intubated at the scene, and showed fixed and dilated pupils. CT scan revealed a left-sided acute subdural hematoma with midline shift and a diffuse subarachnoid hemorrhage. He was clinically consistent with brain death upon hospital examination, and after discussion with his wife, was admitted to the trauma ICU with DNR status for medical futility. Organ donation was considered, but the family eventually declined brain death testing. He ultimately underwent a formal brain death examination, resulting in a confirmed brain death diagnosis, and was pronounced deceased.}
\vspace{4pt}
\hlblue{JSON Response:}
\hlblue{\{}
\hlblue{~~~~"Answer": "6"}
\hlblue{\}}
\vspace{8pt}
Task: \hlgreen{What is the patient's ASA Physical Status Classification?}
\vspace{4pt}
Context:
Procedure: \hlyellow{A-FIB ABLATION}
Procedure Description: \hlyellow{A-Fib Ablation (N/A )}
Diagnosis: \hlyellow{Paroxysmal atrial fibrillation (HCC)}
Provider Service: \hlyellow{Cardiovascular}
\vspace{4pt}
Medical Record Summary:
\hlpink{62-year-old male patient Homer Simpson with a history of paroxysmal atrial fibrillation presented for evaluation via telemedicine. He experienced palpitations and an episode requiring hospitalization and cardioversion earlier in the year. Post-cardioversion, he feels generally well with no recurrence of rapid heart action, preserved LV function (LVEF 0.65), and no significant abnormalities on recent TEE. He has continued with anticoagulation therapy using Xarelto. After discussing treatment options, Mr. Simpson elected to proceed with catheter ablation with a scheduled date of August 22, 2020, and plans to continue anticoagulation through ablation and for at least 6 weeks post-procedure.}
\vspace{4pt}
Give your response in JSON format using the provided template. The desired response type is provided in angle brackets \verb|< >|. For example, \verb|<int>| means to provide an integer response. Provide a single value response without ranges.
\vspace{4pt}
JSON Response:
\begin{jsoncode}
{
    "Answer": "<int>",
    "Explanation": "<str>"
}
\end{jsoncode}
\tcblower
\begin{jsoncode}
{
    "Answer": "2",
    "Explanation": "The patient's ASA Physical Status Classification is likely 2, indicating a patient with mild systemic disease. In this case, although the patient has a history of paroxysmal atrial fibrillation and an episode requiring cardioversion, he is currently feeling generally well with no recurrence of symptoms, preserved left ventricular function, and has been stable on anticoagulation therapy."
}
\end{jsoncode}
\end{tcolorbox}

\subsubsection{Zero-shot Chain-of-Thought Q\&A From Original Notes}\label{suppfig-zeroshot_cot_notes}
\begin{tcolorbox}[enhanced jigsaw,breakable,pad at break*=1mm,
  colback=yellow!5!white,colframe=yellow!50!black,
  colbacktitle=yellow!75!black,title=Prompt \& LLM Output]
\obeylines
You are given a task and context. The context contains information from the proposed procedure and patient's medical record. Assess the patient in the context of the proposed procedure and then provide an answer.
\vspace{8pt}
Task: \hlgreen{What is the patient's ASA Physical Status Classification?}
\vspace{4pt}
Context:
Procedure: \hlyellow{A-FIB ABLATION}
Procedure Description: \hlyellow{A-Fib Ablation (N/A )}
Diagnosis: \hlyellow{Paroxysmal atrial fibrillation (HCC)}
Provider Service: \hlyellow{Cardiovascular}
\vspace{4pt}
Medical Record Notes:
\hlred{Progress Notes written by Physician at 2020-07-02 12:23:00:}
\hlpurple{ELECTROPHYSIOLOGY CLINIC INITIAL CONSULT	REASON FOR CONSULT:  Atrial fibrillation 	REFERRING:  Self	PRIMARY CARE: No primary care provider on file.	Distant Site Telemedicine Encounter	I conducted this encounter from Marvin Monroe Memorial Hospital via secure, live, face-to-face video conference with the patient. Homer was located at home with his wife. I reviewed the risks and benefits of telemedicine as pertinent to this visit and the patient agreed to proceed.  	PROBLEM LIST  1. Atrial fibrillation 	HPI:   Mr. Simpson is a 62 year old male referred for evaluation of atrial fibrillation. He has a long standing history of episodic palpitations for several years. After an episode of the same earlier in 2020 he was seen by his PCP who noted that 12 lead ECG showed atrial fibrillation. These episodes were paroxysmal in nature without clear provoking triggers or palliating factors. He had a more sustained episode of palpitations and light headedness on a flight to Springfield where he ultimately was admitted to a hospital for evaluation. ECG and telemetry showed persistent atrial fibrillation and he underwent a TEE cardioversion. He was started on Xarelto and advised to continue for 4 weeks. He had a TTE showing normal LV function and no valvular heart disease. Imaging also showed possible cholecystitis though HIDA scan was most in keeping with biliary dyskinesia. 	Today he joins me via telemedicine. He has felt generally well since his cardioversion and discharge. He has not had a recurrence of rapid heart action. He has continued Xarelto. He is an avid outdoorsman and has been avoiding higher risk physical activity in light of his anticoagulation use. He denies pre-syncope or syncope. He has not had chest pain. He denies peripheral edema, orthopnea or PND. 	ECG June 19, 2020 (personally reviewed) shows atrial fibrillation with IVCD. 	CARDIOLOGY STUDIES	TEE [June 2020]: NO LAA thrombus, LVEF 0.65, trace  TR.   	PAST MEDICAL HISTORY  Patient Active Problem List	Diagnosis Date Noted   • Paroxysmal atrial fibrillation (HCC) [I48.0] 07/02/2020 	Added automatically from request for surgery 127730     	SOCIAL HISTORY:   Married. Non-smoker. Rare EtOH. Works for local tech company. 	FAMILY HISTORY:  No heart failure, SCD.	ALLERGIES:  Patient has no allergy information on record.	Current Outpatient Medications   Medication Sig Dispense Refill   • rivaroxaban (Xarelto) 20 MG tablet Take 1 tablet (20 mg) by mouth daily. 90 tablet 0 	No current facility-administered medications for this visit.    	REVIEW OF SYSTEMS 	Other than HPI, all other systems are negative. 	PHYSICAL EXAM  VITAL SIGNS: There were no vitals taken for this visit.  Appears well   	IMPRESSION:   Mr. Simpson is a 62 year old male with the following salient arrhythmic issues: 	1. Atrial fibrillation 	We discussed atrial fibrillation in detail, including the risks factors, triggers, natural history, stroke risk prediction and treatment options.  Specifically, we discussed a rate control strategy with medications versus a rhythm control strategy.  For many patients, rate control is sufficient to maintain quality of life and cardiovascular function.  We discussed the antiarrhythmic drugs in detail, including class Ic and class III AADs.  We also discussed the importance of lifestyle changes in improving both ablative and non-ablative approaches to AF, namely weight reduction and exercise.	Given his clearly symptomatic atrial fibrillation we discussed the rationale for rhythm control. Given his low resting heart rate and background significant physical activity, we discussed that AAD options are not likely to be tolerated or less well tolerated. 	I therefore discussed ablation.  I explained that the chances of success with this procedure is about 75\% but may need more than one procedure.  I explained how a CT or MRI  of his heart is completed ahead of time to assess his pulmonary vein anatomy.  The procedure is completed under general anesthesia, possibly with a transesophageal echocardiogram.  I explained that a transseptal puncture is completed and a wide-area circumferential ablation around the pulmonary veins is completed.  I explained that he may go home the same day or may spend one night in the hospital and go home the next day.  I said the risks of the procedure included, but were not limited to bleeding, infection, cardiac perforation and tamponade, 1 in 200 chance of a stroke, gastroparesis, pulmonary vein stenosis, atrial-esophageal fistula, and 1 in a 1000 chance of death.	After discussion, he would like to proceed to catheter ablation. 	PLAN:  1. AF ablation August 22, 2020  2. Cardiac CT for PV anatomy   3. Continue Xarelto through ablation and minimum 6 weeks after 	FOLLOW-UP:  Post ablation per protocol 	I spent a total of 60 minutes for the patient's care on the date of the service including chart review, history taking, physical exam, counseling, referrals and communication with other providers, and documentation.	Julius M Hibbert, MD  Cardiac Electrophysiology Section  Division of Cardiology}
\vspace{4pt}
\hlbrown{Think step by step} and give your response in JSON format using the provided template. The desired response type is provided in angle brackets \verb|< >|. For example, \verb|<int>| means to provide an integer response. Provide a single value response without ranges.
\vspace{4pt}
JSON Response:
\begin{jsoncode}
{
    "Step By Step Explanation": "<str>",
    "Answer": "<int>"
}
\end{jsoncode}
\tcblower
\begin{jsoncode}
{
    "Step By Step Explanation": "The ASA Physical Status Classification system is used to assess the preoperative health of patients before they undergo a procedure. It ranges from ASA I, a normal healthy patient, to ASA VI, a brain-dead patient awaiting organ removal. Based on the information provided, Mr. Simpson has a history of paroxysmal atrial fibrillation, which qualifies as at least a systemic disease without substantive functional limitations (ASA II). He is not described as having a severe systemic disease (ASA III) since he does not have a constant threat to life, nor is he incapacitated (ASA IV). There is no mention of emergency surgery (ASA E) or organ donation (ASA VI). Additionally, he has normal LV function, no valvular heart disease, and has been described as feeling generally well since his cardioversion, with no other systemic diseases noted, and is not incapacitated. Even though he has been started on Xarelto (anticoagulant therapy) due to A-Fib, this does not automatically increase his ASA status if he is otherwise healthy. Therefore, he seems to fit the description of ASA II.",
    "Answer": "2"
}
\end{jsoncode}
\end{tcolorbox}

\subsubsection{Zero-shot Chain-of-Thought Q\&A From Notes Summary}\label{suppfig-zeroshot_cot_summary}
\begin{tcolorbox}[enhanced jigsaw,breakable,pad at break*=1mm,
  colback=yellow!5!white,colframe=yellow!50!black,
  colbacktitle=yellow!75!black,title=Prompt \& LLM Output]
\obeylines
You are given a task and context. The context contains information from the proposed procedure and patient's medical record. Assess the patient in the context of the proposed procedure and then provide an answer.
\vspace{8pt}
Task: \hlgreen{What is the patient's ASA Physical Status Classification?}
\vspace{4pt}
Context:
Procedure: \hlyellow{A-FIB ABLATION}
Procedure Description: \hlyellow{A-Fib Ablation (N/A )}
Diagnosis: \hlyellow{Paroxysmal atrial fibrillation (HCC)}
Provider Service: \hlyellow{Cardiovascular}
\vspace{4pt}
Medical Record Summary:
\hlpink{62-year-old male patient Homer Simpson with a history of paroxysmal atrial fibrillation presented for evaluation via telemedicine. He experienced palpitations and an episode requiring hospitalization and cardioversion earlier in the year. Post-cardioversion, he feels generally well with no recurrence of rapid heart action, preserved LV function (LVEF 0.65), and no significant abnormalities on recent TEE. He has continued with anticoagulation therapy using Xarelto. After discussing treatment options, Mr. Simpson elected to proceed with catheter ablation with a scheduled date of August 22, 2020, and plans to continue anticoagulation through ablation and for at least 6 weeks post-procedure.}
\vspace{4pt}
\hlbrown{Think step by step} and give your response in JSON format using the provided template. The desired response type is provided in angle brackets \verb|< >|. For example, \verb|<int>| means to provide an integer response. Provide a single value response without ranges.
\vspace{4pt}
JSON Response:
\begin{jsoncode}
{
    "Step By Step Explanation": "<str>",
    "Answer": "<int>"
}
\end{jsoncode}
\tcblower
\begin{jsoncode}
{
    "Step By Step Explanation": "The ASA Physical Status Classification system is used to assess and communicate a patient's pre-anesthesia medical comorbidities. The classifications are as follows: ASA 1 - Normal healthy patient, ASA 2 - Patients with mild systemic disease, ASA 3 - Patients with severe systemic disease, ASA 4 - Patients with severe systemic disease that is a constant threat to life, ASA 5 - Moribund patients who are not expected to survive without the operation. In the case of Homer Simpson, a 62-year-old male with paroxysmal atrial fibrillation, who experienced an episode requiring hospitalization and cardioversion but otherwise feels well with preserved cardiac function (LVEF 0.65) and is on anticoagulation therapy, he would be classified as ASA 2. This is due to his atrial fibrillation being a mild systemic disease; he is not otherwise healthy (ASA 1), yet his disease does not constitute a severe systemic disease (ASA 3), and it is not a constant threat to his life (ASA 4) nor is he at the brink of death (ASA 5).",
    "Answer": "2"
}
\end{jsoncode}
\end{tcolorbox}

\subsubsection{Few-shot Chain-of-Thought Q\&A From Notes Summary}\label{suppfig-fewshot_cot_summary}
\begin{tcolorbox}[enhanced jigsaw,breakable,pad at break*=1mm,
  colback=yellow!5!white,colframe=yellow!50!black,
  colbacktitle=yellow!75!black,title=Prompt \& LLM Output]
\obeylines
\hlblue{You are given examples of task, context, logical reasoning, and answer. The context contains information from the proposed procedure and patient's medical record which can be used to determine the answer. The logical reasoning contains a step by step explanation leading to the answer.}
\vspace{8pt}
\hlblue{Task: What is the patient's ASA Physical Status Classification?}
\vspace{4pt}
\hlblue{Context:}
\hlblue{Procedure: HYSTERECTOMY, TOTAL, VAGINAL, WITH VAGINAL VAULT SUSPENSION}
\hlblue{Procedure Description: LAPAROSCOPIC ASSISTED  VAGINAL HYSTERECTOMY REMOVAL BILATERALTUBES AND OVARIES WITH UTEROSACRAL LIGAMENT SUSPENSION AND CYSTOSCOPY.   (N/A Vagina)}
\hlblue{Diagnosis: Uterine prolapse|Cystocele, midline}
\hlblue{Provider Service: Gynecology}
\vspace{4pt}
\hlblue{Medical Record Summary:}
\hlblue{Marge Simpson is a 61-year-old female with a history of uterine prolapse treated with a pessary for 10-15 years, endometrial hyperplasia managed with progesterone cream, bladder infections, pelvic pressure, lipidemia, and thyroid disease. She is scheduled for a LAVH, BSO, USLS with a preoperative Caprini score of 3 indicating a high risk for VTE, for which heparin and SCDs are planned. She takes levothyroxine and liothyronine for thyroid disease and has documented allergies to nitrofurantoin and levofloxacin, causing GI upset. Physical examination reveals a well-nourished woman with stable vital signs and no signs of anxiety or agitation.}
\vspace{4pt}
\hlblue{JSON Response:}
\hlblue{\{}
\hlorange{~~~~"Step By Step Explanation": "The ASA Physical Status Classification System is a system used by anesthesiologists to assess the fitness of patients before surgery. It is important because it helps predict the risk of complications during or after surgery. The ASA physical status classification is as follows: }
\hlorange{1. ASA 1: A normal healthy patient.}
\hlorange{2. ASA 2: A patient with mild systemic disease.}
\hlorange{3. ASA 3: A patient with severe systemic disease.}
\hlorange{4. ASA 4: A patient with severe systemic disease that is a constant threat to life.}
\hlorange{5. ASA 5: A moribund patient who is not expected to survive without the operation.}
\hlorange{6. ASA 6: A declared brain-dead patient whose organs are being removed for donor purposes.}
\hlorange{~~Given the information from the patient's medical record, Marge Simpson has systemic diseases that are well controlled. She has endometrial hyperplasia managed with progesterone cream, bladder infections, pelvic pressure, thyroid disease for which she is taking medications, and lipidemia. While she has multiple medical issues, none are described as 'severe.' Her vital signs are stable, and there is no indication in the provided information that her systemic diseases are a constant threat to her life or a severe systemic disease. Therefore, she does not fit the criteria for ASA 3 or higher. }
\hlorange{~~Additionally, the patient does not qualify for ASA 1, as she does have systemic diseases, albeit mild. The patient's preoperative Caprini score of 3 indicates a high risk for venous thromboembolism (VTE), which is being managed with prophylactic measures such as heparin and sequential compression devices (SCDs). An ASA 2 status is typically assigned to patients who have a mild systemic disease that does not limit their activity. }
\hlorange{Based on her controlled medical conditions and her physical examination, which does not reveal signs of severe systemic disease, the correct ASA Physical Status Classification for Marge Simpson is ASA 2.",}
\hlblue{~~~~"Answer": "2"}
\hlblue{\}}
\vspace{8pt}
\hlblue{Task: What is the patient's ASA Physical Status Classification?}
\vspace{4pt}
\hlblue{Context:}
\hlblue{Procedure: MICROLARYNGOSCOPY, USING LASER, WITH BRONCHOSCOPY}
\hlblue{Procedure Description: MICROLARYNGOSCOPY, DILATIN, STEROID INJECTION, WITH BRONCHOSCOPY (Bronchus)}
\hlblue{Diagnosis: Subglottic stenosis|Dyspnea, unspecified type|Dysphonia}
\hlblue{Provider Service: Otolaryngology}
\vspace{4pt}
\hlblue{Medical Record Summary:}
\hlblue{Selma Bouvier, a 59-year-old female with a long history of subglottic stenosis, has experienced recurrent symptoms of dyspnea, dysphonia, and mucus build-up leading to multiple interventions. Most recently, she underwent a microlaryngoscopy using CO2 laser with bronchoscopy. Postoperative reports indicate significant improvement in breathing and ability to carry out activities with intermittent post-nasal drip and productive cough attributed to allergies. Vital signs are stable, and recent physical exams show slight stridor, no acute distress, and a patent airway with 50-60\% subglottic stenosis. Current plan includes consideration of further dilation and potential steroid injections for long-term management. She is not interested in in-office balloon dilation or open tracheal resection at this time.}
\vspace{4pt}
\hlblue{JSON Response:}
\hlblue{\{}
\hlorange{~~~~"Step By Step Explanation": "The ASA Physical Status Classification System is used to assess and communicate a patient's pre-anesthesia medical comorbidities. This classification ranges from ASA 1 to ASA 6. The ASA 1 indicates a normal healthy patient. ASA 2 indicates a patient with a mild systemic disease. ASA 3 is assigned to a patient with severe systemic disease. ASA 4 is for a patient with a severe systemic disease that is a constant threat to life. ASA 5 indicates a moribund patient who is not expected to survive without the operation. ASA 6 is declared for a patient who is declared brain-dead and whose organs are being removed for donor purposes. Selma Bouvier's classification as ASA 3 likely stems from several factors mentioned in her medical record: }
\hlorange{1. She has a 'long history of subglottic stenosis,' which is a chronic condition that affects the airway. This is a significant piece of medical history indicative of a severe systemic disease that has required multiple interventions. }
\hlorange{2. Ms. Bouvier had 'recurrent symptoms of dyspnea and dysphonia,' symptoms that can limit function and indicate that her systemic disease significantly impacts her health. }
\hlorange{3. Ms. Bouvier's recent physical exam showed 'slight stridor,' and a 'patent airway with 50-60\% subglottic stenosis,' which means that despite previous interventions, there is still notable impairment which could be considered severe. }
\hlorange{~~These points suggest that Ms. Bouvier's medical condition significantly impacts her daily life, thereby fitting the criteria for an ASA 3 classification. Ms. Bouvier does not appear to be at a constant threat to life nor moribund, which would be required for an ASA 4 or ASA 5 classification, respectively.",}
\hlblue{~~~~"Answer": "3"}
\}
\vspace{8pt}
\hlblue{Task: What is the patient's ASA Physical Status Classification?}
\vspace{4pt}
\hlblue{Context:}
\hlblue{Procedure: LAPAROTOMY, EXPLORATORY}
\hlblue{Procedure Description: LAPAROTOMY, EXPLORATORY, PORTAL VEIN REPAIR, TEMPORARY ABDOMINAL CLOSURE (N/A Abdomen)}
\hlblue{Diagnosis: Intraabdominal hemorrhage}
\hlblue{Provider Service: General Surgery}
\vspace{4pt}
Medical Record Summary:
\hlblue{Patty Bouvier is a 63-year-old G2P2 female with a history of breast cancer treated with tamoxifen, presenting with post-menopausal bleeding (PMB). Initial assessment revealed likely endometrial polyp and a simple paratubal cyst, with a negative endometrial biopsy for hyperplasia and malignancy. Patient subsequently underwent hysteroscopic polypectomy with pathology showing complex endometrial hyperplasia without atypia. A Mirena IUD was inserted for management of the hyperplasia with endometrial sampling planned for follow-up. Patient later requested IUD removal after discontinuing tamoxifen and reports minimal spotting. Advised to have repeat endometrial biopsy in 4-6 months. The patient's medical history also includes spinal fusion, appendectomy, and surgeries due to an accident; family history includes heart disease and stroke. Notable allergies include Lymphazurin, Vancomycin, Dexamethasone, Lamisil, and Pcn. At the age of 67, she experienced intraabdominal hemorrhage, suspected to be related to a pancreatic head mass with adjacent hemorrhage.}
\vspace{4pt}
\hlblue{JSON Response:}
\hlblue{\{}
\hlorange{~~~~"Step By Step Explanation": "To determine Patty Bouvier's American Society of Anesthesiologists (ASA) Physical Status Classification, we assess the information given about her health status and the urgency and severity of her medical condition. The ASA classification is a system used to evaluate the fitness of patients before surgery. The scale ranges from 1 (a healthy patient) to 6 (a declared brain-dead patient whose organs are being removed for donor purposes). The ASA Physical Status Classification defines class 5 as 'a moribund patient who is not expected to survive without the operation.' This classification applies to patients facing an immediate threat to life. In this context, Ms. Bouvier has suffered an intraabdominal hemorrhage, which is a life-threatening condition. The procedure planned is an exploratory laparotomy with portal vein repair and temporary abdominal closure, which indicates imminent surgery to control hemorrhage and is likely an emergency situation. Given this information and considering that intraabdominal hemorrhage puts Ms. Bouvier's life in immediate danger, the correct classification would be ASA 5. The other medical history such as breast cancer, endometrial hyperplasia, and previous surgeries also corroborate a more complicated medical background which further supports the assignment of a higher ASA class.",}
\hlblue{~~~~"Answer": "5"}
\hlblue{\}}
\vspace{8pt}
\hlblue{Task: What is the patient's ASA Physical Status Classification?}
\vspace{4pt}
\hlblue{Context:}
\hlblue{Procedure: SURGICAL PROCUREMENT, ORGAN}
\hlblue{Procedure Description: SURGICAL PROCUREMENT, ORGAN / LIVER AND KIDNEYS (N/A Abdomen)}
\hlblue{Diagnosis: Other accident}
\hlblue{Provider Service: Organ Donor}
\vspace{4pt}
\hlblue{Medical Record Summary:}
\hlblue{Mr. Doe, a 60-year-old male with unknown past medical history, suffered a severe traumatic brain injury from a fall from a ladder. He had an initial Glasgow Coma Scale of 3, was intubated at the scene, and showed fixed and dilated pupils. CT scan revealed a left-sided acute subdural hematoma with midline shift and a diffuse subarachnoid hemorrhage. He was clinically consistent with brain death upon hospital examination, and after discussion with his wife, was admitted to the trauma ICU with DNR status for medical futility. Organ donation was considered, but the family eventually declined brain death testing. He ultimately underwent a formal brain death examination, resulting in a confirmed brain death diagnosis, and was pronounced deceased.}
\vspace{4pt}
\hlblue{JSON Response:}
\hlblue{\{}
\hlorange{~~~~"Step By Step Explanation": "The ASA Physical Status Classification System is a way for anesthesiologists to categorize a patient's preoperative health. This system ranges from ASA 1, denoting a normal healthy patient, to ASA 6, indicating a declared brain-dead patient whose organs are being removed for donor purposes. In this particular case, we can determine the ASA classification by following these logical steps: }
\hlorange{1. Identify the patient's current medical status from the medical record summary provided. Mr. Doe has been confirmed to be brain-dead following a severe traumatic brain injury and a conclusive formal brain death examination. }
\hlorange{2. Recognize the type of procedure described. The text mentions 'SURGICAL PROCUREMENT, ORGAN / LIVER AND KIDNEYS' which indicates organ retrieval for transplantation purposes. }
\hlorange{3. Understand that Mr. Doe is designated as an organ donor ('Provider Service: Organ Donor'), and the procedure is taking place posthumously after he was pronounced deceased. }
\hlorange{4. Recall that the ASA classification of '6' corresponds specifically to patients who are declared brain-dead and are organ donors, according to the definition of ASA 6. }
\hlorange{5. Conclude that the correct ASA Physical Status Classification for Mr. Doe is ASA 6, based on his brain-dead status and the fact that his organs are being procured for donation. This aligns with the provided answer.",}
\hlblue{~~~~"Answer": "6"}
\hlblue{\}}
\vspace{8pt}
Task: \hlgreen{What is the patient's ASA Physical Status Classification?}
\vspace{4pt}
Context:
Procedure: \hlyellow{A-FIB ABLATION}
Procedure Description: \hlyellow{A-Fib Ablation (N/A )}
Diagnosis: \hlyellow{Paroxysmal atrial fibrillation (HCC)}
Provider Service: \hlyellow{Cardiovascular}
\vspace{4pt}
Medical Record Summary:
\hlpink{62-year-old male patient Homer Simpson with a history of paroxysmal atrial fibrillation presented for evaluation via telemedicine. He experienced palpitations and an episode requiring hospitalization and cardioversion earlier in the year. Post-cardioversion, he feels generally well with no recurrence of rapid heart action, preserved LV function (LVEF 0.65), and no significant abnormalities on recent TEE. He has continued with anticoagulation therapy using Xarelto. After discussing treatment options, Mr. Simpson elected to proceed with catheter ablation with a scheduled date of August 22, 2020, and plans to continue anticoagulation through ablation and for at least 6 weeks post-procedure.}
\vspace{4pt}
\hlorange{Think step by step} and give your response in JSON format using the provided template. The desired response type is provided in angle brackets \verb|< >|. For example, \verb|<int>| means to provide an integer response. Provide a single value response without ranges.
\vspace{4pt}
JSON Response:
\begin{jsoncode}
{
    "Step By Step Explanation": "<str>",
    "Answer": "<int>"
}
\end{jsoncode}
\tcblower
\begin{jsoncode}
{
    "Step By Step Explanation": "The ASA Physical Status Classification System is a way to assess and communicate a patient's pre-anesthesia medical co-morbidities. The categories are as follows: \n\n- ASA 1: A normal healthy patient.\n- ASA 2: A patient with mild systemic disease.\n- ASA 3: A patient with severe systemic disease.\n- ASA 4: A patient with severe systemic disease that is a constant threat to life.\n- ASA 5: A moribund patient who is not expected to survive without the operation.\n- ASA 6: A declared brain-dead patient whose organs are being removed for donor purposes.\n\nHomer Simpson is a 62-year-old male with a history of paroxysmal atrial fibrillation. This condition is considered a systemic disease which can affect multiple systems in the body; however, it is noted that his left ventricular ejection fraction (LVEF) is preserved at 0.65 and there are no significant abnormalities on recent transesophageal echocardiogram (TEE). Additionally, there has been no recurrence of rapid heart action since cardioversion, and he has been managing his condition with anticoagulation therapy.\n\nAtrial fibrillation by itself can vary in severity, but given that the patient feels generally well, has preserved LV function, and is currently stable on medication, it indicates a 'mild systemic disease'. No other systemic diseases or significant health issues are mentioned. The patient's ability to actively participate in telemedicine consultation and make an informed treatment choice further suggests that his systemic disease does not incapacitate him.\n\nAccording to the details provided, he does not have a severe systemic disease that dramatically limits his activity or poses a constant threat to life, which excludes classifications ASA 3 and ASA 4. Therefore, the correct ASA classification for Mr. Simpson would be ASA 2. This classification reflects a patient who has a mild systemic disease without substantive functional limitations.",
    "Answer": "2"
}
\end{jsoncode}
\end{tcolorbox}

\subsubsection{Summary Generation}\label{suppfig-summary_generation}
\begin{tcolorbox}[enhanced jigsaw,breakable,pad at break*=1mm,
  colback=blue!5!white,colframe=blue!50!black,
  colbacktitle=blue!75!black,title=Prompt \& LLM Output]
\obeylines
You are given information from the patient's medical record. Summarize this information, making sure to include the most important positive clinical findings.
\vspace{8pt}
\hlred{Progress Notes written by Physician at 2020-07-02 12:23:00:}
\hlpurple{ELECTROPHYSIOLOGY CLINIC INITIAL CONSULT	REASON FOR CONSULT:  Atrial fibrillation 	REFERRING:  Self	PRIMARY CARE: No primary care provider on file.	Distant Site Telemedicine Encounter	I conducted this encounter from Marvin Monroe Memorial Hospital via secure, live, face-to-face video conference with the patient. Homer was located at home with his wife. I reviewed the risks and benefits of telemedicine as pertinent to this visit and the patient agreed to proceed.  	PROBLEM LIST  1. Atrial fibrillation 	HPI:   Mr. Simpson is a 62 year old male referred for evaluation of atrial fibrillation. He has a long standing history of episodic palpitations for several years. After an episode of the same earlier in 2020 he was seen by his PCP who noted that 12 lead ECG showed atrial fibrillation. These episodes were paroxysmal in nature without clear provoking triggers or palliating factors. He had a more sustained episode of palpitations and light headedness on a flight to Springfield where he ultimately was admitted to a hospital for evaluation. ECG and telemetry showed persistent atrial fibrillation and he underwent a TEE cardioversion. He was started on Xarelto and advised to continue for 4 weeks. He had a TTE showing normal LV function and no valvular heart disease. Imaging also showed possible cholecystitis though HIDA scan was most in keeping with biliary dyskinesia. 	Today he joins me via telemedicine. He has felt generally well since his cardioversion and discharge. He has not had a recurrence of rapid heart action. He has continued Xarelto. He is an avid outdoorsman and has been avoiding higher risk physical activity in light of his anticoagulation use. He denies pre-syncope or syncope. He has not had chest pain. He denies peripheral edema, orthopnea or PND. 	ECG June 19, 2020 (personally reviewed) shows atrial fibrillation with IVCD. 	CARDIOLOGY STUDIES	TEE [June 2020]: NO LAA thrombus, LVEF 0.65, trace  TR.   	PAST MEDICAL HISTORY  Patient Active Problem List	Diagnosis Date Noted   • Paroxysmal atrial fibrillation (HCC) [I48.0] 07/02/2020 	Added automatically from request for surgery 127730     	SOCIAL HISTORY:   Married. Non-smoker. Rare EtOH. Works for local tech company. 	FAMILY HISTORY:  No heart failure, SCD.	ALLERGIES:  Patient has no allergy information on record.	Current Outpatient Medications   Medication Sig Dispense Refill   • rivaroxaban (Xarelto) 20 MG tablet Take 1 tablet (20 mg) by mouth daily. 90 tablet 0 	No current facility-administered medications for this visit.    	REVIEW OF SYSTEMS 	Other than HPI, all other systems are negative. 	PHYSICAL EXAM  VITAL SIGNS: There were no vitals taken for this visit.  Appears well   	IMPRESSION:   Mr. Simpson is a 62 year old male with the following salient arrhythmic issues: 	1. Atrial fibrillation 	We discussed atrial fibrillation in detail, including the risks factors, triggers, natural history, stroke risk prediction and treatment options.  Specifically, we discussed a rate control strategy with medications versus a rhythm control strategy.  For many patients, rate control is sufficient to maintain quality of life and cardiovascular function.  We discussed the antiarrhythmic drugs in detail, including class Ic and class III AADs.  We also discussed the importance of lifestyle changes in improving both ablative and non-ablative approaches to AF, namely weight reduction and exercise.	Given his clearly symptomatic atrial fibrillation we discussed the rationale for rhythm control. Given his low resting heart rate and background significant physical activity, we discussed that AAD options are not likely to be tolerated or less well tolerated. 	I therefore discussed ablation.  I explained that the chances of success with this procedure is about 75\% but may need more than one procedure.  I explained how a CT or MRI  of his heart is completed ahead of time to assess his pulmonary vein anatomy.  The procedure is completed under general anesthesia, possibly with a transesophageal echocardiogram.  I explained that a transseptal puncture is completed and a wide-area circumferential ablation around the pulmonary veins is completed.  I explained that he may go home the same day or may spend one night in the hospital and go home the next day.  I said the risks of the procedure included, but were not limited to bleeding, infection, cardiac perforation and tamponade, 1 in 200 chance of a stroke, gastroparesis, pulmonary vein stenosis, atrial-esophageal fistula, and 1 in a 1000 chance of death.	After discussion, he would like to proceed to catheter ablation. 	PLAN:  1. AF ablation August 22, 2020  2. Cardiac CT for PV anatomy   3. Continue Xarelto through ablation and minimum 6 weeks after 	FOLLOW-UP:  Post ablation per protocol 	I spent a total of 60 minutes for the patient's care on the date of the service including chart review, history taking, physical exam, counseling, referrals and communication with other providers, and documentation.	Julius M Hibbert, MD  Cardiac Electrophysiology Section  Division of Cardiology}
\vspace{4pt}
Give your response in JSON format using the provided template. The desired response type is provided in angle brackets \verb|< >|. For example, \verb|<int>| means to provide an integer response. Provide a single value response without ranges.
\vspace{4pt}
JSON Response:
\begin{jsoncode}
{
    "Brief Patient Summary": "<str>"
}
\end{jsoncode}
\tcblower
\begin{jsoncode}
{
    "Brief Patient Summary": "62-year-old male patient Homer Simpson with a history of paroxysmal atrial fibrillation presented for evaluation via telemedicine. He experienced palpitations and an episode requiring hospitalization and cardioversion earlier in the year. Post-cardioversion, he feels generally well with no recurrence of rapid heart action, preserved LV function (LVEF 0.65), and no significant abnormalities on recent TEE. He has continued with anticoagulation therapy using Xarelto. After discussing treatment options, Mr. Simpson elected to proceed with catheter ablation with a scheduled date of August 22, 2020, and plans to continue anticoagulation through ablation and for at least 6 weeks post-procedure."
}
\end{jsoncode}
\end{tcolorbox}

\subsubsection{Chain-of-thought Rationale Generation}\label{suppfig-cot_rationale_generation}
\begin{tcolorbox}[enhanced jigsaw,breakable,pad at break*=1mm,
  colback=orange!5!white,colframe=orange!50!black,
  colbacktitle=orange!75!black,title=Prompt \& LLM Output]
\obeylines
You are given a task, the answer, and context which contains information from the proposed procedure and patient's medical record. Provide the logical reasoning steps that lead to the answer using information from the proposed procedure and patient's medical record.
\vspace{8pt}
Task: \hlgreen{What is the patient's ASA Physical Status Classification?}
Answer: \hlgreen{5}
\vspace{4pt}
Context:
Procedure: \hlyellow{REPAIR, ANEURYSM, AORTA, THORACIC, ASCENDING}
Procedure Description: \hlyellow{REPAIR, ANEURYSM, AORTA, THORACIC, ASCENDING (Chest)}
Diagnosis: \hlyellow{Dissection of ascending aorta (HCC)}
Provider Service: \hlyellow{Cardiovascular Surgery}
\vspace{4pt}
Medical Record Summary:
\hlpink{Barney Gumble is a 56-year-old male with a history of alcohol use disorder who suffered a ground-level fall resulting in a right frontal lobe contusion, subarachnoid hemorrhage, and a Type A aortic dissection. He was found with elevated blood alcohol level, hypotension, and altered mental status necessitating intubation. Imaging revealed aneurysmal dilatation of the aortic root and dissection involving major arteries. He was admitted to ICU for stabilization and impulse control with IV antihypertensives, and was evaluated for potential surgical intervention for the aortic dissection. The patient showed neurological improvement with response to commands while sedated, and had stable vitals without evidence of malperfusion or ischemic complications from the dissection at the time. There is an incidental finding of a right upper lobe nodule and small bowel intussusception.}
\vspace{4pt}
JSON Response:
\begin{jsoncode}
{
    "Step By Step Explanation": "<str>"
}
\end{jsoncode}
\tcblower
\begin{jsoncode}
{
    "Step By Step Explanation": "The ASA Physical Status Classification System classifies a patient's preoperative health to help assess anesthetic risk before surgery. The system ranges from ASA 1 (a normal healthy patient) to ASA 6 (a brain-dead patient whose organs are being removed for donor purposes). ASA 5 is classified as a moribund patient who is not expected to survive without the operation. Here are the logical reasoning steps that lead to the answer, ASA 5, for Barney Gumble based on the information provided in the medical record summary:\n\nStep 1: Identify the patient's medical condition. The patient has suffered a Type A aortic dissection, which is a life-threatening condition that can lead to death if not treated promptly and effectively.\n\nStep 2: Assess the urgency and severity of the underlying condition. A Type A aortic dissection involves the ascending aorta and is often considered a surgical emergency due to the risk of rapid deterioration and fatal outcomes.\n\nStep 3: Evaluate the patient's overall health and comorbidities. Barney has a history of alcohol use disorder. He has also suffered a fall resulting in a right frontal lobe contusion and a subarachnoid hemorrhage. These additional health issues add to his operative risk.\n\nStep 4: Consider the patient's current status. At the time of evaluation, Barney had been admitted to ICU for stabilization, was intubated due to altered mental status which further suggests severe systemic disease and an unstable condition.\n\nStep 5: Determine if the patient is expected to survive without the operation. Given the life-threatening nature of a Type A aortic dissection and the fact that Barney was in the ICU for impulse control with IV antihypertensives, it can be inferred that without the surgery, Barney's chance of survival is poor.\n\nStep 6: Confirm the lack of less acute findings that could suggest a less critical state. The incidental finding of a right upper lobe nodule and small bowel intussusception do not mitigate the acute and critical nature of the patient's condition.\n\nStep 7: Assign the appropriate ASA classification. Based on Barney's critical state, need for urgent surgery, life-threatening diagnosis, and the expectation that he would not survive without the operation, he fits the criteria for ASA 5. Therefore, the assignment of ASA 5 is justified."
}
\end{jsoncode}
\end{tcolorbox}

\subsection{Supplemental Figure 2: Dataset Creation Flow Diagram}\label{suppfig2-section}
\begin{figure}[H]
    \centering
    \includegraphics[width=1\linewidth]{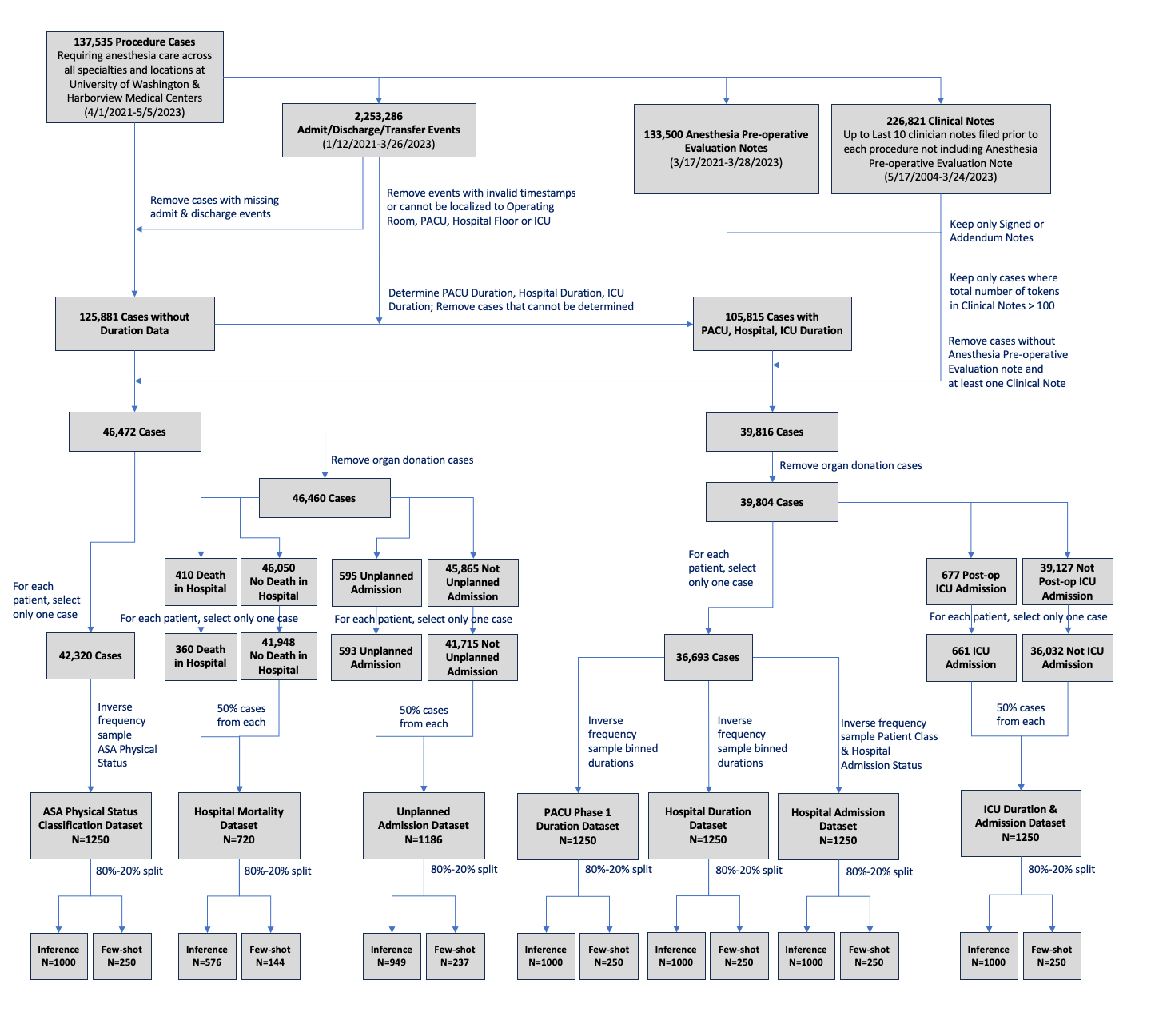}
    \caption{Flow diagram showing how the task-specific datasets were constructed from Electronic Health Record data. The natural occurrence of certain outcomes such as ICU admission, unplanned admission, and hospital mortality are rare, so datasets are constructed to balance the task label. If patients have multiple procedure cases, only a single case for that patient was included in the final dataset. “Clinical Notes” refers to up to the last 10 clinician-written notes filed prior to each procedure, excluding notes directly associated with the procedure itself. Due to the rarity of ICU admission, the datasets for ICU Duration and Admission tasks are identical.}
    \label{fig:suppfig2-flowdiagram}
\end{figure}

\subsection{Supplemental Figure 3: Dataset Case Overlap}\label{suppfig3-section}
\begin{figure}[H]
    \centering
    \includegraphics[width=1\linewidth]{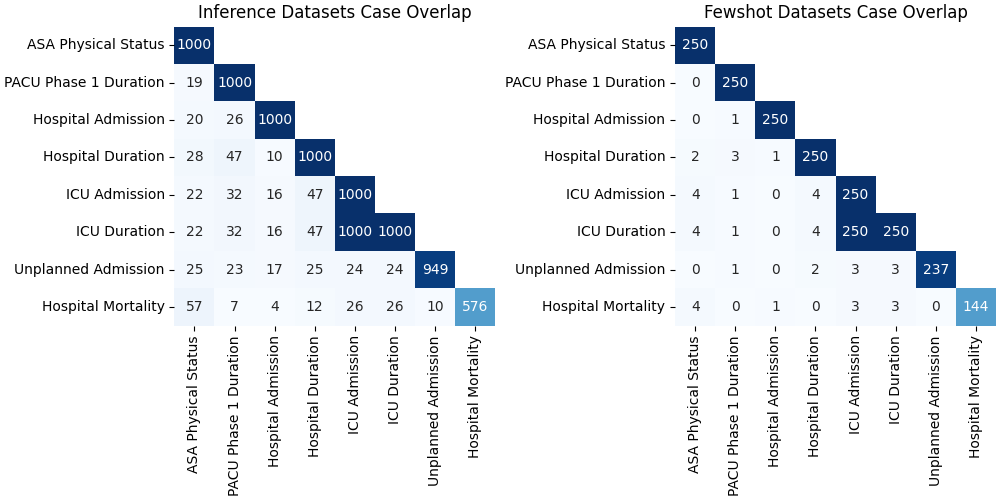}
    \caption{Number of cases that overlap between any of the task-specific datasets in the inference and few-shot data splits. Datasets for ICU admission and ICU duration prediction tasks are identical. Most datasets have \texttt{<}5\% overlap in cases.}
    \label{fig:suppfig3-datasetoverlap}
\end{figure}

\subsection{Supplemental Figure 4: ASA Physical Status - Statistical Significance}\label{suppfig4-section}
\begin{figure}[H]
    \centering
    \includegraphics[width=1\linewidth]{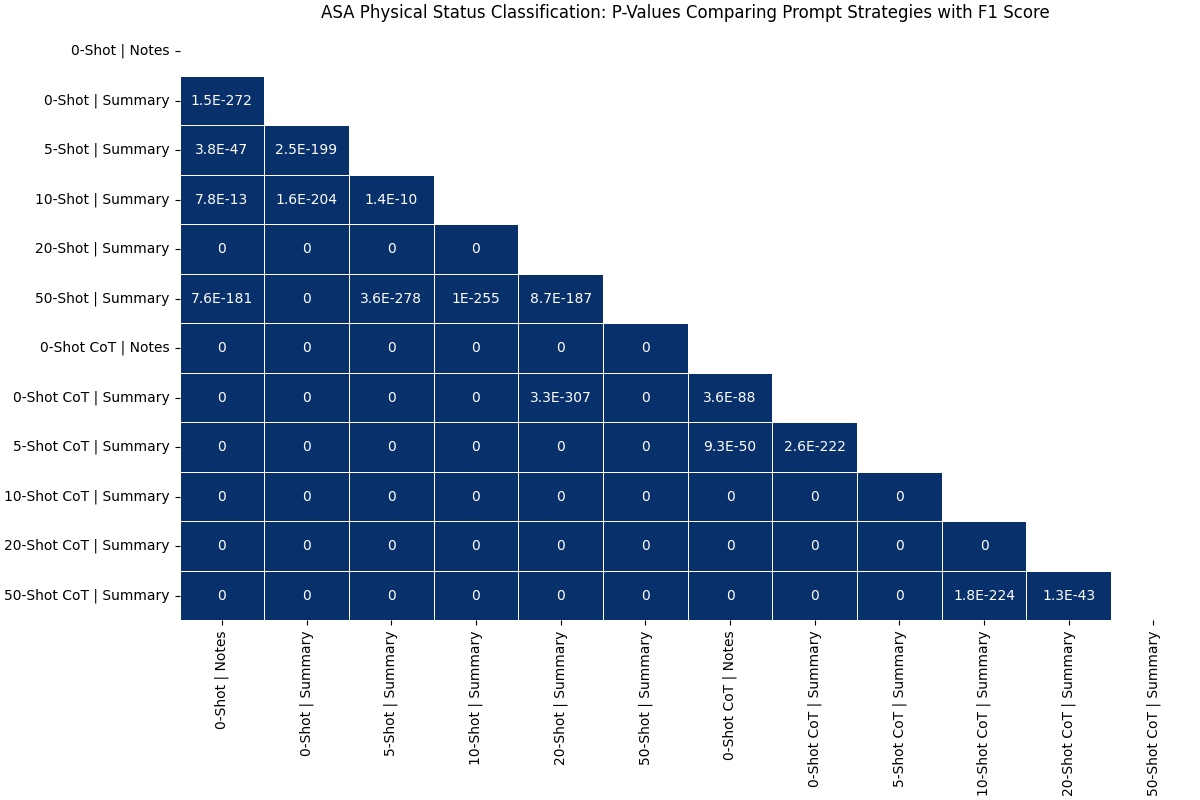}
    \caption{P-values comparing prompt strategies with F1 score for ASA physical status classification prediction. P-values are computed using Wilcoxon signed-rank test with Benjamini-Hochberg procedure for false-discovery rate control level $\alpha=0.05$. Statistically significant p-values \texttt{<} 0.05 are depicted in cells with a dark blue background.}
    \label{fig:suppfig4-asa_pvalues}
\end{figure}

\subsection{Supplemental Figure 5: Hospital Admission - Statistical Significance}\label{suppfig5-section}
\begin{figure}[H]
    \centering
    \includegraphics[width=1\linewidth]{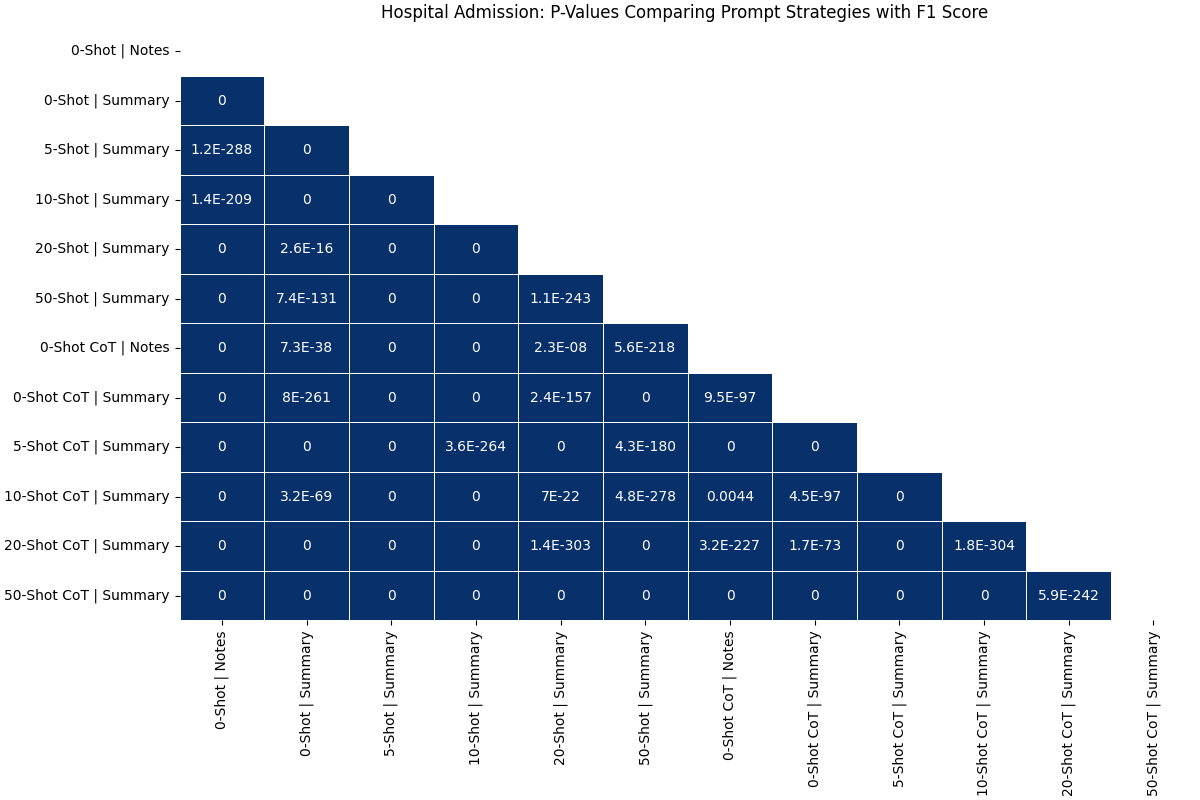}
    \caption{P-values comparing prompt strategies with F1 score for hospital admission prediction. P-values are computed using Wilcoxon signed-rank test with Benjamini-Hochberg procedure for false-discovery rate control level $\alpha=0.05$. Statistically significant p-values \texttt{<} 0.05 are depicted in cells with a dark blue background.}
    \label{fig:suppfig5-hospitaladmission_pvalues}
\end{figure}

\subsection{Supplemental Figure 6: ICU Admission - Statistical Significance}\label{suppfig6-section}
\begin{figure}[H]
    \centering
    \includegraphics[width=1\linewidth]{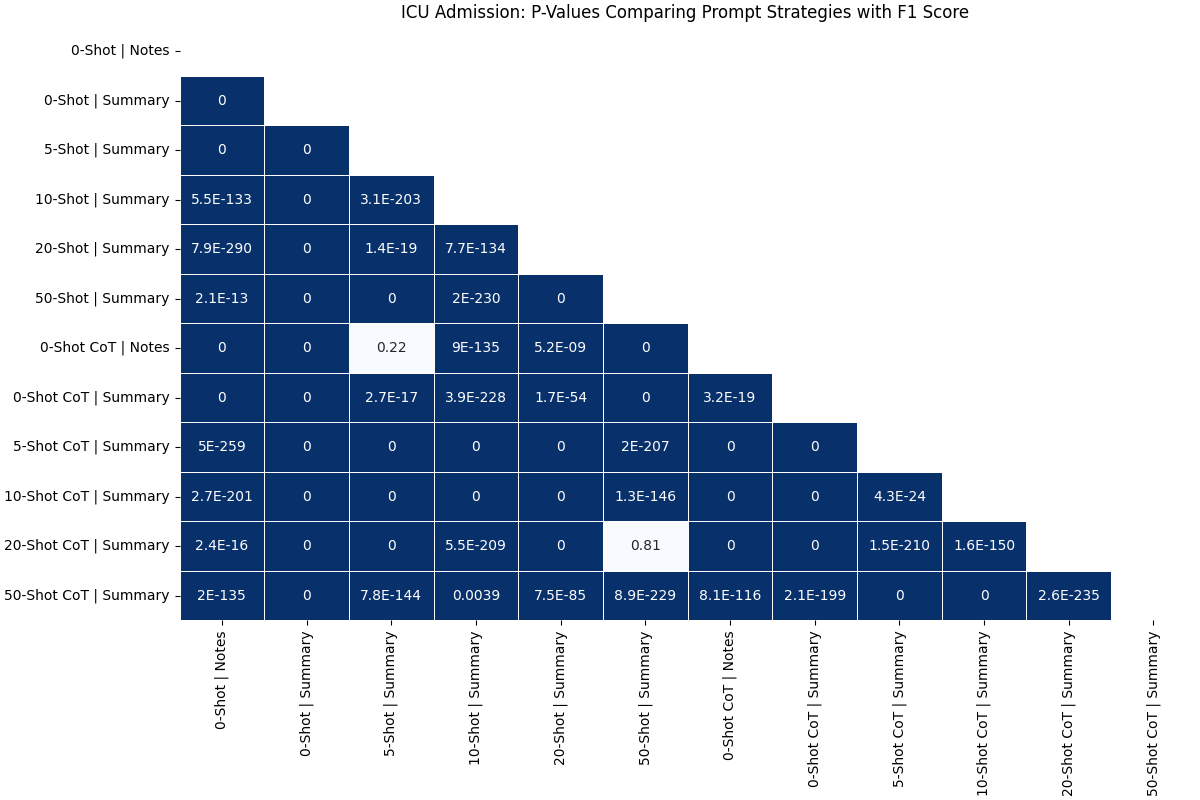}
    \caption{P-values comparing prompt strategies with F1 score for ICU admission prediction. P-values are computed using Wilcoxon signed-rank test with Benjamini-Hochberg procedure for false-discovery rate control level $\alpha=0.05$. Statistically significant p-values \texttt{<} 0.05 are depicted in cells with a dark blue background.}
    \label{fig:suppfig6-icuadmission_pvalues}
\end{figure}

\subsection{Supplemental Figure 7: Unplanned Admission - Statistical Significance}\label{suppfig7-section}
\begin{figure}[H]
    \centering
    \includegraphics[width=1\linewidth]{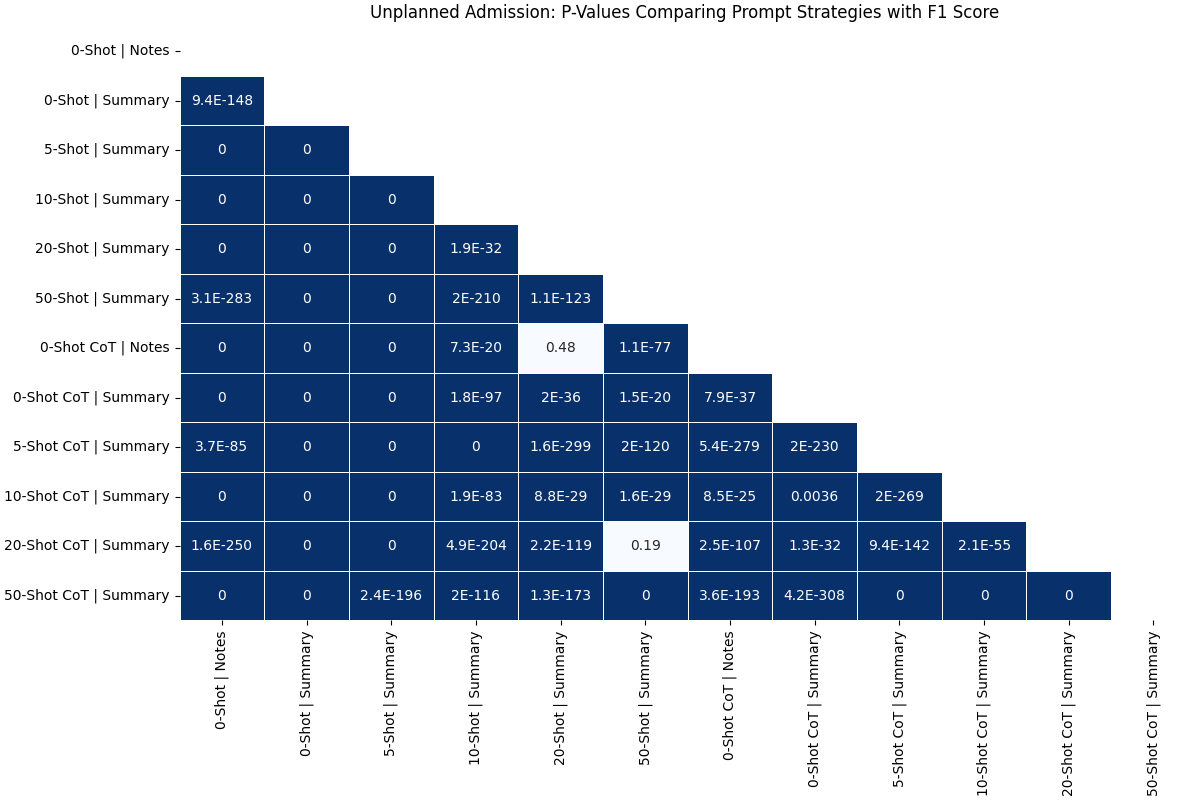}
    \caption{P-values comparing prompt strategies with F1 score for unplanned admission prediction. P-values are computed using Wilcoxon signed-rank test with Benjamini-Hochberg procedure for false-discovery rate control level $\alpha=0.05$. Statistically significant p-values \texttt{<} 0.05 are depicted in cells with a dark blue background.}
    \label{fig:suppfig7-unplannedadmission_pvalues}
\end{figure}

\subsection{Supplemental Figure 8: Hospital Mortality - Statistical Significance}\label{suppfig8-section}
\begin{figure}[H]
    \centering
    \includegraphics[width=1\linewidth]{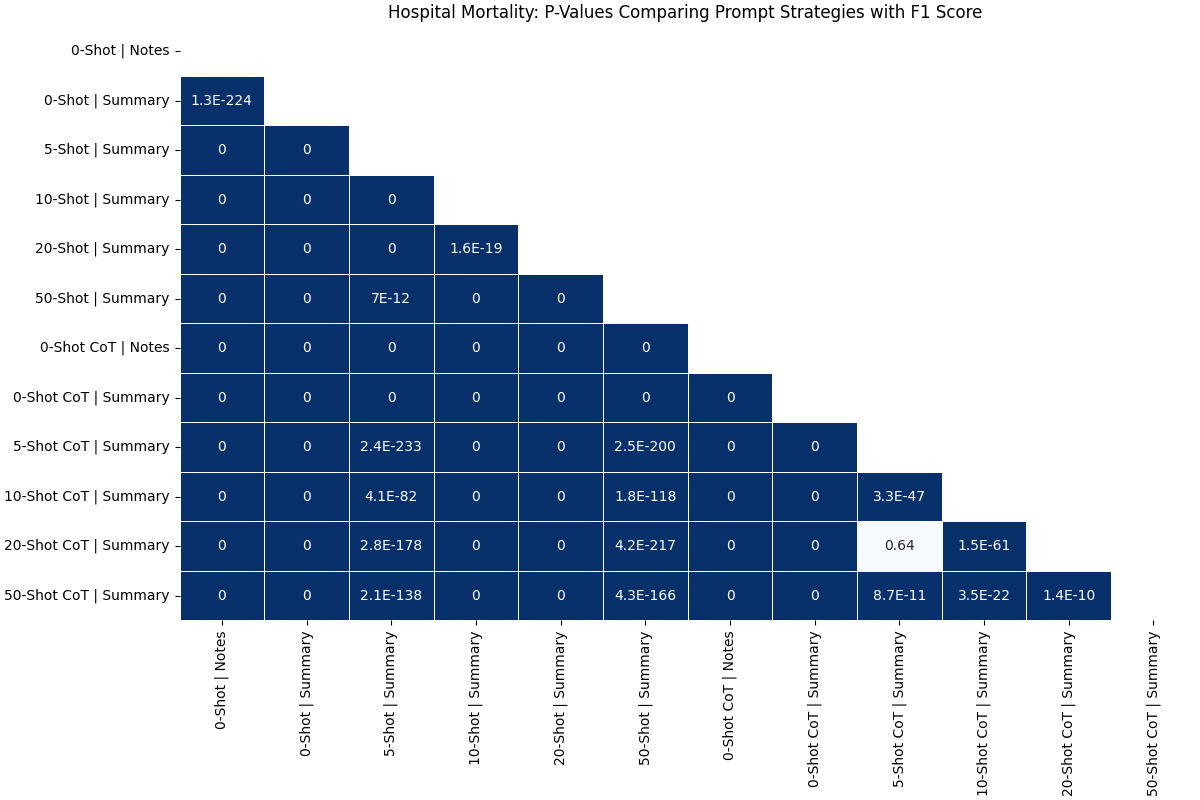}
    \caption{P-values comparing prompt strategies with F1 score for hospital mortality prediction. P-values are computed using Wilcoxon signed-rank test with Benjamini-Hochberg procedure for false-discovery rate control level $\alpha=0.05$. Statistically significant p-values \texttt{<} 0.05 are depicted in cells with a dark blue background.}
    \label{fig:suppfig8-hospitalmortality_pvalues}
\end{figure}

\subsection{Supplemental Figure 9: Post-Anesthesia Care Unit (PACU) - Phase 1 Duration - Statistical Significance}\label{suppfig9-section}
\begin{figure}[H]
    \centering
    \includegraphics[width=1\linewidth]{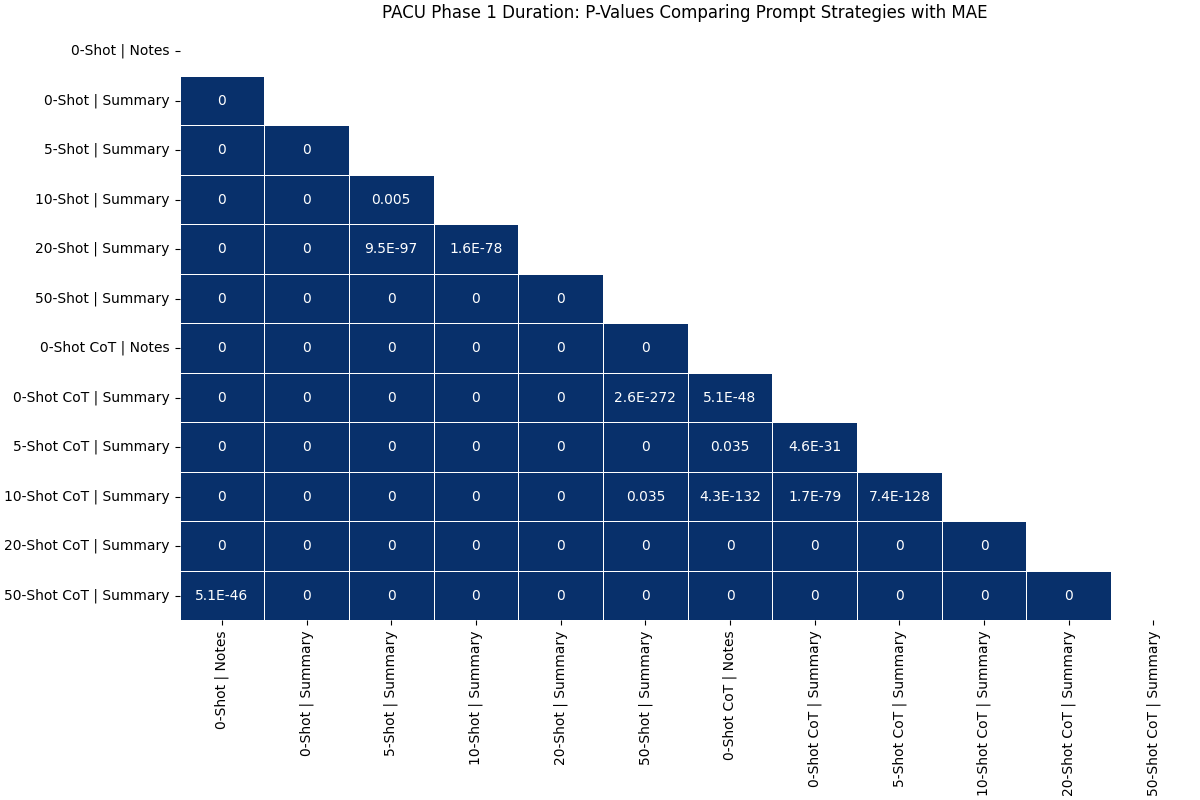}
    \caption{P-values comparing prompt strategies with mean absolute error (minutes) for PACU phase 1 duration prediction. P-values are computed using Wilcoxon signed-rank test with Benjamini-Hochberg procedure for false-discovery rate control level $\alpha=0.05$. Statistically significant p-values \texttt{<} 0.05 are depicted in cells with a dark blue background.}
    \label{fig:suppfig9-phase1duration_pvalues}
\end{figure}

\subsection{Supplemental Figure 10: Hospital Duration - Statistical Significance}\label{suppfig10-section}
\begin{figure}[H]
    \centering
    \includegraphics[width=1\linewidth]{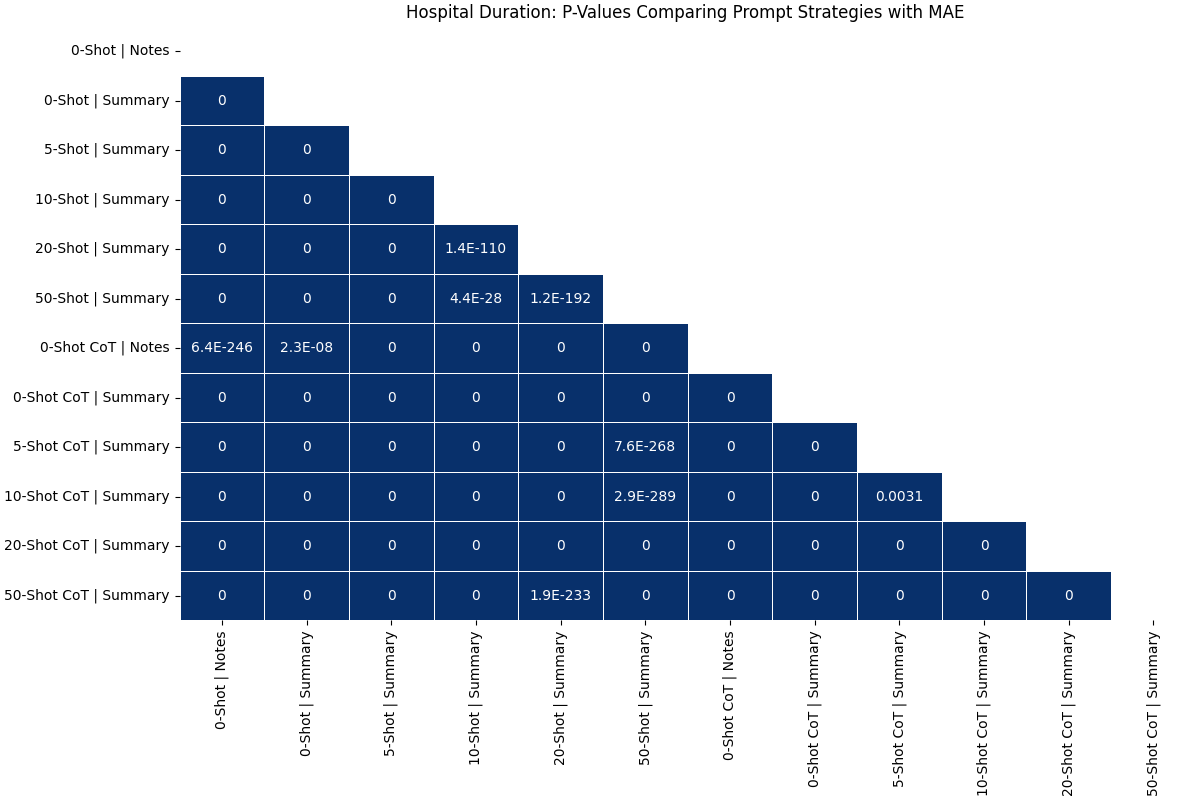}
    \caption{P-values comparing prompt strategies with mean absolute error (days) for hospital duration prediction. P-values are computed using Wilcoxon signed-rank test with Benjamini-Hochberg procedure for false-discovery rate control level $\alpha=0.05$. Statistically significant p-values \texttt{<} 0.05 are depicted in cells with a dark blue background.}
    \label{fig:suppfig10-hospitalduration_pvalues}
\end{figure}

\subsection{Supplemental Figure 11: Intensive Care Unit (ICU) Duration - Statistical Significance}\label{suppfig11-section}
\begin{figure}[H]
    \centering
    \includegraphics[width=1\linewidth]{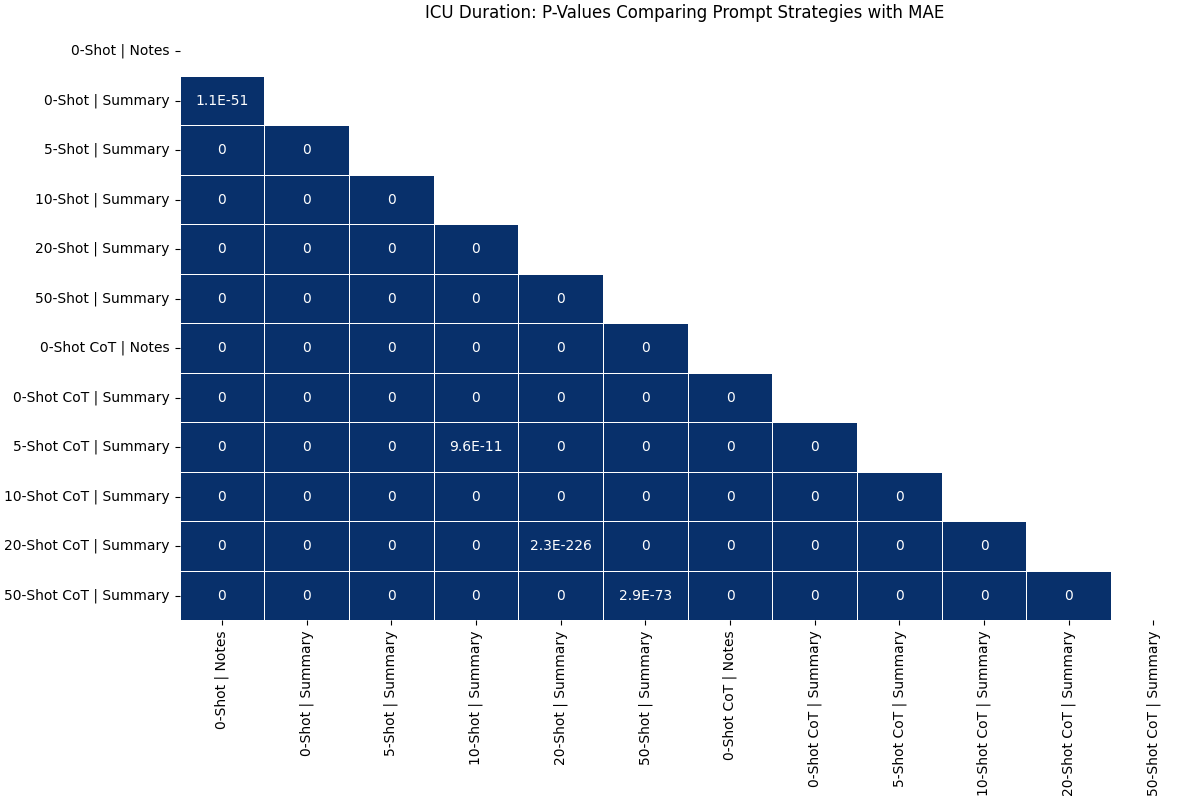}
    \caption{P-values comparing prompt strategies with mean absolute error (days) for ICU duration prediction. P-values are computed using Wilcoxon signed-rank test with Benjamini-Hochberg procedure for false-discovery rate control level $\alpha=0.05$. Statistically significant p-values \texttt{<} 0.05 are depicted in cells with a dark blue background.}
    \label{fig:suppfig11-icuduration_pvalues}
\end{figure}

\subsection{Supplemental Figure 12: ASA Physical Status - Predicted vs. Actual for Each Prompt Strategy}\label{suppfig12-section}
\begin{figure}[H]
    \centering
    \includegraphics[width=1\linewidth]{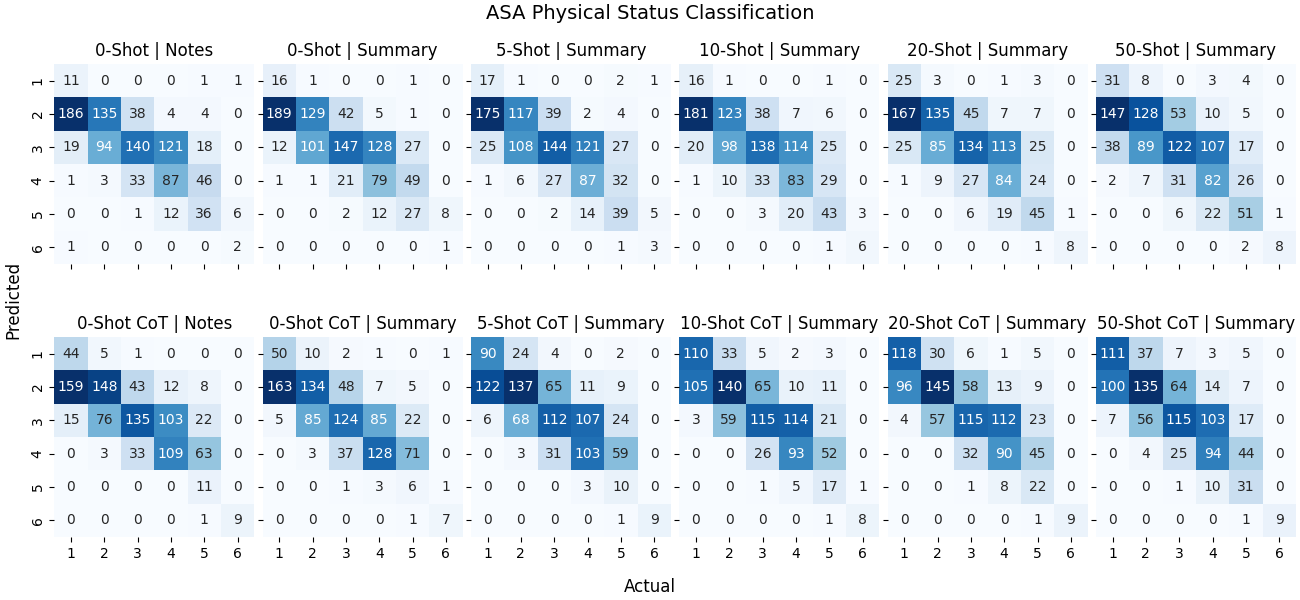}
    \caption{Confusion matrix of predicted and actual ASA Physical Status (ASA-PS) across all 12 prompt strategies. Without few-shot and CoT prompting, predictions are heavily localized to predicting ASA 2 \& 3 (the most common classifications) and the LLM rarely predicts ASA 1 and ASA 6. The addition of few-shot and CoT prompting helps the LLM correctly predict ASA 1 \& 6. When the LLM does make a mis-classification error, it generally predicts an adjacent ASA-PS class and rarely makes egregious prediction errors.}
    \label{fig:suppfig12-asa_confusionmatrix}
\end{figure}

\subsection{Supplemental Figure 13: Hospital Admission - Predicted vs. Actual for Each Prompt Strategy}\label{suppfig13-section}
\begin{figure}[H]
    \centering
    \includegraphics[width=1\linewidth]{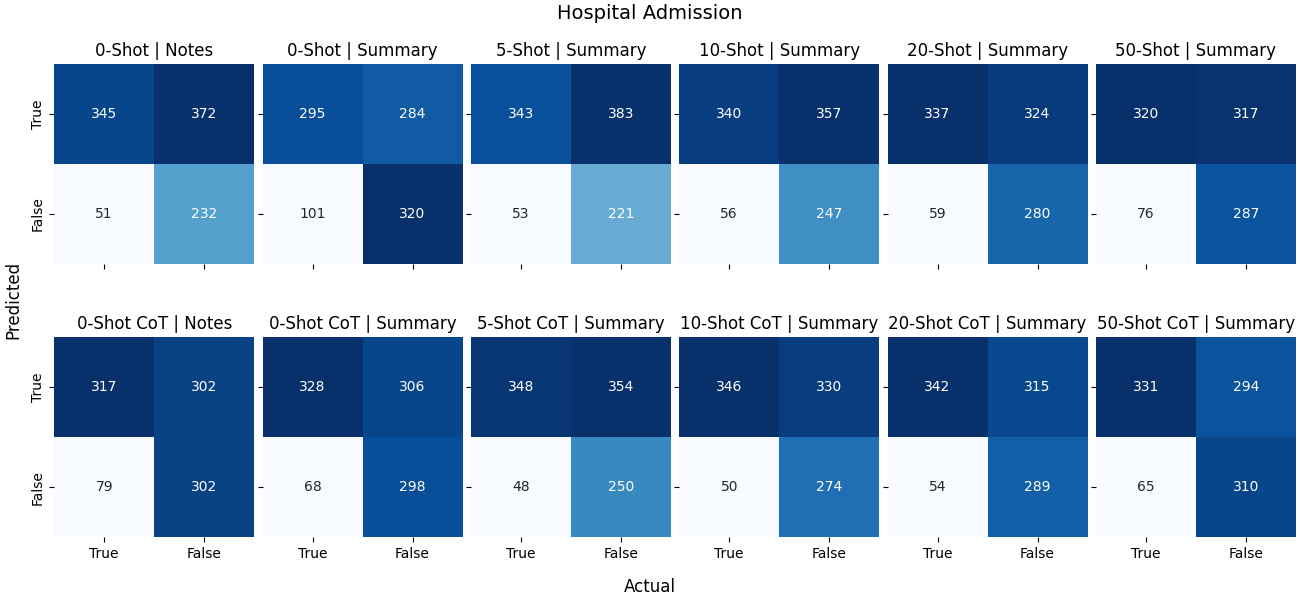}
    \caption{Confusion matrix of predicted and actual postoperative hospital admission across all 12 prompt strategies. All prompt strategies exhibit a high number of false positives, though this is reduced with the addition of few-shot and CoT prompting. The number of false negatives remains low across all prompt strategies.}
    \label{fig:suppfig13-hospitaladmission_confusionmatrix}
\end{figure}

\subsection{Supplemental Figure 14: ICU Admission - Predicted vs. Actual for Each Prompt Strategy}\label{suppfig14-section}
\begin{figure}[H]
    \centering
    \includegraphics[width=1\linewidth]{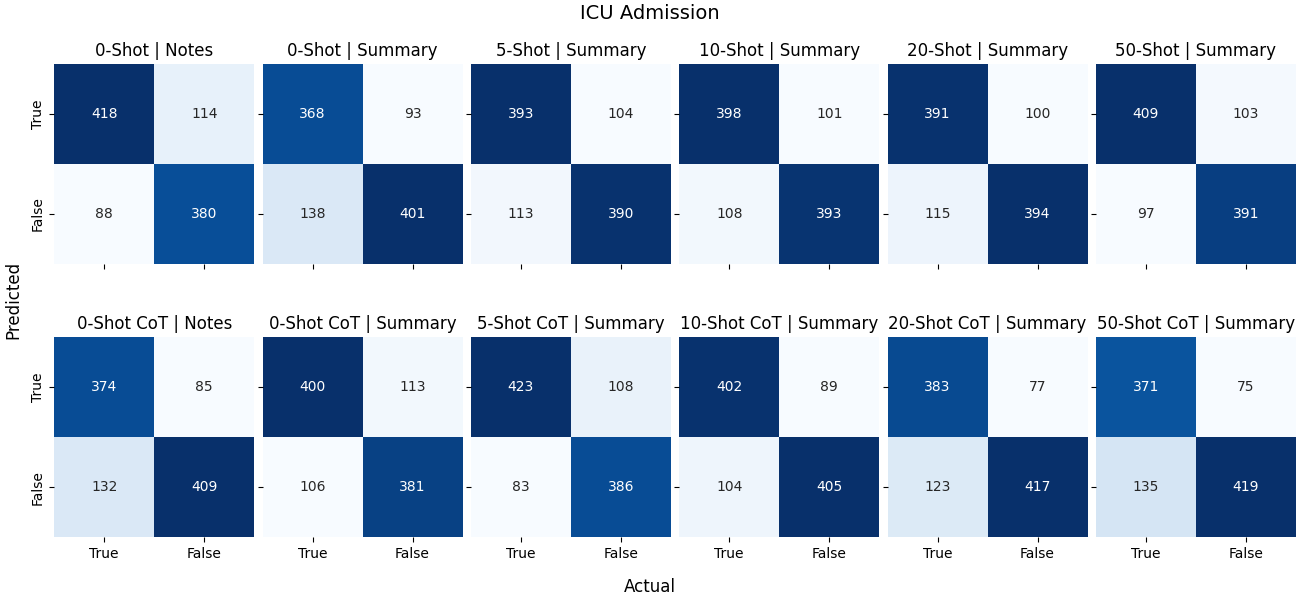}
    \caption{Confusion matrix of predicted and actual postoperative ICU admission across all 12 prompt strategies. All prompt strategies exhibit a high number of true positive and true negatives relative to false positive and false negatives.}
    \label{fig:suppfig14-hospitaladmission_confusionmatrix}
\end{figure}

\subsection{Supplemental Figure 15: Unplanned Admission - Predicted vs. Actual for Each Prompt Strategy}\label{suppfig15-section}
\begin{figure}[H]
    \centering
    \includegraphics[width=1\linewidth]{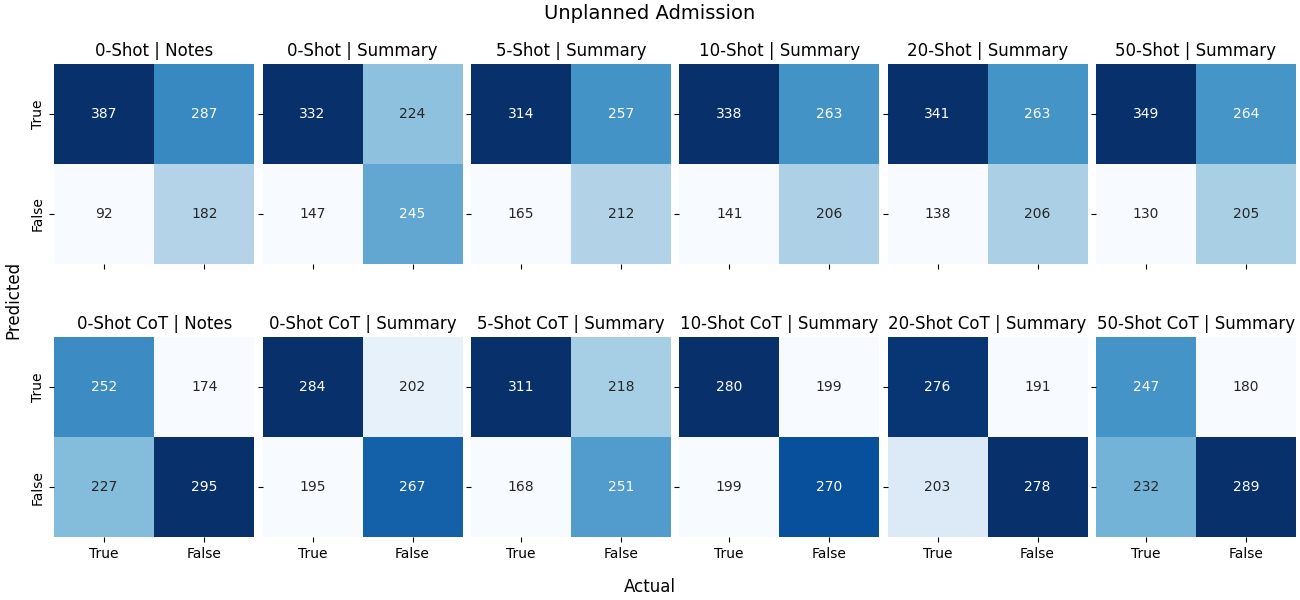}
    \caption{Confusion matrix of predicted and actual postoperative unplanned admissions across all 12 prompt strategies. Zero-shot approaches without CoT have the best performance with low false negatives, but high false positives. Few-shot and CoT strategies steer the LLM from predicting unplanned admissions and increase the number of “False” predictions. Even though this improves the true negative rate, it results in a significant increase in false negatives and worsens overall prediction performance.}
    \label{fig:suppfig15-unplannedadmit_confusionmatrix}
\end{figure}

\subsection{Supplemental Figure 16: Hospital Mortality - Predicted vs. Actual for Each Prompt Strategy}\label{suppfig16-section}
\begin{figure}[H]
    \centering
    \includegraphics[width=1\linewidth]{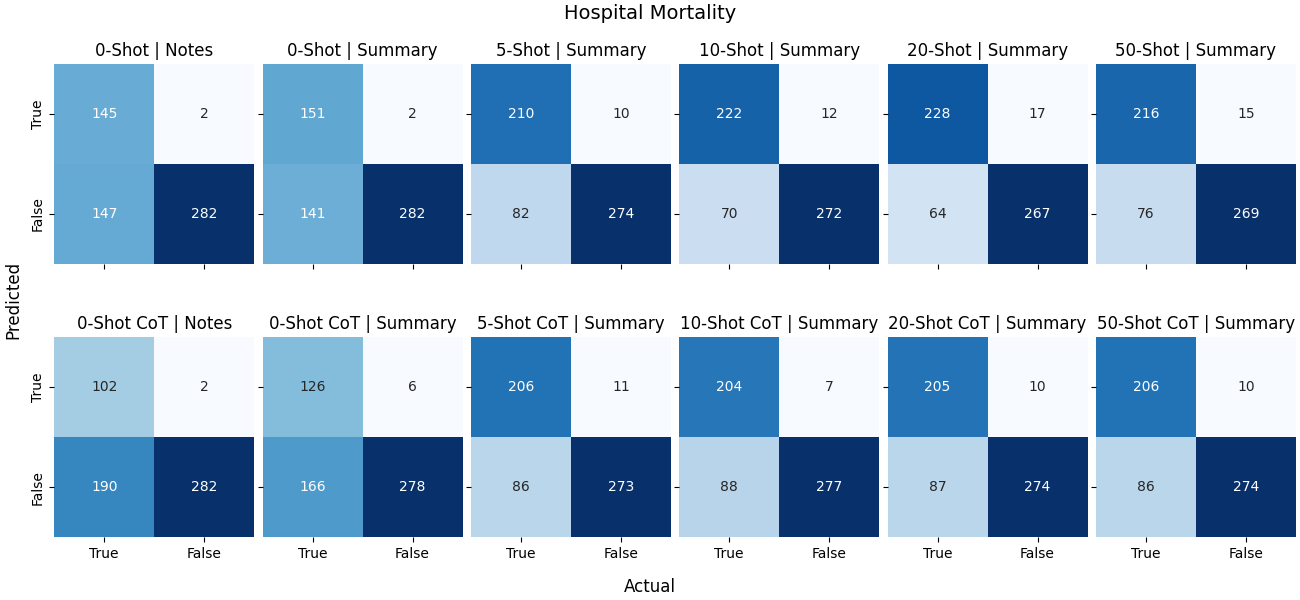}
    \caption{Confusion matrix of predicted and actual postoperative hospital mortality across all 12 prompt strategies. The LLM rarely makes a false positive prediction, regardless of prompt strategy. The addition of few-shot and CoT dramatically reduces the number of false negatives and results in improved overall prediction performance.}
    \label{fig:suppfig16-hospitalmortality_confusionmatrix}
\end{figure}


\subsection{Supplemental Figure 17: PACU Phase 1 Duration - Predicted vs. Actual for Each Prompt Strategy}\label{suppfig17-section}
\begin{figure}[H]
    \centering
    \includegraphics[width=1\linewidth]{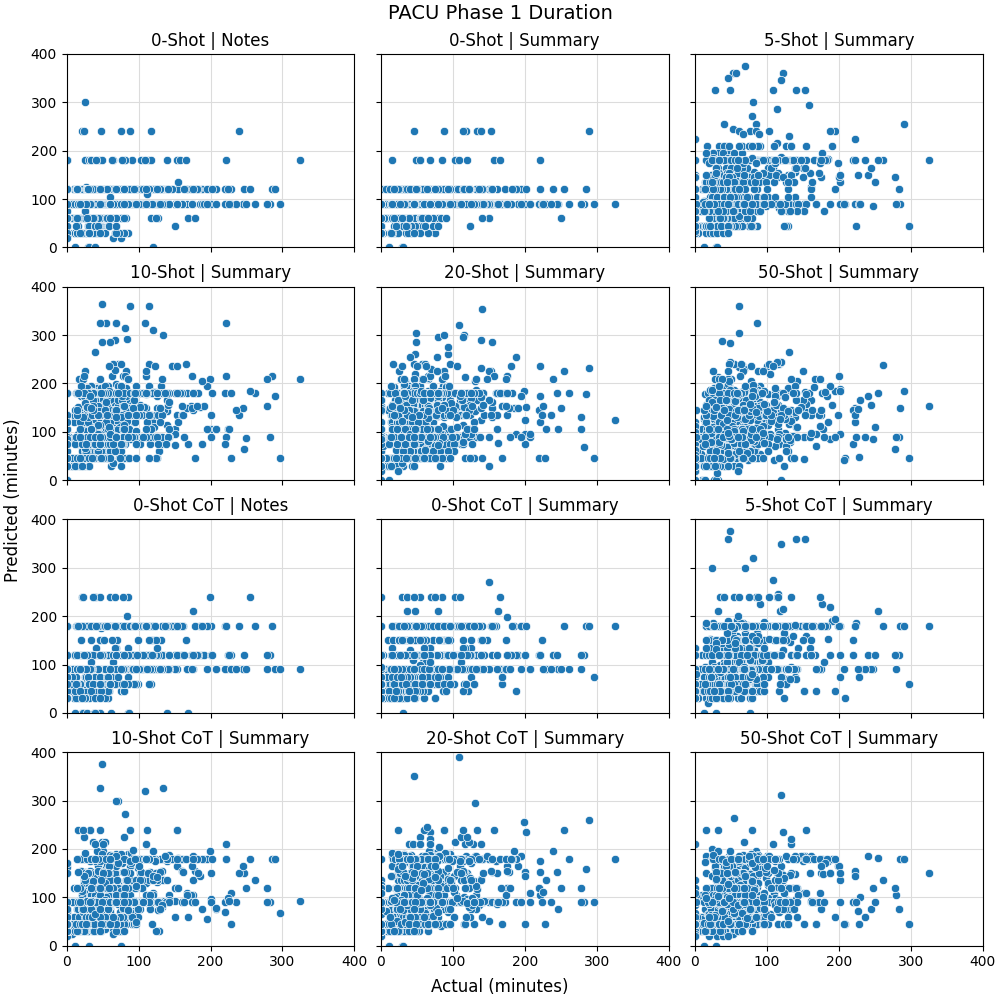}
    \caption{This is the same figure as Figure 4, replicated here for completeness. Scatter plot of predicted and actual post-anesthesia care unit (PACU) Phase 1 recovery durations across all 12 prompt strategies. Without few-shot and CoT prompting, predictions are heavily quantized to specific values and exhibit a ceiling effect where the LLM rarely predicts beyond 180 minutes. The progressive addition of few-shot and CoT prompting removes this effect, but predictive performance remains poor.}
    \label{fig:suppfig17-phase1duration_scatter}
\end{figure}

\subsection{Supplemental Figure 18: Hospital Duration - Predicted vs. Actual for Each Prompt Strategy}\label{suppfig18-hospitalduration_scatter}
\begin{figure}[H]
    \centering
    \includegraphics[width=1\linewidth]{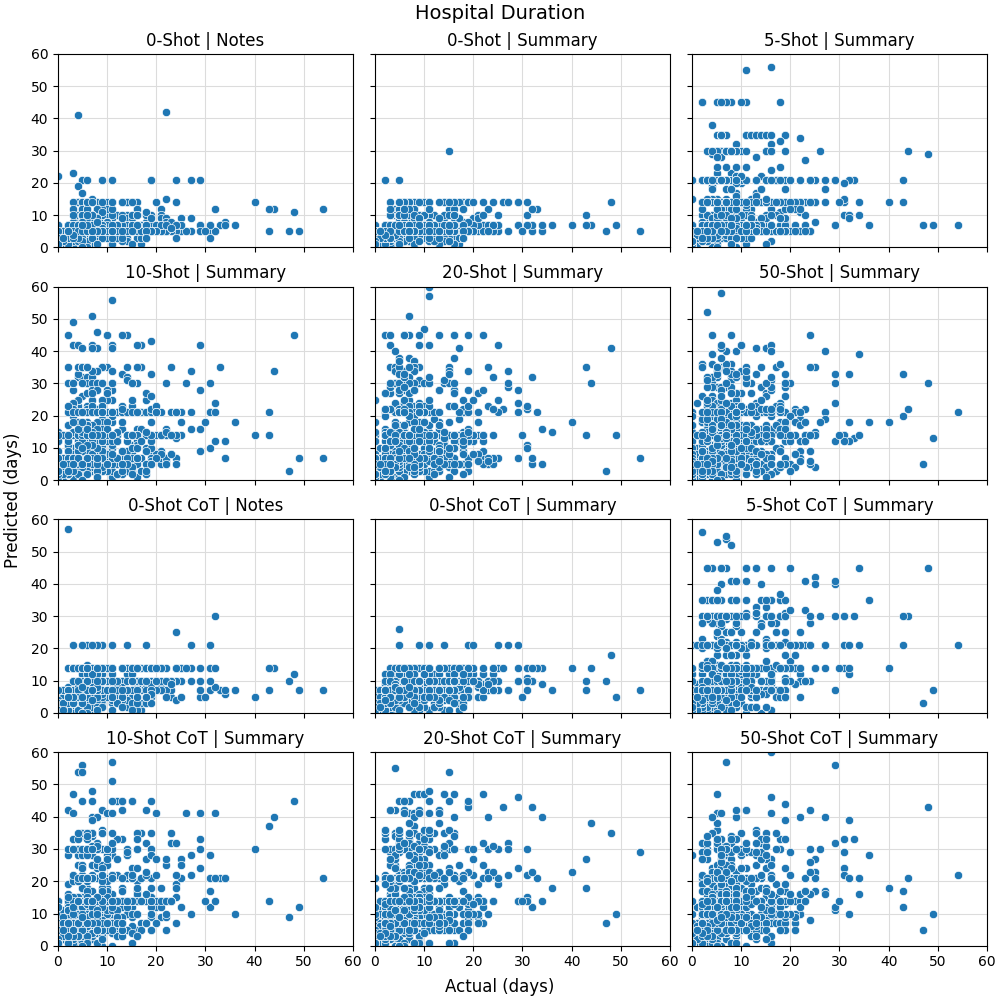}
    \caption{Scatter plot of predicted versus actual postoperative hospital admission durations across all 12 prompt strategies. Without few-shot and CoT prompting, predictions are heavily quantized to specific values and exhibit a ceiling effect where the LLM rarely predicts beyond 14 days. The progressive addition of few-shot and CoT prompting removes this effect, but predictive performance remains poor.}
    \label{fig:suppfig18-hospitalduration_scatter}
\end{figure}

\subsection{Supplemental Figure 19: ICU Duration - Predicted vs. Actual for Each Prompt Strategy}\label{suppfig19-icuduration_scatter}
\begin{figure}[H]
    \centering
    \includegraphics[width=1\linewidth]{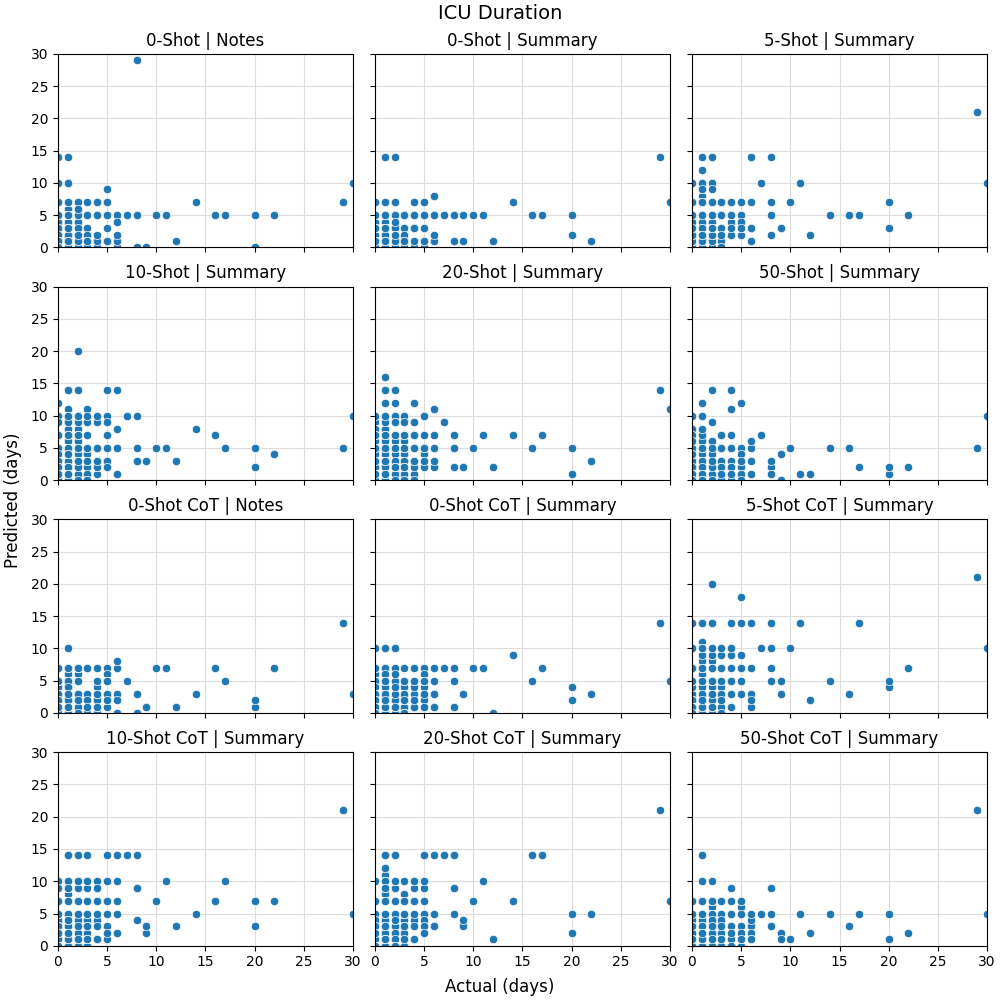}
    \caption{Scatter plot of predicted versus actual postoperative ICU admission durations across all 12 prompt strategies. Without few-shot prompting, predictions exhibit a ceiling effect where the LLM rarely predicts beyond 10 days. The progressive addition of few-shot prompting removes this effect, but predictive performance remains poor.}
    \label{fig:suppfig19-icuduration_scatter}
\end{figure}




\end{appendices}

\pagebreak
\bibliography{article}

\end{document}